\documentclass[english]{svjour3}

\PassOptionsToPackage{boxruled,linesnumbered}{algorithm2e}
\usepackage[T1]{fontenc}
\usepackage[latin9]{inputenc}
\usepackage{setspace}
\usepackage{geometry}
\makeatletter
\let\vec\relax
\DeclareMathAccent{\vec}{\mathord}{letters}{"7E}

\usepackage{
algorithm2e,
amsmath,
amssymb,
amsthm,
array,
appendix,
bm,
bbm,
color,
epsf,
enumitem,
float,
graphicx,
hyperref,
listings,
mathtools,
mathrsfs,
pifont,
setspace,
textcomp,
tikz,
titletoc,
units,
url,
ulem,
xcolor,
comment,
cite
}

\hypersetup{
    colorlinks=true, 
    linktoc=all,     
    linkcolor=black,  
}

\graphicspath{{figures/}}

\newcommand{\Cbb}{\mathbb{C}}
\newcommand{\Ebb}{\mathbb{E}}

\newcommand{\Vbb}{\mathbb{V}}
\newcommand{\Zbb}{\mathbb{Z}}

\newcommand{\bbf}{\mathbf{b}}
\newcommand{\Bbf}{\mathbf{B}}

\newcommand{\Cbf}{\mathbf{C}}

\newcommand{\ebf}{\mathbf{e}}

\newcommand{\fbf}{\mathbf{f}}
\newcommand{\Fbf}{\mathbf{F}}

\newcommand{\Gbf}{\mathbf{G}}
\newcommand{\hbf}{\mathbf{h}}
\newcommand{\Hbf}{\mathbf{H}}

\newcommand{\Ibf}{\mathbf{I}}

\newcommand{\Lbf}{\mathbf{L}}
\newcommand{\mbf}{\mathbf{m}}

\newcommand{\Pbf}{\mathbf{P}}

\newcommand{\Qbf}{\mathbf{Q}}
\newcommand{\rbf}{\mathbf{r}}
\newcommand{\Rbf}{\mathbf{R}}

\newcommand{\Sbf}{\mathbf{S}}
\newcommand{\tbf}{\mathbf{t}}

\newcommand{\ubf}{\mathbf{u}}
\newcommand{\Ubf}{\mathbf{U}}

\newcommand{\Vbf}{\mathbf{V}}
\newcommand{\wbf}{\mathbf{w}}
\newcommand{\Wbf}{\mathbf{w}}

\newcommand{\ep}{\epsilon}

\newcommand{\CalF}{{\mathcal{F}}}

\newcommand{\CalL}{{\mathcal{L}}}

\newcommand{\CalN}{{\mathcal{N}}}
\newcommand{\CalO}{{\mathcal{O}}}

\newcommand{\nrm}[1]{\left\Vert {#1} \right\Vert}

\newcommand{\supp}[1]{\text{supp}\left(#1\right)}

\renewcommand{\tilde}[1]{\widetilde{#1}}

\newcommand{\argmin}{\text{argmin}}

\newcommand{\wstar}{\wbf^\star}
\newcommand{\what}{{\widehat{\wbf}}}

\def\bigdot#1{\overset{\hbox{\LARGE .}}{#1}}

\newcommand{\flr}[1]{{\lfloor {#1} \rfloor}}

\providecommand{\tabularnewline}{\\}

\RequirePackage{fix-cm}

\smartqed  

\usepackage{cite}

\makeatother

\usepackage{babel}

\begin{document}
\title{Direct Estimation of Parameters in ODE Models Using WENDy: Weak-form Estimation of Nonlinear Dynamics
\thanks{This research was supported in part by the following grants: NSF Mathematical Biology MODULUS grant 2054085 to DMB; NSF/NIH Joint DMS/NIGMS Mathematical Biology Initiative grant R01GM126559 to DMB; DOE ASCR MMICC grant DE-SC0023346 to DMB; NIFA Biological Sciences grant 2019-67014-29919 to VD; and NSF Division Of Environmental Biology grant 2109774 to VD.  Software and code for reproducing the examples is available at \url{https://github.com/MathBioCU/WENDy}.}}
\titlerunning{Weak form Estimation of Nonlinear Dynamics}
\author{David M.~Bortz  \and Daniel A.~Messenger \and Vanja Dukic\footnote{Prof.~Dukic holds concurrent appointments as a Professor at the University of Colorado at Boulder and as an Amazon Scholar. This publication describes work performed at the University of Colorado at Boulder and is not associated with Amazon.}}
\authorrunning{Bortz, Messenger, Dukic}
\institute{David M. Bortz\at Department of Applied Mathematics\\University of Colorado, Boulder, CO 80309-0526\\ Tel.: +1 303-492-7569\\ Fax: +1 303-492-4066\\ \email{david.bortz@colorado.edu}}
\date{Received: date / Accepted: date}
\maketitle
\begin{abstract}
We introduce the Weak-form Estimation of Nonlinear Dynamics (WENDy) method for estimating model parameters for non-linear systems of ODEs. Without relying on any numerical differential equation solvers, WENDy computes accurate estimates and is robust to large (biologically relevant) levels of measurement noise. For low dimensional systems with modest amounts of data, WENDy is competitive with conventional forward solver-based nonlinear least squares methods in terms of speed and accuracy.  For both higher dimensional systems and stiff systems, WENDy is typically both faster (often by orders of magnitude) and more accurate than forward solver-based approaches.

The core mathematical idea involves an efficient conversion of the strong form representation of a model to its weak form, and then solving a regression problem to perform parameter inference. The core statistical idea  rests on the  Errors-In-Variables framework, which necessitates the use of the iteratively reweighted least squares algorithm.   Further improvements are obtained by using orthonormal test functions, created from a set of $C^{\infty}$ bump functions of varying support sizes.
 
We demonstrate the high robustness and computational efficiency by applying WENDy to estimate parameters in some common models from population biology, neuroscience, and biochemistry, including logistic growth, Lotka-Volterra, FitzHugh-Nagumo, Hindmarsh-Rose, and a Protein Transduction Benchmark model. Software and code for reproducing the examples is available at \url{https://github.com/MathBioCU/WENDy}.
\end{abstract}



\section{Introduction\label{sec:Introduction}}

Accurate estimation of parameters for a given model is central to modern scientific discovery. It is particularly important in the modeling of biological systems which can involve both first principles-based and phenomenological models and for which measurement errors can be substantial, often in excess of 20\%. The dominant methodologies for parameter inference are either not capable of handling realistic errors, or  are  computationally costly relying on forward solvers or Markov chain Monte Carlo methods. In this work, we propose an accurate, robust and efficient weak form-based approach to estimate parameters for parameter inference. We demonstrate that our ``Weak form Estimation of Nonlinear Dynamics'' (WENDy) method offers many advantages including high accuracy, robustness to substantial noise, and computational efficiency often up to several orders of magnitude over the existing methods. 

In the remainder of this section, we provide an overview of modern parameter estimation methods in ODE systems, as well as a discussion of the literature that led to the WENDy idea. Section \ref{sec:WENDy} contains the core weak-form estimation ideas as well as the WENDy algorithm itself.  In Section \ref{sec:WfOLS}, we introduce the idea of weak-form parameter estimation, including a simple algorithm to illustrate the idea.  In Section \ref{sec:WENDy_IRLS}, we describe the WENDy method in detail.  
We describe the Errors-In-Variables (EiV) framework, and derive a Taylor expansion of the residual which allows us to  formulate the 
(in Section \ref{sec:WENDy_IRLS}) Iteratively Reweighted Least Squares (IRLS) approach to inference. The EiV and IRLS modifications are important as they offers significant improvements to the Ordinary Least Squares approach. In Section \ref{sec:TestFunction}, we  present a strategy for computing an  orthogonal set of test functions that facilitate  a successful weak-form implementation.  In Section \ref{sec:Illustrating-Examples} we illustrate the performance of WENDy using five common mathematical models from the biological sciences and in Section \ref{sec:ConcDisc} we offer some concluding remarks.

\subsection{Background}
A ubiquitous version of the parameter estimation problem in the biological sciences is
\begin{equation}
\widehat{\wbf}:={\arg \min_{\wbf\in \mathbb{R}^{J}}} \|u(\mathbf{t};\wbf)-\mathbf{U}\|_{2}^{2},\label{eq:ParEstProblem}
\end{equation}
where the function $u:\mathbb{R}\to\mathbb{R}^d$ is a solution to a differential equation model\footnote{While we restrict ourselves to deterministic differential equations, there is nothing in the WENDy approach that inhibits extension to discrete or stochastic models.}
\begin{equation}
\begin{array}{rl}
\dot{u}&=\sum_{j=1}^{J}w_j f_j(u),\label{eq:UPE_DE}\\
u(t_0)&=u_0\in\mathbb{R}^{d},
\end{array}
\end{equation}
The ODE system in \eqref{eq:UPE_DE} is parameterized by $\wbf\in \mathbb{R}^{J}$, the vector of $J$ true parameters  which are to be estimated by $\widehat{\wbf}$. 
The solution to the equation is then compared (in a least squares sense) with data $\mathbf{U}\in\mathbb{R}^{(M+1)\times d}$ that is sampled at $M+1$ timepoints $t:=\{t_i\}_{i=0}^{M}$.  We note that in this work, we will restrict the differential equations to those with right sides that are linear combinations of the $f_j$ functions with coefficients $w_j$, as in equation \eqref{eq:UPE_DE}.

Conventionally, the standard approach for parameter estimation methodologies has been forward solver-based nonlinear least squares (FSNLS). In that framework, 1) a candidate parameter vector is proposed, 2) the resulting equation is numerically solved on a computer, 3) the output is compared (via least squares) to data, and 4) then this process is repeated until a convergence criteria is met. This is a mature field and we direct the interested reader to references by Ljung \cite{Ljung1999,Ljung2017WileyEncyclopediaofElectricalandElectronicsEngineering} and,  for those interested in a more theoretical perspective, to the monograph by Banks and Kunisch \cite{BanksKunisch1989}. 

The FSNLS methodology is very well understood and its use is ubiquitous in the biological, medical, and bioengineering sciences. However, as models get larger and more realism is demanded of them, there remain several important challenges that do not have fully satisfying answers.  For example, the accuracy of the solver can have a huge impact on parameter estimates; see \cite{NardiniBortz2019InverseProbl} for an illustration with PDE models and \cite{Bortz2006JCritCare} for an example with ODE and DDE models. There is no widespread convention on detection of this type of error and the conventional strategy would be to simply increase the solution accuracy (usually at significant computational cost) until the estimate stabilizes.  Perhaps more importantly, the choice of the initial candidate parameter vector can have a huge impact upon the final estimate, given that nonlinear least squares cost functions frequently have multiple local minima in differential equations applications.  There are several algorithms designed to deal with the multi-modality, such as particle swarm optimization \cite{BonyadiMichalewicz2017EvolComput} and simulated annealing \cite{vanLaarhovenAarts1987}; however, all come at the cost of additional forward solves and unclear dependence on the hyperparameters used in the solver and optimization algorithms.

Given the above, it is reasonable to consider alternatives to fitting via comparing an approximate model solution with the measured data.  A natural idea would be to avoid performing forward solves altogether via substituting the data directly into the model equation \eqref{eq:UPE_DE}. The derivative could be approximated via differentiating a projection of the data onto, e.g., orthogonal polynomials, and the parameters could then be estimated by minimizing the norm of the residual of the equation \eqref{eq:UPE_DE} --  i.e., via a gradient matching criteria. Indeed, Richard Bellman proposed exactly this strategy in 1969 \cite{Bellman1969MathematicalBiosciences}. There have been similar ideas in the literature of chemical and aerospace engineering, which can be traced back even further \cite{PerdreauvilleGoodson1966JBasicEng, Greenberg1951NACATN2340}.  However, these methods are known to  perform poorly in the presence of even modest noise.

To account for the noise in the measurements while estimating the parameters (and in some cases the state trajectories), researchers have proposed a variety of different non-solver-based methods. The most popular modern approaches involve denoising the measured state via Gaussian Processes \cite{YangWongKou2021ProcNatlAcadSciUSA,Martina-PerezSimpsonBaker2021ProcRSocA,WangZhou2021IntJUncertaintyQuantification,WenkAbbatiOsborneEtAl2020AAAI,CalderheadGirolamiLawrence2008AdvNeuralInfProcessSyst} and collocations projecting onto a polynomial or spline basis \cite{Varah1982SIAMJSciandStatComput, RamsayHookerCampbellEtAl2007JRStatSocSerBStatMethodol,LiangWu2008JournaloftheAmericanStatisticalAssociation,PoytonVarziriMcAuleyEtAl2006ComputersChemicalEngineering, Brunel2008ElectronJStat,ZhangNanshanCao2022StatComput}. For example, Yang et al. \cite{YangWongKou2021ProcNatlAcadSciUSA}, restricted a Gaussian Process to the manifold of solutions to an ODE to infer both the parameters and the state using a Hamiltonian Markov chain Monte Carlo method.  Ramsey et al. \cite{RamsayHookerCampbellEtAl2007JRStatSocSerBStatMethodol} proposed a collocation-type method in which the solution is projected onto a spline basis. In a two-step procedure, both the basis weights and the unknown parameters are iteratively estimated. The minimization identifies the states and the parameters by penalizing poor faithfulness to the model equation (i.e., gradient matching) and deviations too far from the measured data.
Liang and Wu \cite{LiangWu2008JournaloftheAmericanStatisticalAssociation} proposed a similar strategy based on local polynomial smoothing to first estimate the state and its derivative, compute derivatives of the smoothed solution, and then estimate the parameters.  Ding and Wu later improved upon this work in \cite{DingWu2014StatSin} by using local polynomial regression instead of the pseudo-least squares estimator used in \cite{LiangWu2008JournaloftheAmericanStatisticalAssociation}. 

There are also a few approaches which focus on transforming the equations with operators that allow efficiently solving for the parameters. 
 In particular Xu and Khanmohamadi created smoothing and derivative smoothing operators based on Fourier theory \cite{XuKhanmohamadi2008Chaos} and Chebyshev operators \cite{KhanmohamadiXu2009Chaos}. However, they have not proven to be as influential as the integral and weak form methods described in the next subsection.

 \subsection{Integral and Weak Form Methods}

Recent efforts by our group and others suggest that there is a considerable advantage in parameter estimation performance to be gained from using an integral-based transform of the model equations.  The two main approaches are to 1) use integral forms of the model equation or 2) convolve the equation with a compactly supported test function to obtain the so-called "weak form" of the equation.  The weak form idea can be traced back to Laurent Schwartz's Theory of Distributions \cite{Schwartz1950}\footnote{See \cite{DuistermaatKolk2010} for a modern introduction.}, which recasts the classical notion of a function acting on a point to one acting on a measurement structure or "test function". In the context of differential equation models, Lax and Milgram pioneered the use of the weak form for relaxing smoothness requirements on unique solutions to parabolic PDE systems in Hilbert spaces \cite{LaxMilgram1955ContributionstotheTheoryofPartialDifferentialEquations}. Since then, the weak form has been heavily used in studying solutions to PDEs as well as numerically solving for the solutions (e.g., the Finite Element Method), but not with the goal of directly estimating parameters.

The idea of weak-form based estimation has been repeatedly discovered over the years (see \cite{PreisigRippin1993ComputChemEng} for a good historical overview). Briefly, in 1954, Shinbrot created a proto-weak-form parameter inference method, called the Equations Of Motion (EOM) method \cite{Shinbrot1954NACATN3288}. In it, he proposes to multiply the model equations by so-called method functions, i.e., what we would now call test functions.  These test functions were based on $\sin^n(\nu t)$ for different values of $\nu$ and $n$. In 1965, Loeb and Cahen \cite{LoebCahen1965Automatisme, LoebCahen1965IEEETransAutomControl} independently discovered the same method, calling it the Modulating Function (MF) method.  They proposed and advocated for the use of polynomial test functions. The issue with these approaches (and indeed all subsequent developments based on these methods) is that the maximum power $n$ is chosen to exactly match the number of derivatives needed to perform integration by parts (IBP). As we have shown, this choice means that these methods are not nearly as effective as they could be. As we initially reported in \cite{MessengerBortz2021MultiscaleModelSimul}, a critical step in obtaining robust and accurate parameter estimation is to use \emph{highly} smooth test functions, e.g., to have $n$ be substantially higher than the minimum needed by the IBP.  This insight led to our use of the $C^{\infty}$ bump functions in WENDy (see Section \ref{sec:TestFunction}).

In the statistics literature, there are several examples of using integral or weak-form equations.  Dattner et al. \cite{DattnerMillerPetrenkoEtAl2017JRSocInterface} illustrate an integral-based approach and Dattner's 2021 review \cite{Dattner2021WIREsCompStat} 
provides a good overview of other efforts to use the integral form for parameter estimation.  Concerning the weak form, several researchers have used it as a core part of their estimation methods \cite{BrunelClairondAlche-Buc2014JAmStatAssoc,Sangalli2021InternationalStatisticalReview}. Unlike WENDy, however, either these approaches smooth the data before substitution into the model equation (which can lead to poor performance) or still require forward solves. As with the EOM and MF method above, the test functions in these methods were also chosen with insufficient smoothness to yield the highly robust parameter estimates we obtain with WENDy.

As the field of SINDy-based equation learning \cite{BruntonProctorKutz2016ProcNatlAcadSci} is built upon direct parameter estimation methods, there are also several relevant contributions from this literature. Schaeffer and McCalla \cite{SchaefferMcCalla2017PhysRevE} showed that parameter estimation and learning an integral form of equations can be done in the presence of significant noise. Broadly speaking, however, the consensus has emerged that the weak form is more effective than a straightforward integral representation. In particular, several groups (including ours) independently proposed weak form-based approaches \cite{PantazisTsamardinos2019Bioinformatics,GurevichReinboldGrigoriev2019Chaos,MessengerBortz2021MultiscaleModelSimul,PantazisTsamardinos2019Bioinformatics,WangHuanGarikipati2019ComputMethodsApplMechEng, MessengerBortz2021JComputPhys}. The weak form is now even implemented in the PySINDy code \cite{KaptanogludeSilvaFaselEtAl2022JOSS} which is actively developed by the authors of the original SINDy papers \cite{BruntonProctorKutz2016ProcNatlAcadSci,RudyBruntonProctorEtAl2017SciAdv}.  However, we do note that the Weak SINDy in PySINDy is based on an early weak form implementation (proposed in \cite{GurevichReinboldGrigoriev2019Chaos,ReinboldGurevichGrigoriev2020PhysRevE}).  A more recent implementation with autotuned hyperparameters can be found at \url{https://github.com/MathBioCU/WSINDy_ODE} for ODEs \cite{MessengerBortz2021MultiscaleModelSimul} and \url{https://github.com/MathBioCU/WSINDy_PDE} for PDEs \cite{MessengerBortz2021JComputPhys}.

While our group wasn't the first to propose a weak form methodology, we have pioneered its use for equation learning in a wide range of model structures and applications including: ODEs \cite{MessengerBortz2021MultiscaleModelSimul}, PDEs \cite{MessengerBortz2021JComputPhys},  interacting particle systems of the first \cite{MessengerBortz2022PhysicaD} and second \cite{MessengerWheelerLiuEtAl2022JRSocInterface} order, and online streaming \cite{MessengerDallAneseBortz2022ProcThirdMathSciMachLearnConf}.  We have also studied and advanced the computational method itself. Among other contributions, we were the first to automate (with mathematical justification) test function hyperparameter specification, feature matrix rescaling (to ensure stable computations), and 
to filter high frequency noise \cite{MessengerBortz2021JComputPhys}.  Lastly we have also studied the theoretical convergence properties for WSINDy in the continuum data limit
\cite{MessengerBortz2022arXiv221116000}.  Among the results are a description of a broad class of models for which the asymptotic limit of continuum data can overcome \emph{any} noise level to yield both an accurately learned equation and a correct parameter estimate (see \cite{MessengerBortz2022arXiv221116000} 
 for more information).  

\section{Weak form Estimation of Nonlinear Dynamics (WENDy)\label{sec:WENDy}}

In this work, we assume that the exact form of a differential equation-based mathematical
model is known, but that the precise values of constituent parameters are
to be estimated using existing data. As the model equation is not being learned,
this is different than the  WSINDy methodology and, importantly, does not use
sparse regression. We thus denote the method presented in this paper as the Weak-form Estimation
of Nonlinear Dynamics (WENDy) method.

In Section \ref{sec:WfOLS}, we start with an  introduction to the idea of weak-form parameter estimation in a simple OLS setting.  In Section \ref{sec:WENDy_IRLS} we describe the WENDy algorithm in detail, along with several strategies for improving the accuracy: in Section \ref{sec:TestFunction} we describe a strategy for optimal test function selection, and in Section \ref{sec:SC} the strategy for improved iteration termination criteria.

\subsection{Weak-form Estimation with Ordinary Least Squares\label{sec:WfOLS}}

We begin by considering a $d$-dimensional matrix form of \eqref{eq:UPE_DE}, i.e., an ordinary differential
equation system model
\begin{equation}
\dot{u}=\Theta(u)W\label{eq:general model}
\end{equation}
with row vector of the $d$ solution states $u(t;W):=[\begin{array}{c|c|c|c}
u_{1}(t;W) & u_{2}(t;W) & \cdots & u_{d}(t;W)]\end{array}$, row vector of $J$ features (i.e., right side terms) $\Theta(u):=[\begin{array}{c|c|c|c}
f_{1}(u) & f_{2}(u) & \cdots & f_{J}(u)]\end{array}$ where $f_j:\mathbb{R}^d\to\mathbb{R}$, and the matrix of unknown parameters $W\in\mathbb{R}^{J\times d}$. We consider
a $C^{\infty}$ test function $\phi$ compactly supported in the time interval $[0,T]$ (e.g.\ $\phi \in C_{c}^{\infty}([0,T])$),
multiply both sides of (\ref{eq:general model}) by $\phi$, and integrate
over $0$ to $T$.  Via integration by parts we obtain
\[
\phi(T)u(T)-\phi(0)u(0) -
\int_{0}^{T}\dot{\phi}u\textsf{d}t
=\int_{0}^{T}\phi\Theta(u)W\textsf{d}t.
\]
As the compact support of $\phi$ implies that $\phi(0)=\phi(T)=0$, this yields
a transform of (\ref{eq:general model}) into 
\begin{equation}
-\int_{0}^{T}\dot{\phi}u\textsf{d}t
=\int_{0}^{T}\phi\Theta(u)W\textsf{d}t.\label{eq:WENDyContinuum}
\end{equation}
This weak form of the equation allows us to define a novel methodology for estimating the entries
in $W$.

Observations of states of this system are (in this paper) assumed to occur at a discrete set of
$M+1$ timepoints $\{t_{m}\}_{m=0}^{M}$ with uniform stepsize $\Delta t$.
The test functions are thus  centered at a subsequence of $K$
timepoints $\{t_{m_{k}}\}_{k=1}^{K}$. We choose the test function
support to be centered at a timepoint $t_{m_{k}}$ with radius $m_{t}\Delta t$
where $m_{t}$ is an integer (to be chosen later). Bold variables denote
evaluation at or dependence on the chosen timepoints, e.g., 
\begin{equation*}
\begin{array}{ccc}
\tbf:=\left[\begin{array}{c}
     t_0\\
     \vdots\\
     t_M\end{array}\right],\phantom{\quad} & \ubf:=\left[\begin{array}{ccc}
         u_1(t_0) & \cdots & u_d(t_0) \\
         \vdots & \ddots & \vdots \\
         u_1(t_M) & \cdots & u_d(t_M)
     \end{array}\right],\phantom{\quad} & \Theta(\ubf):=\left[\begin{array}{ccc}
        f_1(u(t_0)) & \cdots & f_J(u(t_0))\\
        \vdots & \ddots & \vdots\\
        f_1(u(t_M)) & \cdots & f_J(u(t_M))
        \end{array}\right].
\end{array}
\end{equation*}
Approximating the integrals in (\ref{eq:WENDyContinuum}) using a
Newton-Cotes quadrature yields 
\begin{equation}
-\dot{\Phi}_{k}\mathbf{u}\approx\Phi_{k}\Theta(\mathbf{u})W,\label{eq:ApproxWENDy}
\end{equation}
where
\[
\begin{array}{ccc}
    \Phi_k:=\left[\begin{array}{c|c|c}
        \phi_k(t_0) & \cdots & \phi_k(t_M)
        \end{array}\right]\bm{\mathcal{Q}},&\phantom{\qquad}& \dot{\Phi}_k:=\left[\begin{array}{c|c|c}
        \dot{\phi}_k(t_0) & \cdots & \dot{\phi}_k(t_M)
        \end{array}\right]\bm{\mathcal{Q}}
\end{array}
\]
and $\phi_{k}$ is a test function centered at timepoint $t_{m_{k}}$. To account for proper scaling, in computations we normalize each test function $\phi_k$ to have unit $\ell_2$-norm, or $\sum_{m=0}^M\phi_k^2(t_m) = 1$.

The $\bm{\mathcal{Q}}$ matrix contains the quadrature weights on the diagonal. In this work we use the composite Trapezoidal  
rule\footnote{The composite Trapezoidal rule works best for the uniform spacing and thus the left and right sides of \eqref{eq:ApproxWENDy} are sums weighted by $\dot{\phi}_{k}(\tbf)$ and $\phi_{k}(\tbf)$, respectively.} for which the matrix is
\[
\bm{\mathcal{Q}}:=\mathsf{diag}(\nicefrac{\Delta t}{2},\Delta t,\ldots,\Delta t,\nicefrac{\Delta t}{2})\in \mathbb{R}^{(M+1)\times(M+1)}.
\]
We defer full consideration of the integration error until Section \ref{sec:TestFunctionminrad} but note that in the case of a non-uniform timegrid, $\bm{\mathcal{Q}}$ would  simply be adapted with the correct stepsize and quadrature weights.

The core idea of the weak-form-based direct parameter estimation is to identify $W$ as a least squares solution
to 
\begin{equation}
\min_{W}\left\Vert \textsf{vec}(\mathbf{G}W-\mathbf{B})\right\Vert _{2}^{2}\label{eq:WENDy}
\end{equation}
 where ``$\textsf{vec}$'' vectorizes a matrix,
\[
\begin{array}{rl}
\mathbf{G} & :=\Phi\Theta(\mathbf{U})\in\mathbb{R}^{K\times J},\\
\mathbf{B} & :=-\dot{\Phi}\mathbf{U}\in\mathbb{R}^{K\times d},
\end{array}
\]
where $\Ubf$ represents the data, and the integration matrices are
\[
\begin{array}{rl}
\Phi=\left[\begin{array}{c}
\Phi_{1}\\
\vdots\\
\Phi_{K}
\end{array}\right]\in\mathbb{R}^{K\times (M+1)}\quad\textsf{and} & \dot{\Phi}=\left[\begin{array}{c}
\dot{\Phi}_{1}\\
\vdots\\
\dot{\Phi}_{K}
\end{array}\right]\in\mathbb{R}^{K\times (M+1)}.\end{array}
\]

The ordinary least squares (OLS) solution to (\ref{eq:WENDy}) is presented in Algorithm \ref{alg:WENDy-with-naive}.  We note that we have written the algorithm this way to promote clarity concerning the weak-form estimation idea.  For actual implementation, we create a different $\Theta_i$ for each variable $i=1\ldots,d$ and use regression for state $i$ to solve for a vector $\widehat{\wbf}_i$ of parameters (instead of a matrix of parameters $W$, which can contain values known to be zero). To increase computational efficiency, we make sure to remove any redundancies and use sparse computations whenever possible.
\begin{algorithm}
\caption{\label{alg:WENDy-with-naive}Weak-form Parameter Estimation with Ordinary Least Squares}

	\SetKwInOut{Input}{input} \SetKwInOut{Output}{output} 
	\Input{Data $\{\mathbf{U}\}$, Feature Map $\{\Theta\}$, Test Function Matrices $\{\Phi,\dot{\Phi}\}$}
	\Output{Parameter Estimate $\{\widehat{W}\}$} 
	\BlankLine
	\BlankLine
	\tcp{Solve Ordinary Least Squares Problem}
	$\Gbf\leftarrow \Phi\Theta(\Ubf)$\\
	$\Bbf\leftarrow -\dot{\Phi}\Ubf$\\
	$\widehat{W}\leftarrow (\Gbf^T\Gbf)^{-1}\Gbf^T\Bbf$
\end{algorithm}

The OLS solution has respectable performance in some cases, but in general there is a clear need for improvement upon OLS. In particular, we note that \eqref{eq:WENDy}
is \emph{not} a standard least squares problem. The (likely noisy)
observations of the state $u$ appear on both sides of
(\ref{eq:ApproxWENDy}). In Statistics, this is known as an Errors
in Variables (EiV) problem.\footnote{
The EiV problem with standard additive i.i.d.\,Gaussian measurement errors is known as a Total Least Squares (TLS) problem in applied and computational mathematics. The literature of EiV and TLS is very similar, but TLS problems are a subset of EiV problems. We direct the interested reader to  \cite{VanHuffelLemmerling2002} for more information.} While a full and rigorous analysis of the statistical properties
of weak-form estimation is beyond the scope of this article\footnote{See our work in \cite{MessengerBortz2022arXiv221116000} for an investigation of the
asymptotic consistency in the limit of continuum data.}, here we will present several formal derivations aimed at improving the accuracy of parameter estimation. These improvements are critical as the OLS approach is not reliably accurate.  Accordingly, we define WENDy (in the next section) as a weak-form parameter estimation method which uses techniques that address the EiV challenges.

\subsection{WENDy: Weak-form estimation using Iterative Reweighting\label{sec:WENDy_IRLS}}

In this subsection, we acknowledge that the 
regression problem does not fit within the framework of ordinary least squares (see Figure \ref{HR_res}) and is actually an Errors-In-Variables problem. We now derive a linearization that yields insight into the  covariance
structure of the problem. First, we denote the vector
of true (but unknown) parameter values used in all state variable equations as $\wbf^{\star}$ and let $u^{\star}:=u(t;\wbf^{\star})$
and $\Theta^{\star}:=\Theta(u^{\star})$. We
also assume that measurements of the system are noisy,
so that at each timepoint $t$ all states are observed with additive
noise
\begin{equation}
U(t)=u^{\star}(t)+\varepsilon(t)\label{eq:additive noise-1}
\end{equation}
where each element of $\varepsilon(t)$ is i.i.d.~$\mathcal{N}(0,\sigma^{2})$.\footnote{Naturally, for real data, there could be different variances for different
states as well as more sophisticated measurement error models. We defer such questions
to future work.} Lastly, we note that there are $d$ variables, $J$ feature terms, and $M+1$ timepoints.  In what follows, we present the expansion using Kronecker products
(denoted as $\otimes$). 

We begin by considering the  sampled data $\Ubf:=\ubf^\star+\pmb{\varepsilon}\in\mathbb{R}^{(M+1)\times d}$ and vector of parameters to be identified $\wbf\in\mathbb{R}^{Jd}$.  The use bolded variables to represent evaluation at the timegrid $\tbf$, and use  superscript $\star$ notation to denote quantities based on true (noise-free) parameter or states.  We now consider the residual
\begin{equation}
\rbf(\mathbf{U},\wbf):=\Gbf\wbf-\bbf,\label{eq:WENDyDataResid}
\end{equation}
where we redefine
\begin{align*}
\Gbf & :=[\mathbb{I}_{d}\otimes(\Phi\Theta(\Ubf))],\\
\bbf & :=-\mathsf{vec}(\dot{\Phi}\Ubf).
\end{align*}
We then note that we can decompose the residual into several components
\begin{align*}
\rbf(\Ubf,\wbf)&= 
\Gbf \wbf - \Gbf^\star\wbf+\Gbf^\star\wbf -\Gbf^\star\wstar+\Gbf^\star\wstar- (\bbf^\star+\bbf^{\pmb{\varepsilon}})\\ 
&= \underbrace{(\Gbf-\Gbf^\star)\wbf}_{\begin{array}{c}\ebf_\Theta\end{array}}+\underbrace{\Gbf^\star(\Wbf-\wstar)}_{\begin{array}{c}\rbf_{0}\end{array}}+\underbrace{(\Gbf^\star\wstar-\bbf^\star)}_{\begin{array}{c}\ebf_{\text{int}}\end{array}}-\bbf^{\pmb{\varepsilon}},
\end{align*}
where 
\begin{align*}
\Gbf^\star & :=[\mathbb{I}_{d}\otimes(\Phi\Theta(\ubf^\star))],\\
\bbf & :=\underbrace{-\mathsf{vec}(\dot{\Phi}\ubf^\star)}_{\begin{array}{c}\bbf^\star\end{array}}+\underbrace{-\textsf{vec}(\dot{\Phi}\,\pmb{\varepsilon})}_{\begin{array}{c}\bbf^{\pmb{\varepsilon}}\end{array}}.
\end{align*}
Here, $\rbf_0$ is the residual without measurement noise or integration errors, and $\ebf_{\text{int}}$ is the numerical integration error induced by the quadrature (and will be analyzed in Section \ref{sec:TestFunction}). 

Let us further consider the leftover terms $\ebf_\Theta-\bbf^{\pmb{\varepsilon}}$ and take a Taylor expansion around the data $\Ubf$ 
\begin{equation}
    \begin{array}{rl}
        \ebf_\Theta-\bbf^{\pmb{\varepsilon}} & = (\Gbf-\Gbf^\star)\wbf +\textsf{vec}(\dot{\Phi}\,\pmb{\varepsilon})\\
        & =  \Big[\mathbb{I}_d\otimes \big(\Phi\left(\Theta(\Ubf)-\Theta(\Ubf-\pmb{\varepsilon})\right)\big)\Big]\wbf + \Big[\mathbb{I}_d\otimes \dot{\Phi}\Big]\textsf{vec}(\pmb{\varepsilon})\\
        & = \Lbf_{\wbf}\mathsf{vec}(\pmb{\varepsilon})+\hbf(\Ubf,\wbf,\pmb{\varepsilon})
    \end{array}\label{eqn:eThetabeps}
\end{equation}
where $\hbf(\Ubf,\wbf,\pmb{\varepsilon})$ is a vector-valued function of higher order terms in the measurement errors $\pmb{\varepsilon}$ (including the Hessian as well as higher order derivatives). Note that the $\hbf$ function will generally produce a bias and higher-order dependencies for all system where $\nabla^2 \Theta \neq \mathbf{0}$, but vanishes when $\pmb{\varepsilon}=\mathbf{0}$.  

The first order matrix in the expansion \eqref{eqn:eThetabeps} is
\[
\mathbf{L}_{\wbf} :=[\mathsf{mat}(\wbf)^{T}\otimes\Phi]\nabla\Theta\Pbf+[\mathbb{I}_{d}\otimes\dot{\Phi}],
\]
where ``$\mathsf{mat}$'' is the matricization operation and $\Pbf$ is a permutation matrix such that $\Pbf\textsf{vec}(\boldsymbol{\varepsilon})=\textsf{vec}(\boldsymbol{\varepsilon}^{T})$. The matrix $\nabla \Theta$ contains derivatives of the features
\begin{align*}
\nabla \Theta & :=\left[\begin{array}{ccc}
\nabla f_{1}(\Ubf_{0})\\
 & \ddots\\
 &  & \nabla f_{1}(\Ubf_{M})\\
\hline  & \vdots\\
\hline \nabla f_{J}(\Ubf_{0})\\
 & \ddots\\
 &  & \nabla f_{J}(\Ubf_{M})
\end{array}\right],
\end{align*}
 where 
\[
\nabla f_{j}(\Ubf_{m})=\left[\begin{array}{c|c|c}
\frac{\partial}{\partial u_{1}}f_{j}(\Ubf_{m}) & \cdots & \frac{\partial}{\partial u_{d}}f_{j}(\Ubf_{m})\end{array}\right],
\]
and $\Ubf_{m}$ is the row vector of true solution
states at $t_{m}$.

As mentioned above, we assume that all elements of $\boldsymbol{\varepsilon}$
are i.i.d.\,Gaussian, i.e., $\mathcal{N}(0,\sigma^2)$ and thus to first order the residual
is characterized by
\begin{equation}
\Gbf \wbf-\bbf-(\rbf_{0}+\ebf_{\text{int}})\sim\mathcal{N}(\mathbf{0},\sigma ^2\mathbf{L}_{\wbf}(\mathbf{L}_{\wbf})^{T})\label{eqn:ResidDistw}.
\end{equation}
In the case where $\wbf=\wstar$ and the integration error is negligible, \eqref{eqn:ResidDistw} simplifies to
\begin{equation}
\Gbf \wstar-\bbf\sim\mathcal{N}(\mathbf{0},\sigma ^2\mathbf{L}_{\wstar}(\mathbf{L}_{\wstar})^{T})\label{eqn:ResidDistwstar}.
\end{equation}
We note that in \eqref{eqn:ResidDistwstar} (and in \eqref{eqn:ResidDistw}), the covariance is dependent upon the parameter vector $\wbf$.
In the statistical inference literature, the Iteratively Reweighted
Least Squares (IRLS) \cite{Jorgensen2012EncyclopediaofEnvironmetrics}
method offers a strategy to account for a parameter-dependent covariance by iterating between solving for $\wbf$ and updating
the covariance matrix $\Cbf$. Furthermore, while the normality in \eqref{eqn:ResidDistwstar} is approximate, the weighted least squares estimator has been shown to be consistent under fairly general conditions even without normality \cite{BollerslevWooldridge1992EconomRev}. In Algorithm \ref{alg:WENDy-IRLS} we present WENDy method, updating $\Cbf^{(n)}$ (at the $n$-th iteration step) in lines 7-8 and then the new parameters $\wbf^{(n+1)}$ are computed in line 9 by 
weighted least squares.

\begin{algorithm}
\caption{\label{alg:WENDy-IRLS}WENDy}

	\SetKwInOut{Input}{input} \SetKwInOut{Output}{output} 
	\Input{Data $\{\mathbf{U}\}$, Feature Map $\{\Theta,\nabla\Theta\}$, Test Function Matrices $\{\Phi,\dot{\Phi}\}$, Stopping Criteria $\{\text{SC}\}$, Covariance Relaxation Parameter $\{\alpha\}$, Variance Filter $\{\fbf\}$}
	\Output{Parameter Estimate $\{\widehat{\wbf},\widehat{\Cbf},\widehat{\sigma},\mathbf{S},\texttt{stdx}\}$} 
	\BlankLine
	\BlankLine

        \tcp{Compute weak-form linear system}

	$\mathbf{G} \leftarrow \left[\mathbb{I}_{d}\otimes(\Phi\Theta(\mathbf{U}))\right]$\\
	$\mathbf{b} \leftarrow -\textsf{vec}(\dot{\Phi}\mathbf{U})$\\

	\BlankLine
	\BlankLine

        \tcp{Solve Ordinary Least Squares Problem}
        $\wbf^{(0)} \leftarrow (\Gbf^T\Gbf)^{-1}\Gbf^T\bbf$\\
	\BlankLine
        \BlankLine

	\tcp{Solve Iteratively Reweighted Least Squares Problem}
        $n \leftarrow 0$\\
        \texttt{check} $\leftarrow$ true\\
        
	\While{\texttt{check} is true}{
            $\Lbf^{(n)} \leftarrow [\textsf{mat}(\wbf^{(n)})^{T}\otimes\Phi]\nabla\Theta(\Ubf)\Pbf+[\mathbb{I}_{d}\otimes\dot{\Phi}]$\\
            $\Cbf^{(n)} = (1-\alpha)\Lbf^{(n)}(\Lbf^{(n)})^T + \alpha \Ibf$\\
		$\wbf^{(n+1)} \leftarrow (\Gbf^T(\Cbf^{(n)})^{-1}\Gbf)^{-1}\Gbf^{T}(\Cbf^{(n)})^{-1}\bbf$\\
            \texttt{check} $\leftarrow \text{SC}(\wbf^{(n+1)},\wbf^{(n)})$ \\ 
		$n \leftarrow n+1$
	}
\BlankLine
\BlankLine
 \tcp{Return estimate and standard statistical quantities}
    $\widehat{\wbf} \leftarrow \wbf^{(n)}$\\
    $\widehat{\Cbf} \leftarrow \Cbf^{(n)}$\\
    $\widehat{\sigma} \leftarrow (Md)^{-1/2}\nrm{\fbf*\Ubf}_\text{F}$ \\
    $\mathbf{S} \leftarrow \widehat{\sigma}^2 ((\Gbf^T\Gbf)^{-1}\Gbf^T)\ 
    \widehat{\Cbf}\ (\Gbf(\Gbf^T\Gbf)^{-1}))$\\
    $\texttt{stdx} \leftarrow \sqrt{\texttt{diag}(\mathbf{S})}$
\end{algorithm}

The IRLS step in line 9 requires inverting $\Cbf^{(n)}$, which is done by computing its Cholesky factorization and then applying the inverse to $\Gbf$ and $\bbf$. Since this inversion may be unstable, we allow for possible regularization of $\Cbf^{(n)}$ in line 8 via a convex combination between the analytical first-order covariance $\Lbf^{(n)}(\Lbf^{(n)})^T$ and the identity via the covariance relaxation parameter $\alpha$. This regularization allows the user to interpolate between the OLS solution ($\alpha=1$) and the unregularized IRLS solution ($\alpha=0$). In this way WENDy extends and encapsulates Algorithm 1. However, in the numerical examples below, we simply set $\alpha=10^{-10}$ throughout, as the aforementioned instability was not an issue. Lastly, any iterative scheme needs a stopping criteria and we will defer discussion of ours until Section \ref{sec:SC}.

The outputs of Algorithm \ref{alg:WENDy-IRLS} include the estimated parameters $\what$ as well as the covariance $\widehat{\Cbf}$ of the response vector $\bbf$ such that approximately 
\[\bbf \sim \CalN(\Gbf\what,\sigma^2\widehat{\Cbf}).\]
A primary benefit of the WENDy methodology is that the parameter covariance matrix $\Sbf$ can be estimated from $\widehat{\Cbf}$ using
\begin{equation}
\Sbf := \widehat{\sigma}^2 ((\Gbf^T\Gbf)^{-1}\Gbf^T)\ 
    \widehat{\Cbf}\ (\Gbf(\Gbf^T\Gbf)^{-1})).
\end{equation}
This yields the variances of individual components of $\what$ along $\textsf{diag}(\Sbf)$ as well as the correlations between elements of $\what$ in the off-diagonals of $\Sbf$. Here $\widehat{\sigma}^2$ is an estimate of the measurement variance $\sigma^2$, which we compute by convolving each compartment of the data $\Ubf$ with a high-order\footnote{The order of a filter is defined as the number of moments that the filter leaves zero (other than the zero-th moment). For more mathematical details see \cite{MessengerBortz2022arXiv221116000} Appendix F.} 
 filter $\fbf$ and taking the Frobenius norm of the resulting convolved data matrix $\fbf *\Ubf$. Throughout we set $\fbf$ to be the centered finite difference weights of order 6 over 15 equally-spaced points (computed using \cite{Fornberg1988MathComput}), so that $\fbf$ has order 5. The filter $\fbf$ is then normalized to have unit 2-norm. This yields a high-accuracy approximation of $\sigma^2$ for underlying data $\ubf^\star$ that is locally well-approximated by polynomials up to degree 5.

\subsection{Choice of Test Functions\label{sec:TestFunction}}

When using WENDy for parameter estimation, a valid question concerns
the choice of test function. This is particularly challenging in the sparse data regime, where integration errors can easily affect parameter estimates. In 
\cite{MessengerBortz2021MultiscaleModelSimul} we reported that using higher order polynomials as test functions yielded more accuracy (up to machine precision). Inspired by this result and to render moot the question of what order polynomial is needed, we have developed a 2-step process for offline computation of highly efficient test functions, given a timegrid $\tbf$.

We first derive an estimator of the integration error that can be computed using the noisy data $\Ubf$ and used to detect a minimal radius $\underline{m}_{t}$ such that $m_t>\underline{m}_{t}$ leads to negligible integration error compared to the errors introduced by random noise. Inspired by wavelet decompositions, we next row-concatenate convolution matrices of test functions at different radii $\mbf_t:= (2^\ell \underline{m}_{t};\ \ell=\{0,\dots,\bar{\ell}\}).$  An SVD of this tall matrix yields an orthonormal test function matrix $\Phi$, which maximally extracts information across different scales. We note that in the later examples we have $\bar{\ell} = 3$, which in many cases leads to a largest test function support covering half of the time domain.

To begin, we consider a $C^\infty$ bump function
\begin{equation}
\psi(t;a) = C\exp\left(-\frac{\eta}{[1-(t/a)^2]_+}\right),\label{eq:CinftyBump}
\end{equation}
where the constant $C$ enforces that $\nrm{\psi}_2=1$, $\eta$ is a shape parameter, and $[\boldsymbol{\cdot}]_+ := \max(\boldsymbol{\cdot},0)$, so that $\psi(t;a)$ is supported only on $[-a,a]$ where
\begin{equation}\label{raddef}
a = m_t\Delta t.
\end{equation}

With the $\psi$ in \eqref{eq:CinftyBump} we have discovered that the accuracy of the parameter estimates is relatively insensitive to a wide range of $\eta$ values.  Therefore, based on empirical investigation we arbitrarily choose $\eta=9$ in all examples and defer more extensive analysis to future work. In the rest of this section, we will describe the computation of $\underline{m}_t$ and how to use $\psi$ to construct $\Phi$ and $\dot{\Phi}$.

\subsubsection{Minimum radius selection\label{sec:TestFunctionminrad}}
\begin{figure}
\begin{center}
		\includegraphics[trim={0 0 0 0},clip,width=0.7\textwidth]{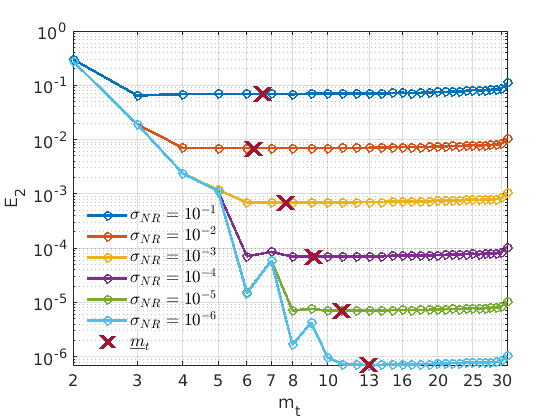}
\end{center}
\caption{Coefficient error $E_2= \|\wstar-\what\|_2/\|\wstar\|_2$ of WENDy applied to the Logistic Growth model vs test function radius $m_t$ for noise levels $\sigma_{NR}\in\{10^{-6},\dots,10^{-1}\}$. For large enough radius, errors are dominated by noise and integration error is negligible. The minimum radius $\underline{m}_t$ computed as in Section \ref{sec:TestFunctionminrad} finds this noise-dominated region, which varies depending on $\sigma_{NR}$.}
\label{noise_err_tradeoff}
\end{figure}
In \eqref{eqn:eThetabeps}, it is clear that reducing the numerical integration errors $\ebf_{\text{int}}$ will improve the estimate accuracy.  Figure \ref{noise_err_tradeoff} illustrates for the Logistic Growth model how the relative error changes as a function of test function radius $m_t$ (for different noise levels). As the radius increases, the error becomee dominated by the measurement noise.  To establish a lower bound $\underline{m}_t$ on the test function radius $m_t$, we create an estimate for the integration error which works for any of the $d$ variables in a model. To promote clarity, we will let $u$ be any of the $d$ variables for the remainder of this section.  However, it is important to note the the final $\widehat{\ebf}_\text{rms}$ sums over all $d$ variables.

We now consider the $k$-th element of $\ebf_\text{int}$
\[\ebf_\text{int}(u^\star,\phi_k,M)_k = (\Gbf^\star\wstar-\bbf^\star)_k  = \sum_{m=0}^{M-1}\left(\phi_k(t_m)\dot{\ubf}_m^\star + \dot{\phi}_k(t_m)\ubf_m^\star\right)\Delta t = \frac{T}{M} \sum_{m=0}^{M-1}\frac{d}{dt}(\phi_k(t_m) \ubf^\star_m),\]
where $\Delta t =T/M$ for a uniform timegrid $\tbf=(0,\Delta t, 2\Delta t,\ldots,M\Delta t)$ with overall length $T$. We also note that the biggest benefit of this approach is that $\ebf_\text{int}$ does not explicitly depend upon $\wstar$.

By expanding $\frac{d}{dt}(\phi_k(t)u^\star(t))$ into its Fourier Series\footnote{We define the $n$th Fourier mode of a function $f:[0,T]\to \Cbb$ as $\CalF_{n}[f] := \frac{1}{\sqrt{T}}\int_0^T f(t)e^{-2\pi in/T}dt$.} we then have
\begin{equation}\label{FFTerrorform}
\ebf_\text{int}(u^\star,\phi_k,M)=\frac{T}{M\sqrt{T}} \sum_{n\in \Zbb} \CalF_n\left[\frac{d}{dt}(\phi_k(t) u^\star(t))\right] \left( \sum_{m=0}^{M-1}e^{2\pi inm/M}\right) =\frac{2\pi i}{\sqrt{T}}\sum_{n\in \Zbb}nM \CalF_{nM}[\phi_k u^\star],
\end{equation}
so that the integration error is entirely represented by aliased modes $\{M,2M,\dots\}$ of $\phi_k u^\star$. Assuming $[-a+t_k,a+t_k]\subset [0,T]$ and $T>2a>1$, we have the relation
\[\CalF_n[\phi_k(\boldsymbol{\cdot};a)] = a\CalF_{na}[\phi_k(\boldsymbol{\cdot};1)],\]
hence increasing $a$ corresponds to higher-order Fourier coefficients of $\phi_k(\boldsymbol{\cdot}; 1)$ entering the error formula \eqref{FFTerrorform}, which shows, using \eqref{FFTerrorform}, that increasing $a$ (eventually) lowers the integration error. For small $m_t$, this leads to the integration error $\ebf_\text{int}$ dominating the noise-related errors, while for large $m_t$, $\ebf_\text{int}$ noise-related effects are dominant. 

\begin{figure}
\begin{tabular}{cc}
		\includegraphics[trim={0 0 35 20},clip,width=0.48\textwidth]{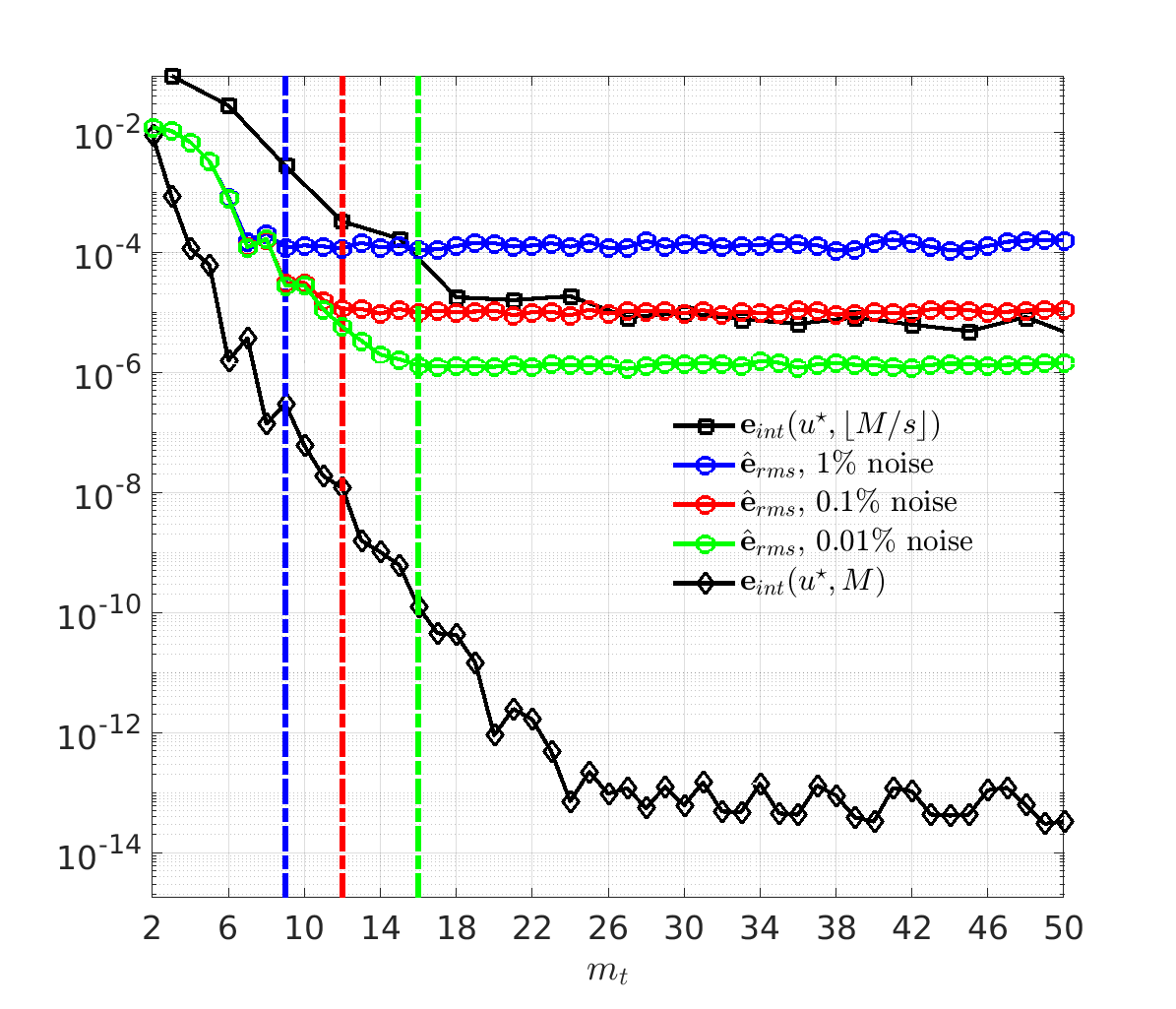} & 
		\includegraphics[trim={0 0 35 20},clip,width=0.48\textwidth]{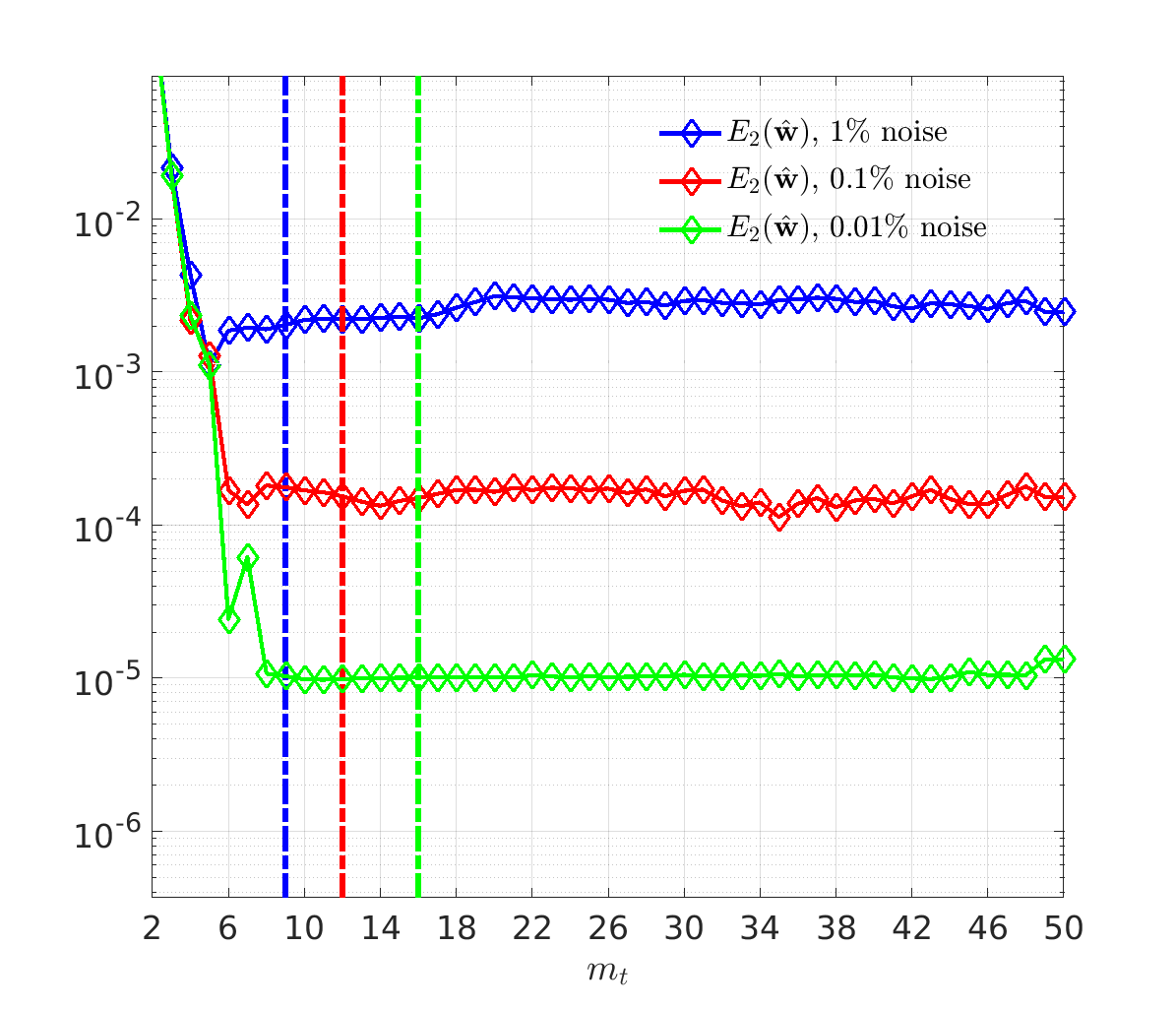}
\end{tabular}
\caption{Visualization of the minimum radius selection using single realizations of Fitzhugh-Nagumo data with 512 timepoints at three different noise levels. Dashed lines indicate the minimum radius $\underline{m}_t$ Left: we see that inequality \eqref{interr_err_ineq} holds empirically for small radii $m_t$. Right: coefficient error $E_2$ as a function of $m_t$ is plotted, showing that for each noise level the identified radius $m_t$ using $\hat{\ebf}_\text{rms}$ lies to right of the dip in $E_2$, as random errors begin to dominate integration errors. In particular, for low levels of noise, $\underline{m}_t$ increases to ensure high accuracy integration.}
\label{interrfig}
\end{figure}

We now derive a surrogate approximation of $\ebf_\text{int}$ using the noisy data $\Ubf$ to estimate this transition from integration error-dominated to noise error-dominated residuals.
 From the noisy data $\Ubf$ on timegrid $\tbf\in\mathbb{R}^M$, we wish to compute $\ebf_\text{int}(u^\star,\phi_k,M)$ by substituting $\Ubf$ for $u^\star$ and using the discrete Fourier transform (DFT), however the highest mode\footnote{We define the $n$th discrete Fourier mode of a function $f$ over a periodic grid $(m\Delta t)_{m=0}^M$ by\newline $\widehat{\CalF}_n[f] := \frac{\Delta t}{\sqrt{M\Delta t}}\sum_{m=0}^{M-1} f(m\Delta t)e^{-2\pi i n m/M}$.} we have access to is $\widehat{\CalF}_{\pm M/2}[\phi \Ubf]$. On the other hand, we \textit{are} able to approximate $\ebf_\text{int}(u^\star,\phi_k,\lfloor M/s\rfloor)$ from $\Ubf$, that is, the integration error over a {\it coarsened} timegrid $(0,\tilde{\Delta t},2\tilde{\Delta t}, \dots, \lfloor M/s\rfloor \tilde{\Delta t})$, where $\tilde{\Delta t} = T / \lfloor M/s\rfloor$ and $s>2$ is a chosen coarsening factor. By introducing the truncated error formula 
\[ \widehat{\ebf}_\text{int}(u^\star,\phi_k,\lfloor M/s\rfloor,s) := \frac{2\pi i}{\sqrt{T}}\sum_{n=-\flr{s/2}}^{\flr{s/2}}n\lfloor M/s\rfloor \CalF_{n\lfloor M/s\rfloor}[\phi_k u^\star],\]
we have that 
\[\widehat{\ebf}_\text{int}(u^\star,\phi_k,\lfloor M/s\rfloor,s)\approx \ebf_\text{int}(u^\star,\phi_k,\lfloor M/s\rfloor),\]
and $\widehat{\ebf}_\text{int}$ can be directly evaluated at $\Ubf$ using the DFT. In particular, with $2<s<4$, we get
\[\widehat{\ebf}_\text{int}(\Ubf,\phi_k,\lfloor M/s\rfloor,s) = \frac{2\pi i \flr{M/s}}{\sqrt{T}}\left(\widehat{\CalF}_{\lfloor M/s\rfloor}[\phi_k \Ubf]-\widehat{\CalF}_{-\lfloor M/s\rfloor}[\phi_k \Ubf]\right) = -\frac{4\pi\flr{M/s}}{\sqrt{T}}\text{Im}\{\widehat{\CalF}_\flr{M/s}[\phi_k \Ubf]\}\]
where $\text{Im}\{z\}$ denotes the imaginary portion of $z\in \Cbb$, so that only a single Fourier mode needs computation. In most practical cases of interest, this leads to (see Figure \ref{interrfig}) 
\begin{equation}\label{interr_err_ineq}
\ebf_\text{int}(u^\star,\phi_k,M) \ \leq \ \widehat{\ebf}_\text{int}(\Ubf,\phi_k,\lfloor M/s\rfloor,s) \ \leq \ \ebf_\text{int}(u^\star,\phi_k,\lfloor M/s\rfloor) 
\end{equation}
so that ensuring $\widehat{\ebf}_\text{int}(\Ubf,\phi_k,\lfloor M/s\rfloor,s)$ is below some tolerance $\tau$ leads also to $\ebf_\text{int}(u,\phi_k,M)<\tau$.

Statistically, under our additive noise model we have that $\widehat{\ebf}_\text{int}(\Ubf,\phi_k,\lfloor M/s\rfloor,s)$ is an unbiased estimator of $\widehat{\ebf}_\text{int}(u^\star,\phi_k,\lfloor M/s\rfloor,s)$, i.e.,
\[\Ebb[\widehat{\ebf}_\text{int}(\Ubf,\phi_k,\lfloor M/s\rfloor,s)]=\Ebb[-(\nicefrac{4\pi\flr{M/s}}{\sqrt{T}})\text{Im}\{\widehat{\CalF}_\flr{M/s}[\phi_k (\ubf^\star+\pmb{\varepsilon})]\}]=\Ebb[\widehat{\ebf}_\text{int}(u^\star,\phi_k,\lfloor M/s\rfloor,s)],\]
where $\Ebb$ denotes expectation.
The variance satisfies, for $2<s<4$,  
\[\textbf{Var}[\widehat{\ebf}_\text{int}(\Ubf,\phi_k,\lfloor M/s\rfloor,s)] := \sigma^2\left(\frac{4\pi\flr{M/s}}{M}\right)^2\sum_{j=1}^{M-1}\phi^2_k(j\Delta t)\sin^2(2\pi 
     \flr{M/s}j/M)\leq \sigma^2\left(\frac{4\pi\flr{M/s}}{M}\right)^2\]
where $\sigma^2 = \mathbf{Var}[\ep]$. The upper bound follows from $\nrm{\phi_k}_2 = 1$, and shows that the variance is not sensitive to the radius of the test function $\phi_k$. 

We pick a radius $\underline{m}_t$ as a changepoint of $\log(\hat{\ebf}_\text{rms})$,  where $\hat{\ebf}_\text{rms}$ is the root-mean-squared integration error over test functions placed along the timeseries, 
\begin{equation}
    \hat{\ebf}_\text{rms}(m_t):= K^{-1}\sum_{k=1}^K\sum_{i=1}^{d}\widehat{\ebf}_\text{int}(\Ubf^{(i)},\phi_k(\cdot;m_t),\lfloor M/s\rfloor,s)^2,
    \label{eq:IntErr}
\end{equation}
where $\Ubf^{(i)}$ is the $i$th variable in the system.
Figure \ref{interrfig} depicts $\widehat{\ebf}_\text{rms}$ as a function of support radius $m_t$. As can be seen, since the variance of $\widehat{\ebf}_\text{int}$ is insensitive to the radius $m_t$, the estimator is approximately flat over the region with negligible integration error, a perfect setting for changepoint detection. Crucially, Figure \ref{interrfig} demonstrates that, in practice, the minimum radius $\underline{m}_t$ lies to the right of the changepoint of the coefficient errors 
\[E_2(\what) := \nrm{\what-\wstar}_2^2/\nrm{\wstar}_2^2,\]
as a function of $m_t$.  Lastly, note that the red $\times$ in Figure \ref{noise_err_tradeoff} depicts the identified $\underline{m}_{t}$ for the Logistic Growth model.

\subsubsection{Orthonormal test functions}\label{sec:orthotf}

\begin{figure}
\begin{tabular}{@{}c@{}c@{}}
    \includegraphics[trim={40 0 40 10},clip,width=0.48\textwidth]{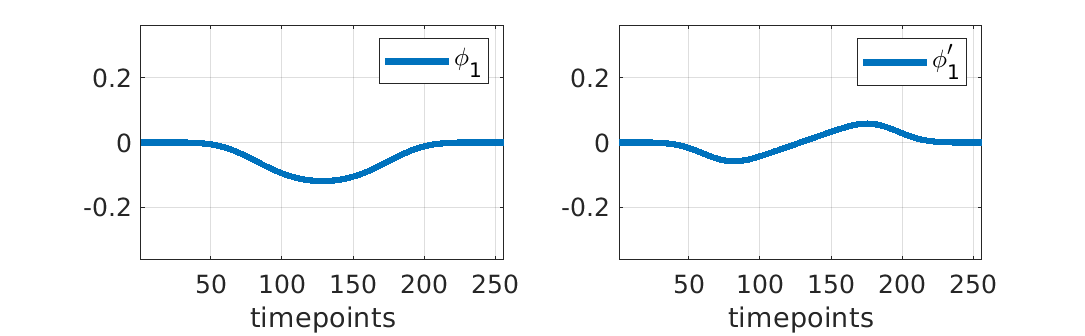} & 
    \includegraphics[trim={40 0 40 10},clip,width=0.48\textwidth]{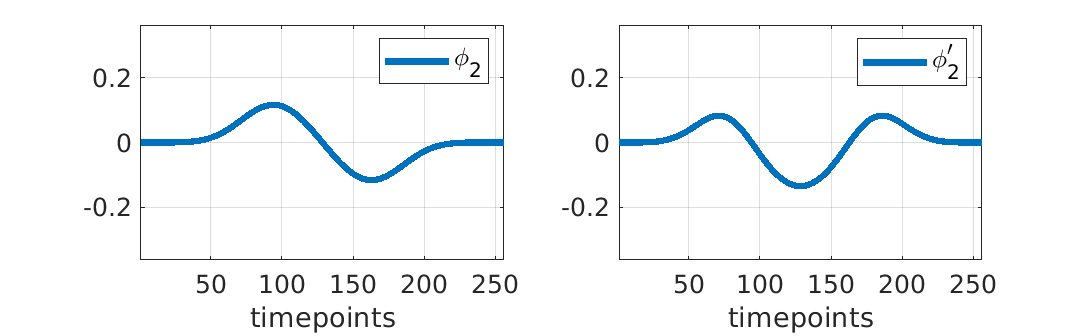} \\ 
    \includegraphics[trim={40 0 40 10},clip,width=0.48\textwidth]{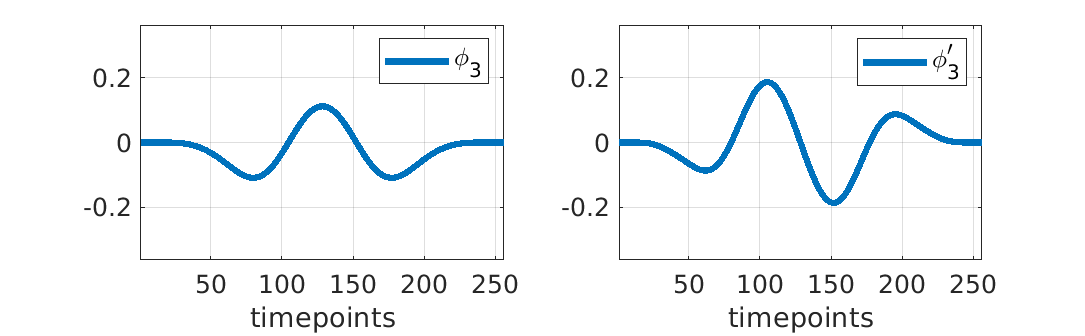} & 
    \includegraphics[trim={40 0 40 10},clip,width=0.48\textwidth]{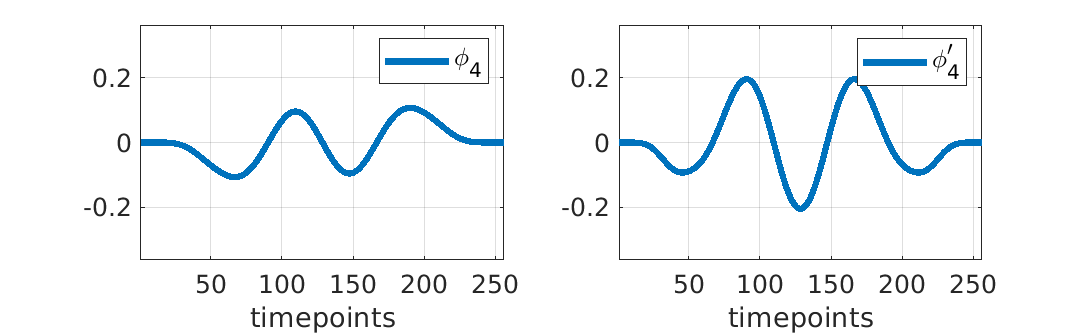} \\ 
    \includegraphics[trim={40 0 40 10},clip,width=0.48\textwidth]{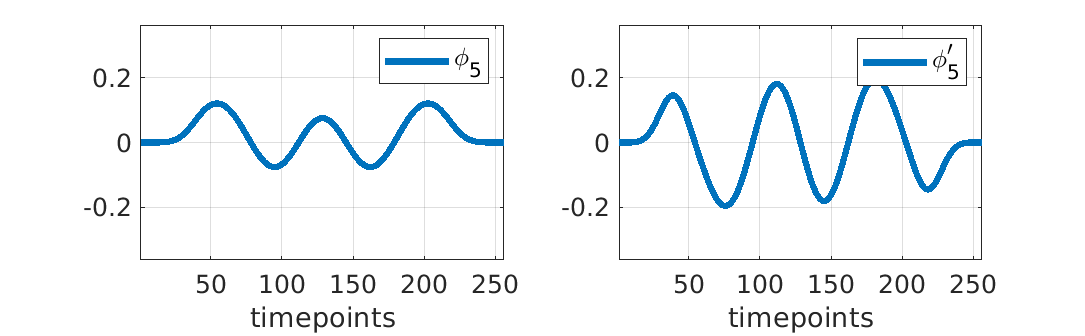} & 
    \includegraphics[trim={40 0 40 10},clip,width=0.48\textwidth]{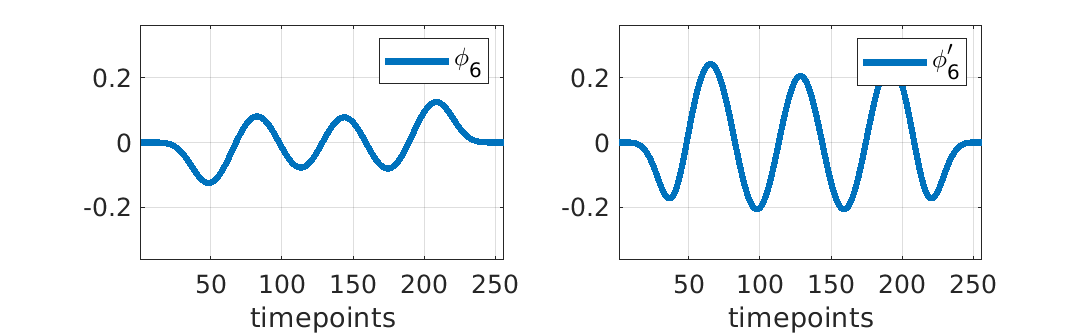} \\ 
\end{tabular}
\caption{First six orthonormal test functions obtained from Hindmarsh-Rose data with 2\% noise and 256 timepoints using the process outlined in Section \ref{sec:orthotf}.}
\label{orthotf}
\end{figure}

Having computed the minimal radius $\underline{m}_t$, we then construct the test function matrices $(\Phi,\dot{\Phi})$ by orthonormalizing and truncating a concatenation of test function matrices with $\mbf_t:= \underline{m}_t\times(1,2,4,8)$. Letting $\Psi_{\ell}$ be the convolution matrix for $\psi(\boldsymbol{\cdot}\ ; 2^\ell \underline{m}_t \Delta t)$, we compute the SVD of
\[\Psi := \begin{bmatrix} \Psi_0 \\ \Psi_1 \\ \Psi_2 \\ \Psi_3 \end{bmatrix}= \Qbf\Sigma\Vbf^T.\]
The right singular vectors $\Vbf$ then form an orthonormal basis for the set of test functions forming the rows of $\Psi$. Letting $r$ be the rank of $\Psi$, we then truncate the SVD to rank $K$, where $K$ is selected as the changepoint in the cumulative sum of the singular values $(\Sigma_{ii})_{i=1}^r$. We then let 
\[\Phi = (\Vbf^{(K)})^T\] 
be the test function basis where $\Vbf^{(K)}$ indicates the first $K$ modes of $\Vbf$. Unlike our previous implementations, the derivative matrix $\dot{\Phi}$ must now be computed numerically, however given the compact support and smoothness of the reference test functions $\psi(\boldsymbol{\cdot} ; 2^\ell \underline{m}_t \Delta t)$, this can be done very accurately with Fourier differentiation. Hence, we let 
\[\dot{\Phi} = \CalF^{-1}\textsf{diag}(i\pmb{k})\CalF\Phi\]
where $\CalF$ is the discrete Fourier transform and $\pmb{k}$ are the requisite wavenumbers. Figure \ref{orthotf} displays the first six orthonormal test functions along with their derivatives obtained from this process applied to Hindmarsh-Rose data.

\subsection{Stopping criteria \label{sec:SC}}

\begin{figure}
\centering
\begin{tabular}{ccc}
    \includegraphics[trim={0 0 35 20},clip,width=0.3\textwidth]{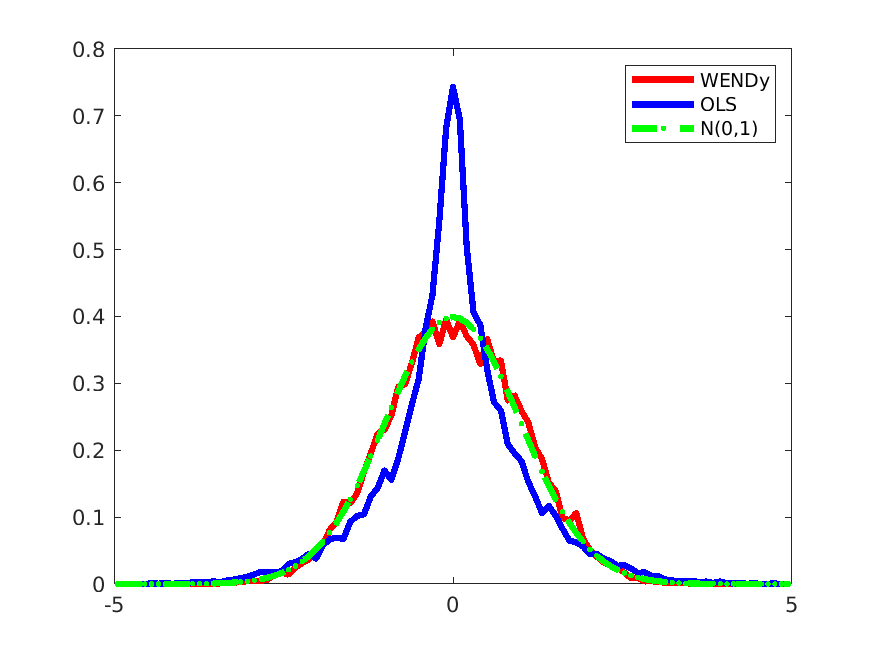}&
    \includegraphics[trim={0 0 35 20},clip,width=0.3\textwidth]{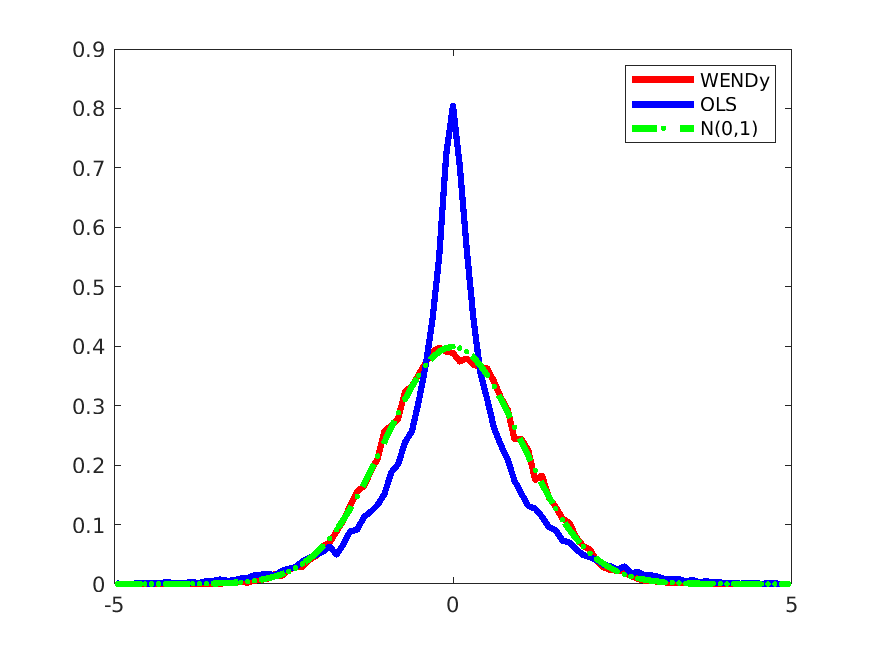}&
    \includegraphics[trim={0 0 35 20},clip,width=0.3\textwidth]{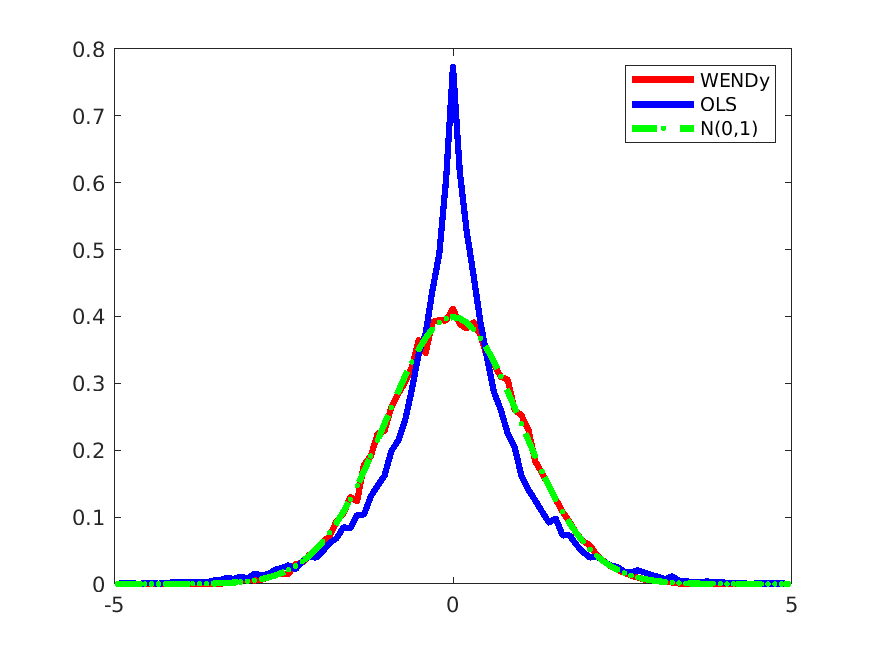}
\end{tabular}
\caption{Histograms of the WENDy (red) and OLS (blue) residuals evaluated at the WENDy output $\what$ applied to the (left-right) Logistic Growth, Lotka-Volterra, and Fitzhugh-Nagumo data, each with 256 timepoints and $20\%$ noise. Curves are averaged over 100 independent trials with each histogram scaled by its empirical standard deviation. In each case, the WENDy residual agrees well with a standard normal, while the OLS residual exhibits distictly non-Gaussian features, indicative that OLS is the wrong statistical regression model.}
\label{HR_res}
\end{figure}

Having formed the test function matrices $\{\Phi,\dot{\Phi}\}$, the remaining unspecified process in Algorithm \ref{alg:WENDy-IRLS} is the stopping criteria $\text{SC}$. The iteration can stop in one of three ways: (1) the iterates reach a fixed point, (2) the number of iterates exceeds a specified limit, or (3) the residuals 
\[\rbf^{(n+1)} := (\Cbf^{(n)})^{-1/2}(\Gbf\wbf^{(n+1)}-\bbf)\]
are no longer approximately normally distributed. (1) and (2) are straightfoward limitations of any iterative algorithm while (3) results from the fact that our weighted 
least-squares framework is only approximate. In ideal scenarios where the discrepancy terms $\ebf_\text{int}$ and $\hbf(\ubf^\star,\wstar;\pmb{\varepsilon})$ are negligible, equation \eqref{eqn:ResidDistw} implies that
\[(\Cbf^\star)^{-1}(\Gbf\wstar-\bbf)\sim \CalN(\pmb{0},\sigma^2\Ibf)\]
where $\Cbf^\star = \Lbf^\star(\Lbf^\star)^T$ is the covariance computed from $\wstar$. Hence we expect $\rbf^{(n)}$ to agree with a normal distribution more strongly as $n$ increases. If the discrepancy terms are non-negligible, it is possible that the reweighting procedure will not result in an increasingly normal $\rbf^{(n)}$, and iterates $\wbf^{(n)}$ may become worse approximations of $\wstar$. A simple way to detect this is with the Shapiro-Wilk (S-W) test for normality \cite{ShapiroWilk1965Biometrika}, which produces an approximate $p$-value under the null hypothesis that the given sample is i.i.d.\ normally distributed. However, the first few iterations are also not expected to yield i.i.d.\ normal residuals (see Figure \ref{HR_res}), so we only check the S-W test after a fixed number of iterations $n_0$. Letting $\text{SW}^{(n)}:=\text{SW}(\rbf^{(n)})$ denote the $p$-value of the S-W test at iteration $n> n_0$, and setting $\text{SW}^{(n_0)}=1$, we specify the stopping criteria as:
\begin{equation}
\text{SC}(\wbf^{(n+1)},\wbf^{(n)}) = \{\|\wbf^{(n+1)}-\wbf^{(n)}\|_2/\|\wbf^{(n)}\|_2>\tau_\text{FP}\}\ \text{and}\ \{n<\texttt{max\_its}\}\ \text{and}\ \{\text{SW}^{(\max\{n,n_0\})}> \tau_\text{SW}\}.
\end{equation}
We set the fixed-point tolerance to $\tau_\text{FP}=10^{-6}$, the S-W tolerance and starting point to $\tau_\text{SW}=10^{-4}$ and $n_0=10$, and $\texttt{max\_its}=100$.

\section{Illustrating Examples\label{sec:Illustrating-Examples}}

Here we demonstrate the effectiveness of WENDy applied to five ordinary differential equations canonical to biology and biochemical modeling. As demonstrated in the works mentioned in Section \ref{sec:Introduction}, it is known that the weak or integral formulations are advantageous, with previous works mostly advocating for a two step process involving (1) pre-smoothing the data before (2) solving for parameters using ordinary least squares. The WENDy approach does not involve smoothing the data, and instead leverages the covariance structure introduced by the weak form to iteratively reduce errors in the ordinary least squares (OLS) weak-form estimation. Utilizing the covariance structure in this way not only reduces error, but reveals parameter uncertainties as demonstrated in Section \ref{sec:uncertainty}.

We compare the WENDy solution to the weak-form ordinary least squares solution (described in Section \ref{sec:WENDy} and denoted simply by OLS in this section) to forward solver-based nonlinear least squares (FSNLS).
Comparison to OLS is important due to the growing use of weak formulations in joint equation learning / parameter estimation tasks, but often without smoothing or further variance reduction steps \cite{MessengerBortz2021JComputPhys,FaselKutzBruntonEtAl2021ArXiv211110992CsMath,NicolaouHuoChenEtAl2023arXiv230102673,BertsimasGurnee2023NonlinearDyn}. In most cases WENDy reduces the OLS error by $60\%$-$90\%$ (see the bar plots in Figures \ref{Logistic_Growth_fig}-\ref{biochemM1_fig}). When compared to FSNLS, WENDy provides a more efficient and accurate solution in typical use cases, however in the regime of highly sparse data and large noise, FSNLS provides an improvement in accuracy at a higher computational cost. Furthermore, we demonstrate that FSNLS may be improved by using the WENDy output as an initial guess. We aim to explore further benefits of combining forward solver-based approaches with solver-free weak-form approaches in a future work. Code to generate all examples is available at \url{https://github.com/MathBioCU/WENDy}.

\subsection{Numerical methods and performance metrics}

In all cases below, we solve for approximate weights $\what$ using Algorithm \ref{alg:WENDy-IRLS} over 100 independent trials of additive Gaussian noise with standard deviation $\sigma = \sigma_{NR}\|\textsf{vec}(\Ubf^\star)\|_\text{rms}$ for a range of noise ratios $\sigma_{NR}$. This specification of the variance implies that 
\[\sigma_{NR} \approx \frac{\|\textsf{vec}(\Ubf^\star-\Ubf)\|_\text{rms}}{\|\textsf{vec}(\Ubf)\|_\text{rms}},\]
so that $\sigma_{NR}$ can be interpreted as the relative error between the true and noisy data. Results from all trials are aggregated by computing the mean and median. Computations of Algorithm \ref{alg:WENDy-IRLS} are performed in MATLAB on a laptop with 40GB of RAM and an 8-core AMD Ryzen 7 pro 4750u processor. Computations of FSNLS are also performed in MATLAB but were run on the University of Colorado Boulder's Blanca Condo Cluster in a trivially parallel manner over a homogeneous CPU set each with Intel Xeon Gold 6130 processors and 24GB RAM. Due to the comparable speed of the two processors (1.7 GHz for AMD Ryzen 7, 2.1 GHz for Intel Xeon Gold) and the fact that each task required less than 5 GB working memory (well below the maximum allowable), we believe the walltime comparisons between WENDy and FSNLS below are fair.

As well as $\sigma_{NR}$, we vary the stepsize $\Delta t$ (keeping the final time $T$ fixed for each example), to demonstrate large and small sample behavior. For each example, a high-fidelity solution is obtained on a fine grid (512 timepoints for Logistic Growth, 1024 for all other examples), which is then subsampled by factors of 2 to obtain coarser datasets.

To evaluate the performance of WENDy, we record the relative coefficient error
\begin{equation}
E_2:= \frac{\|\what-\wstar\|_2}{\|\wstar\|_2}
\end{equation}
as well as the forward simulation error
\begin{equation}
E_\text{FS}:= \frac{\|\textsf{vec}(\Ubf^\star-\widehat{\Ubf})\|_2}{\|\textsf{vec}(\Ubf^\star)\|_2}.
\end{equation}
The data $\widehat{\Ubf}$ is obtained by simulating forward the model using the learned coefficients $\what$ from the exact initial conditions $u(0)$ using the same $\Delta t$ as the data. The RK45 algorithm is used for all forward simulations (unless otherwise specified) with relative and absolute tolerances of $10^{-12}$. Comparison with OLS solutions is displayed in bar graphs which give the drop in error from the OLS solution to the WENDy solution as a percentage of the error in the OLS solution.

\begin{table}
\centering
\begin{tabular}{@{} |p{3cm}|l|p{6cm}|@{} }
\hline
Name & ODE & Parameters \\ 
\hline 
Logistic Growth & \hspace{0.3cm}$\dot{u} = w_1u+w_2u^2$ & $T = 10$, $u(0) = 0.01$, \newline $\|\textsf{vec}(\Ubf^\star)\|_\text{rms} = 0.66$,\newline $\wstar = (1,-1)$ \\
\hline 
Lotka-Volterra & 
$\begin{dcases} \dot{u}_1 = w_1u_1+w_2u_1u_2 \\ \dot{u}_2 = w_3u_2 + w_4u_1u_2 \end{dcases}$ & $T = 5$, $u(0) = (1,1)$, \newline $\|\textsf{vec}(\Ubf^\star)\|_\text{rms} = 6.8$,\newline  $\wstar = (3,-1,-6,1)$  \\
\hline
Fitzhugh-Nagumo & 
$\begin{dcases} \dot{u}_1 = w_1u_1+w_2u_1^3 + w_3u_2 \\ \dot{u}_2 = w_4u_1 + w_5(1) + w_6u_2 \end{dcases}$ & $T = 25$, $u(0) = (0,0.1)$, \newline $\|\textsf{vec}(\Ubf^\star)\|_\text{rms} = 0.68$,\newline $\wstar = (3,-3,3,-1/3,17/150,1/15)$  \\
\hline
Hindmarsh-Rose & $\begin{dcases} \dot{u}_1 = w_1u_2+w_2u_1^3 + w_3u_1^2 + w_4 u_3 \\ \dot{u}_2 = w_5(1) + w_6u_1^2 + w_7u_2 \\ \dot{u}_3 = w_8u_1+w_9(1)+w_{10}u_3\end{dcases}$  & \vspace{-0.8cm}$T = 10$, $u(0) = (-1.31,-7.6,-0.2)$,  \newline $\|\textsf{vec}(\Ubf^\star)\|_\text{rms} = 2.8$,\newline $\wstar = (10,-10,30,-10,10,-50,-10,$ \newline $ 0.04,0.0319,-0.01)$  \\
\hline
Protein Transduction Benchmark (PTB) & $\begin{dcases} \dot{u}_1 = w_1u_1+w_2u_1u_3 + w_3u_4 \\ \dot{u}_2 = w_4u_1 \\ \dot{u}_3 = w_5u_1u_3+w_6u_4+w_7\frac{u_5}{0.3 + u_5} \\ \dot{u}_4 = w_8 u_1u_3 + w_9u_4 \\ \dot{u}_5 = w_{10}u_4 + w_{11}\frac{u_5}{0.3 + u_5}\end{dcases}$ & \vspace{-1cm}$T = 25$, $u(0) = (1,0,1,0,1)$,  \newline $\|\textsf{vec}(\Ubf^\star)\|_\text{rms} = 0.81$,\newline $\wstar = (-0.07,-0.6,0.35,0.07,-0.6,0.05,$ \newline $ 0.17,0.6,-0.35,0.3,-0.017)$ \\
\hline
\end{tabular}
\caption{Specifications of ODE examples. Note that $\|\textsf{vec}(\Ubf^\star)\|_\text{rms}$ is included for reference in order to compute the noise variance using $\sigma = \sigma_{NR}/\|\textsf{vec}(\Ubf^\star)\|_\text{rms}$.}
\end{table}

\subsection{Summary of results}

\subsubsection{Logistic Growth}

\begin{figure}
\centering
\begin{tabular}{@{}c@{}c@{}}
    \includegraphics[trim={-20 0 30 15},clip,width=0.48\textwidth]{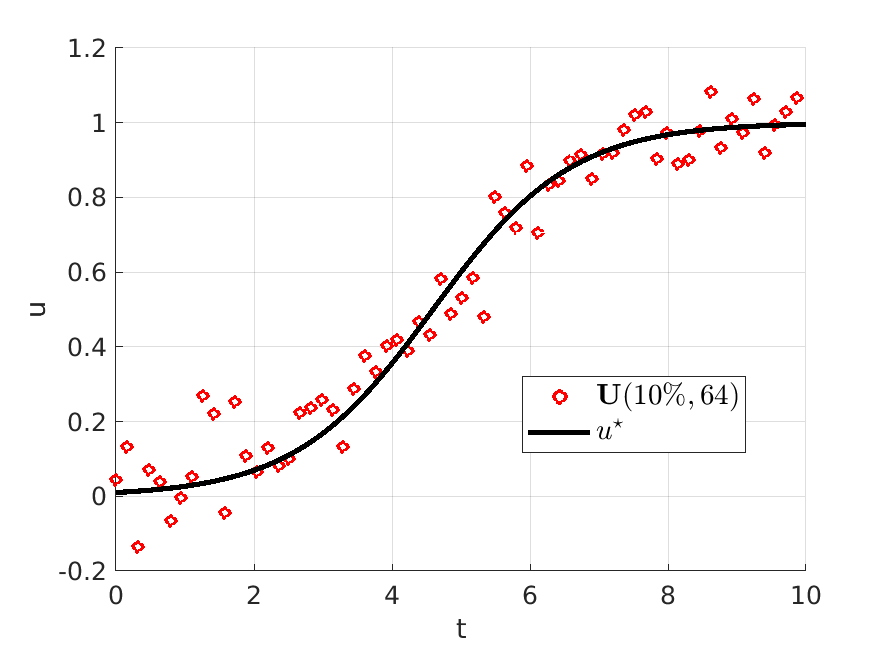} &
    \includegraphics[trim={-20 0 30 20},clip,width=0.48\textwidth]{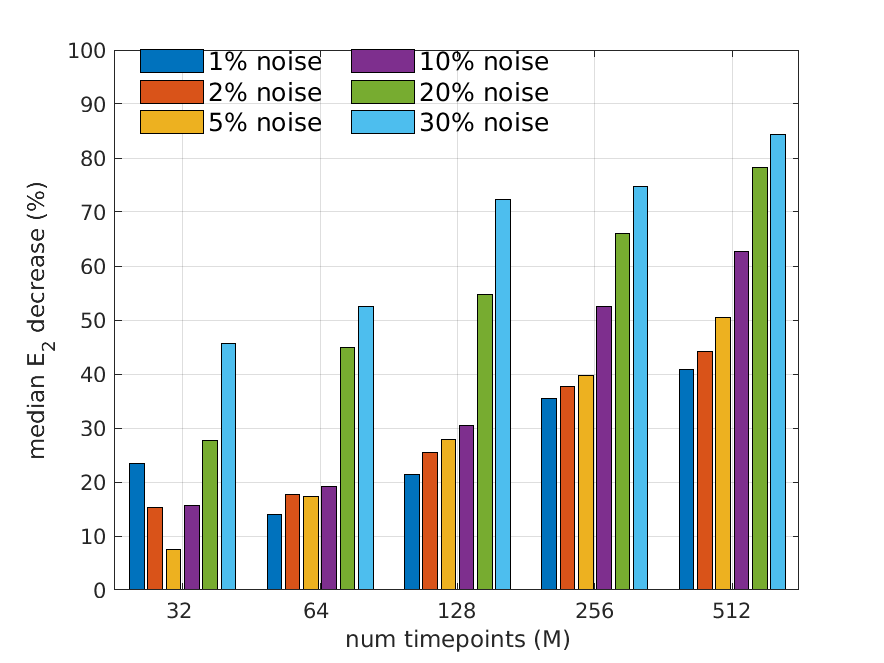} \\
    \includegraphics[trim={0 0 30 20},clip,width=0.48\textwidth]{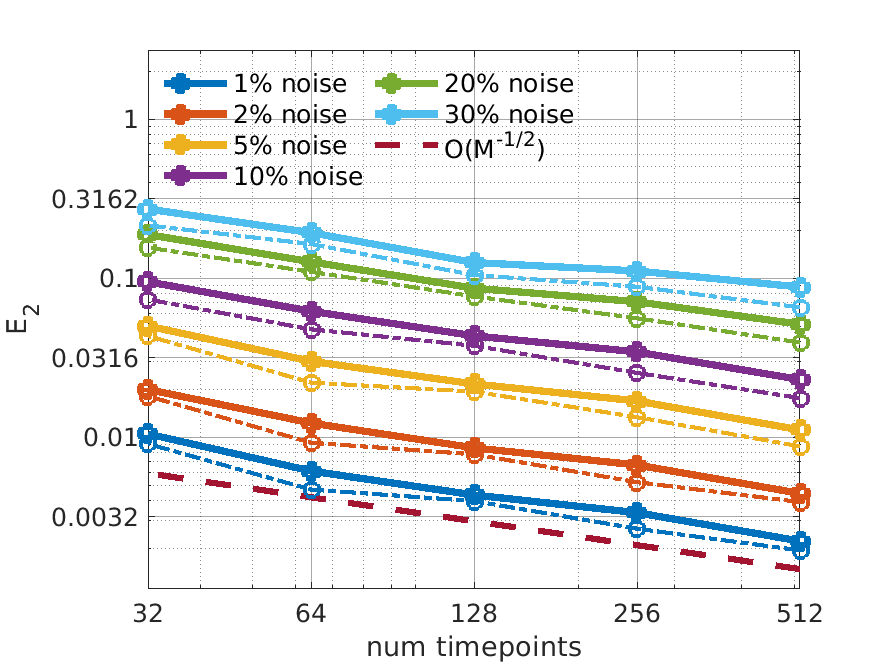}&
    \includegraphics[trim={0 0 30 20},clip,width=0.48\textwidth]{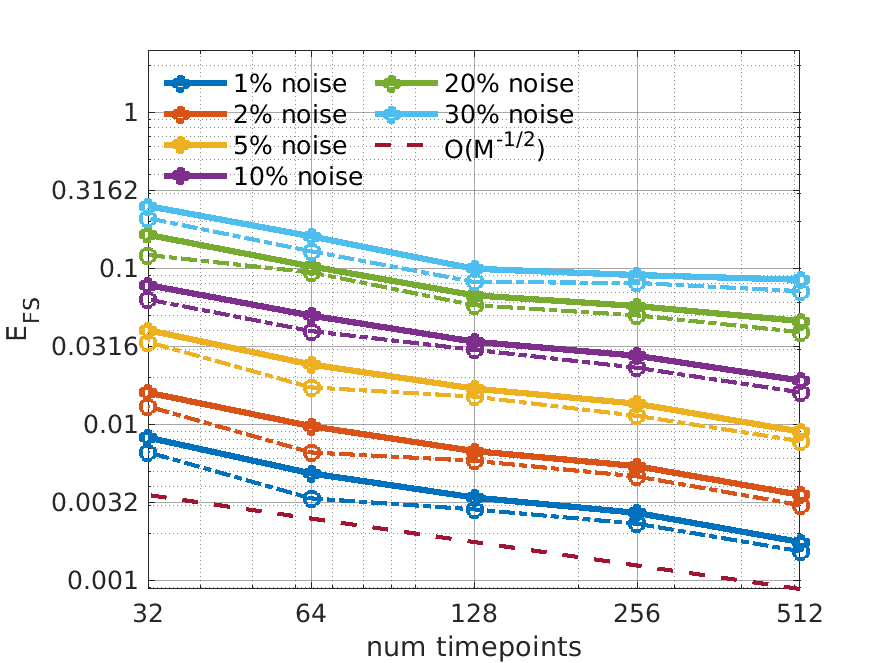}
\end{tabular}
\caption{\textbf{Logistic Growth}: Estimation of parameters in the Logistic Growth model. Left and middle panels display parameter errors $E_2$ and forward simulation error $E_{FS}$, with solid lines showing mean error and dashed lines showing median error. Right: median percentage drop in $E_2$ from the OLS solution to the WENDy output (e.g. at $30\%$ noise and 512 timepoints WENDy results in a 85\% reduction in error).}
\label{Logistic_Growth_fig}
\end{figure}

The logistic growth model is the simplest nonlinear model for population growth, yet the $u^2$ nonlinearity generates a bias that affects the OLS solution more strongly as noise increases. Figure \ref{Logistic_Growth_fig} (top right) indicates that when $M\geq 256$ WENDy decreases the error by 50\%-85\% from the OLS solution for noise level is 10\% or higher. WENDy also leads to a robust fit for smaller $M$, providing coefficient errors $E_2$ and forward simulation errors $E_\text{FS}$ that are both less than $6\%$ for data with only 64 points and $10\%$ noise (Figure \ref{Logistic_Growth_fig} (top left) displays an example dataset at this resolution). 

\subsubsection{Lotka-Volterra}

\begin{figure}
\centering
\begin{tabular}{@{}c@{}c@{}}
    \includegraphics[trim={-20 0 25 15},clip,width=0.48\textwidth]{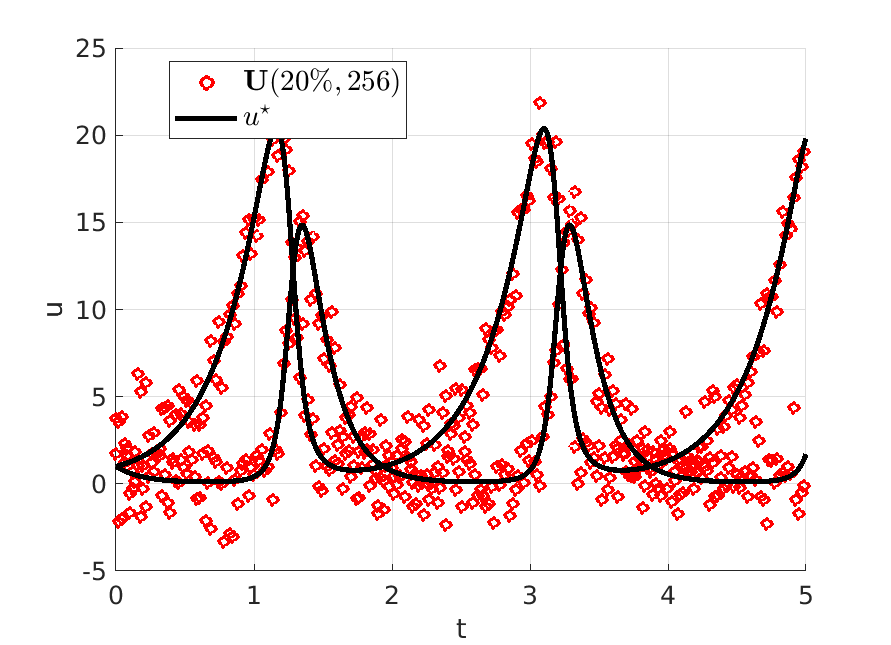} &
    \includegraphics[trim={-20 0 25 20},clip,width=0.48\textwidth]{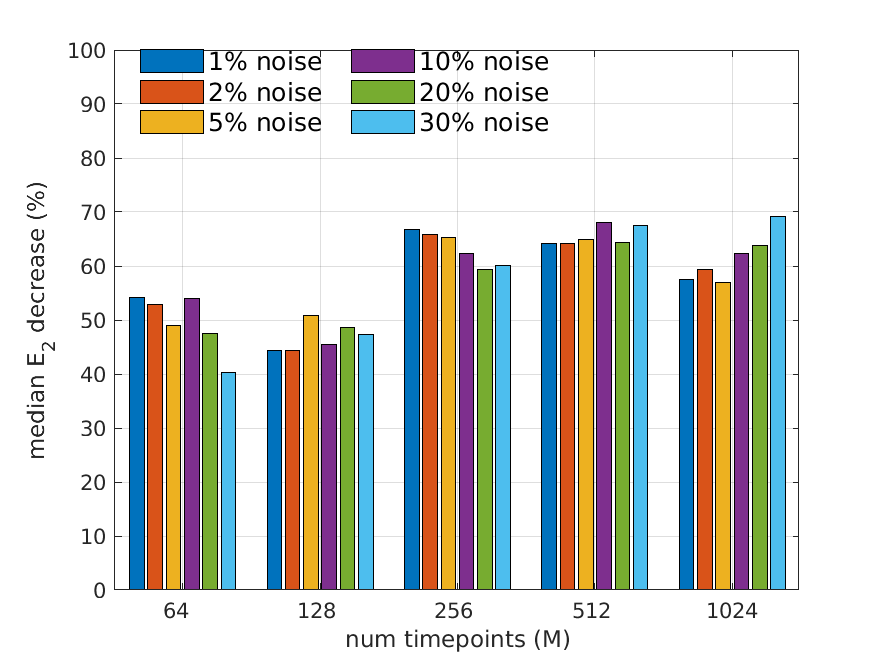} \\
    \includegraphics[trim={0 0 25 20},clip,width=0.48\textwidth]{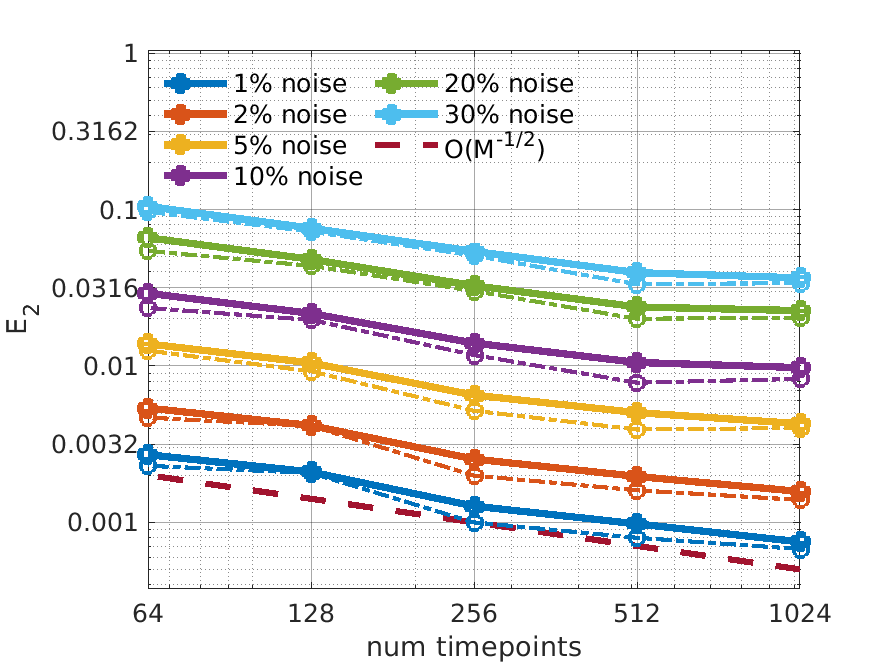}&
    \includegraphics[trim={0 0 25 20},clip,width=0.48\textwidth]{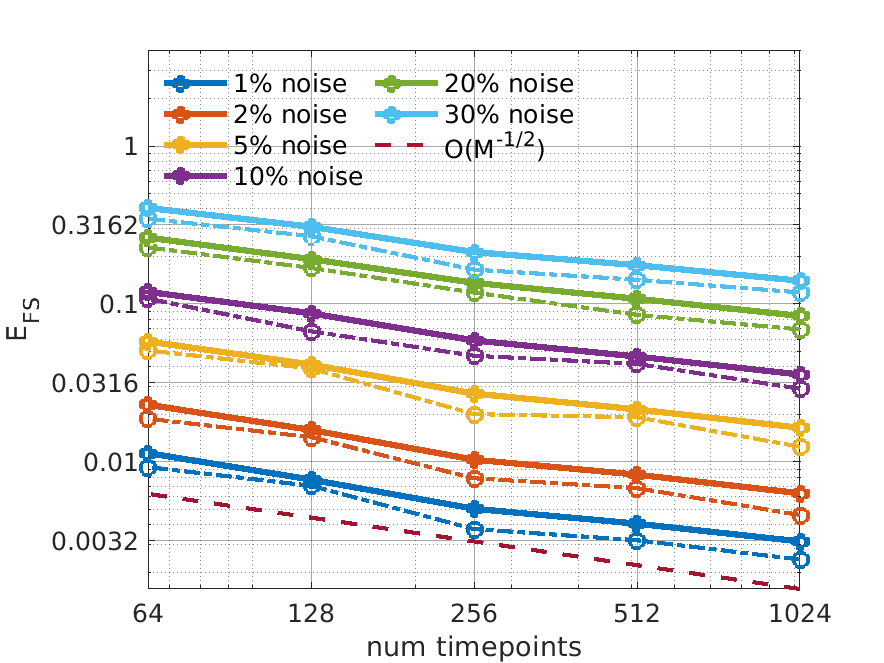}
\end{tabular}
\caption{\textbf{Lotka-Volterra}: Estimation of parameters in the Lotka-Volterra model (for plot details see Figure \ref{Logistic_Growth_fig} caption).}
\label{Lotka_Volterra_fig}
\end{figure}

The Lotka-Volterra model is a system of equations designed to capture predator-prey dynamics \cite{Lotka1978TheGoldenAgeofTheoreticalEcology1923-1940}. Each term in the model is unbiased when evaluated at noisy data (under the i.i.d. assumption), so that the first-order residual expansion utilized in WENDy is highly accurate. The bottom right plot in Figure \ref{Lotka_Volterra_fig} shows even with $30\%$ noise and only 64 timepoints, the coefficient error is still less than $10\%$. WENDy reduces the error by $40\%$-$70\%$ on average from the OLS (top right panel).

\subsubsection{Fitzhugh-Nagumo}

\begin{figure}
\centering
\begin{tabular}{@{}c@{}c@{}}
    \includegraphics[trim={-20 0 30 15},clip,width=0.48\textwidth]{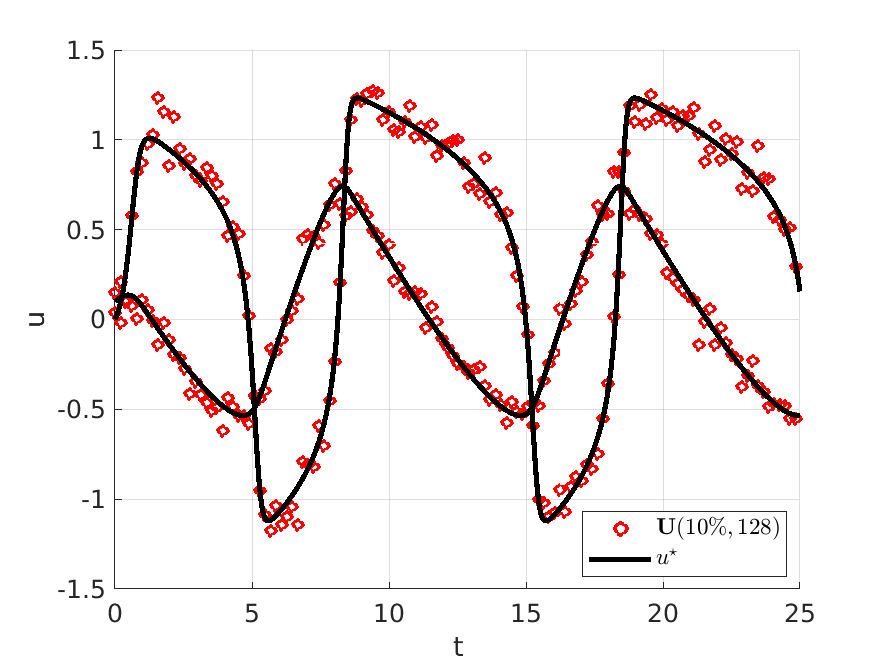} &
    \includegraphics[trim={-20 0 26 20},clip,width=0.48\textwidth]{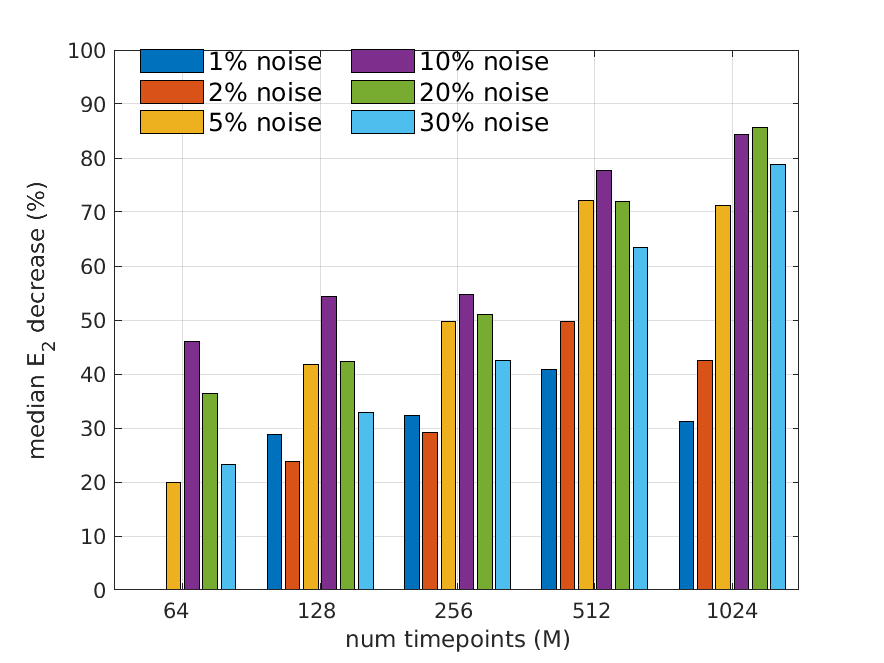} \\
    \includegraphics[trim={0 0 26 20},clip,width=0.48\textwidth]{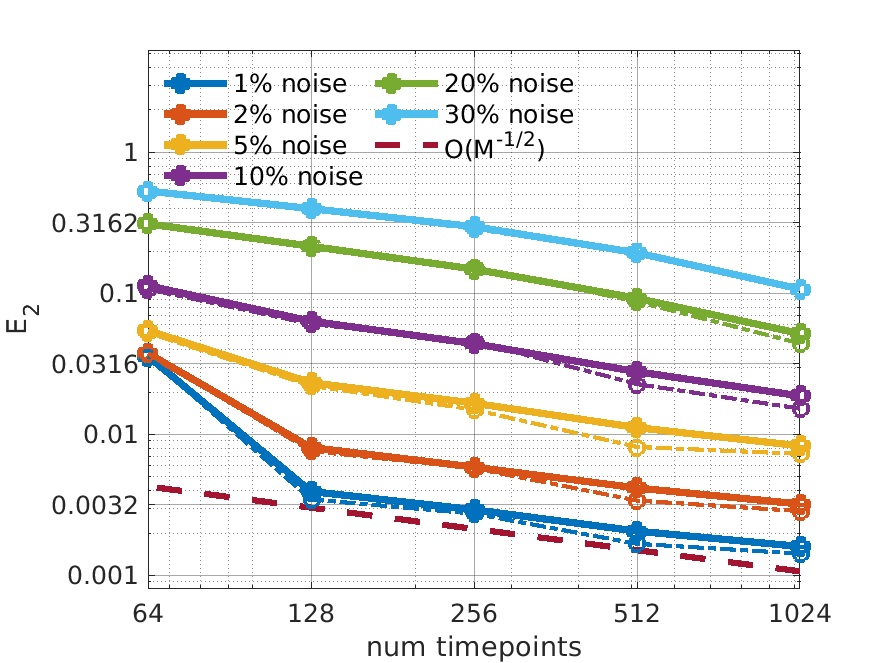}&
    \includegraphics[trim={0 0 26 20},clip,width=0.48\textwidth]{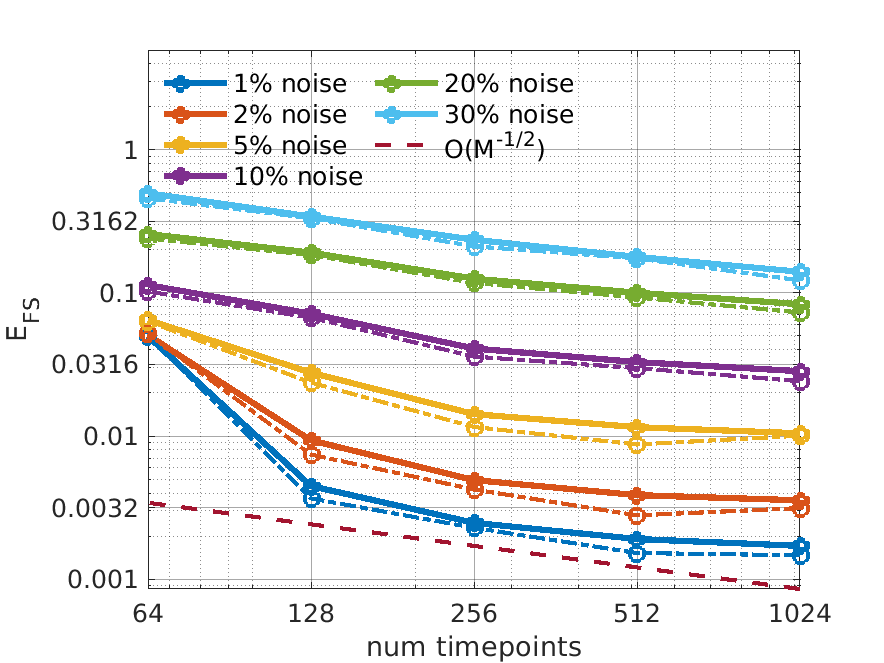}
\end{tabular}
\caption{\textbf{FitzHugh-Nagumo}: Estimation of parameters in the FitzHugh-Nagumo model (for plot details see Figure \ref{Logistic_Growth_fig} caption).}
\label{FitzHugh-Nagumo_fig}
\end{figure}

The Fitzhugh-Nagumo equations are a simplified model for an excitable neuron \cite{FitzHugh1961BiophysJ}. The equations contain six fundamental terms with coefficients to be identified. The cubic nonlinearity implies that the first-order covariance expansion in WENDy becomes inaccurate at high levels of noise. Nevertheless, Figure \ref{FitzHugh-Nagumo_fig} (lower plots) shows that WENDy produces on average $6\%$ coefficient errors at $10\%$ noise with only 128 timepoints, and only $7\%$ forward simulation errors (see upper left plot for an example dataset at this resolution). In many cases WENDy reduces the error by over $50\%$ from the FSNLS solution, with $80\%$ reductions for high noise and $M=1024$ timepoints (top right panel). For sparse data (e.g.\ 64 timepoints), numerical integration errors prevent estimation of parameters with lower than $3\%$ error, as the solution is nearly discontinuous in this case (jumps between datapoints are $\CalO(1)$). 

\subsubsection{Hindmarsh-Rose}

\begin{figure}
\centering
\begin{tabular}{@{}c@{}c@{}}
    \includegraphics[trim={-20 0 30 15},clip,width=0.48\textwidth]{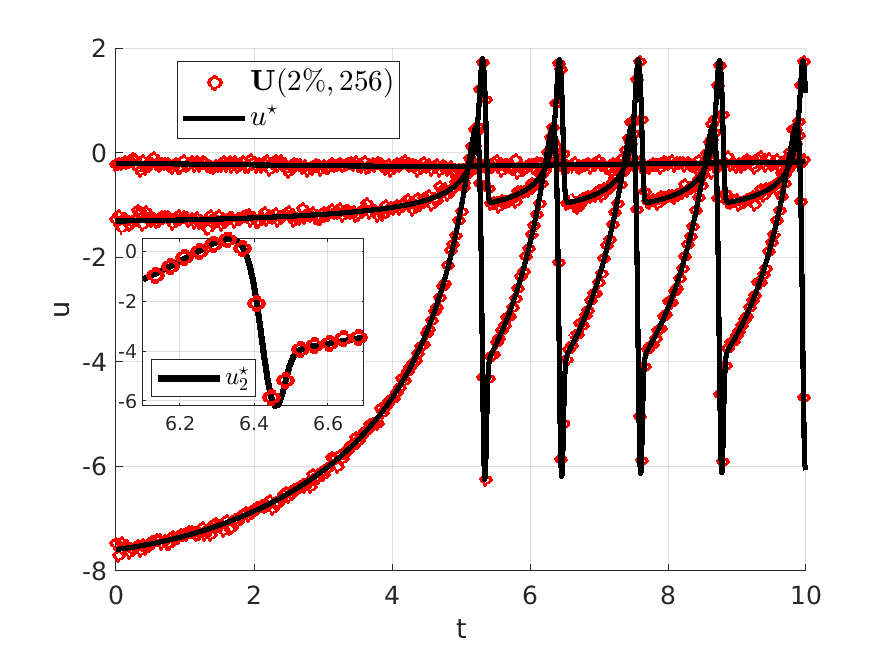} &
    \includegraphics[trim={-20 0 30 20},clip,width=0.48\textwidth]{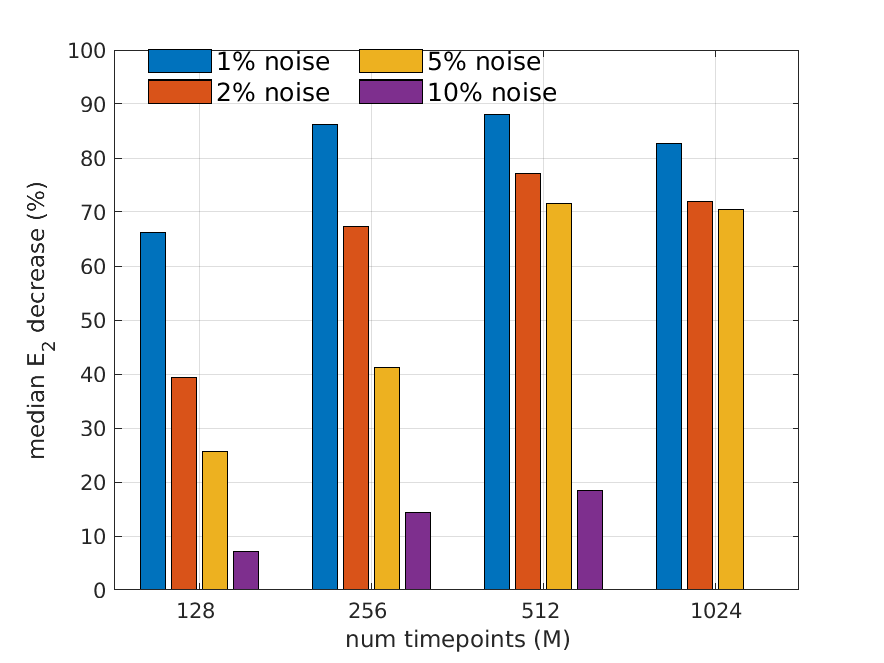} \\
    \includegraphics[trim={0 0 30 20},clip,width=0.48\textwidth]{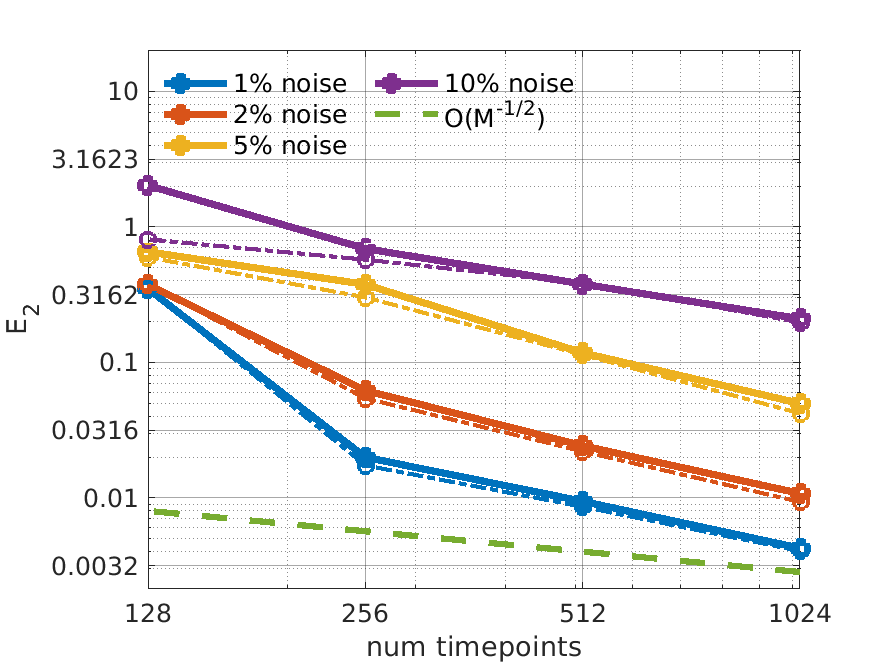}&
    \includegraphics[trim={0 0 30 20},clip,width=0.48\textwidth]{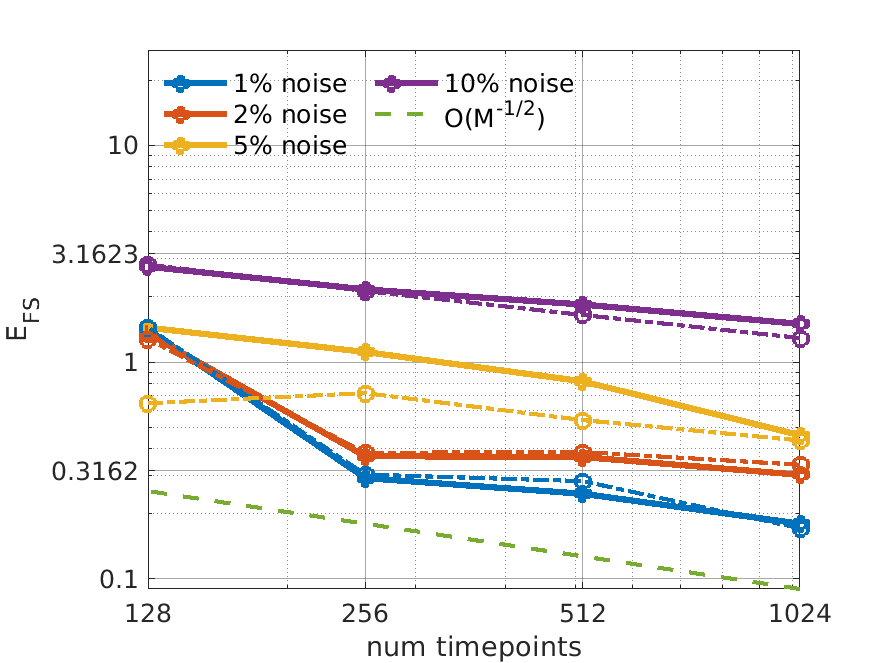}
\end{tabular}
\caption{\textbf{Hindmarsh-Rose}: Estimation of parameters in the Hindmarsh-Rose model (for plot details see Figure \ref{Logistic_Growth_fig} caption).}
\label{Hindmarsh-Rose_fig}
\end{figure}

The Hindmarsh-Rose model is used to emulate neuronal bursting and features 10 fundamental parameters which span 4 orders of magnitude \cite{HindmarshRose1984ProcRSocLondB}. Bursting behavior is observed in the first two solution components, while the third component represents slow neuronal adaptation with dynamics that are two orders of magnitude smaller in amplitude. Bursting produces steep gradients which render the dynamics numerically discontinuous at $M=128$ timepoints, while at $M=256$ there is at most one data point between peaks and troughs of bursts (see Figure \ref{Hindmarsh-Rose_fig}, upper left). Furthermore, cubic and quadratic nonlinearities lead to inaccuracies at high levels of noise. Thus, in a multitude of ways (multiple coefficient scales, multiple solution scales, steep gradients, higher-order nonlinearities, etc.) this is a challenging problem, yet an important one as it exhibits a canonical biological phenomenon. Figure \ref{Hindmarsh-Rose_fig} (lower left) shows that WENDy is robust to $2\%$ noise when $M\geq 256$, robust to $5\%$ noise when $M\geq 512$, and robust to $10\%$ noise when $M\geq 1024$.  It should be noted that since our noise model applies additive noise of equal variance to each component, relatively small noise renders the slowly-varying third component $u_3$ unidentifiable (in fact, the noise ratio of only $\Ubf^{(3)}$ exceeds $100\%$ when the total noise ratio is $10\%$). In the operable range of $1\%$-$2\%$ noise and $M\geq 256$, WENDy results in $70\%$-$90\%$ reductions in errors from the naive OLS solution, indicating that inclusion of the approximate covariance is highly beneficial under conditions which can be assumed to be experimentally relevant. We note that the forward simulation error here is not indicative of performance, as it will inevitably be large in all cases due to slight misalignment with bursts in the true data. 

\subsubsection{Protein Transduction Benchmark (PTB)}

\begin{figure}
\centering
\begin{tabular}{@{}c@{}c@{}}
    \includegraphics[trim={-20 0 30 15},clip,width=0.48\textwidth]{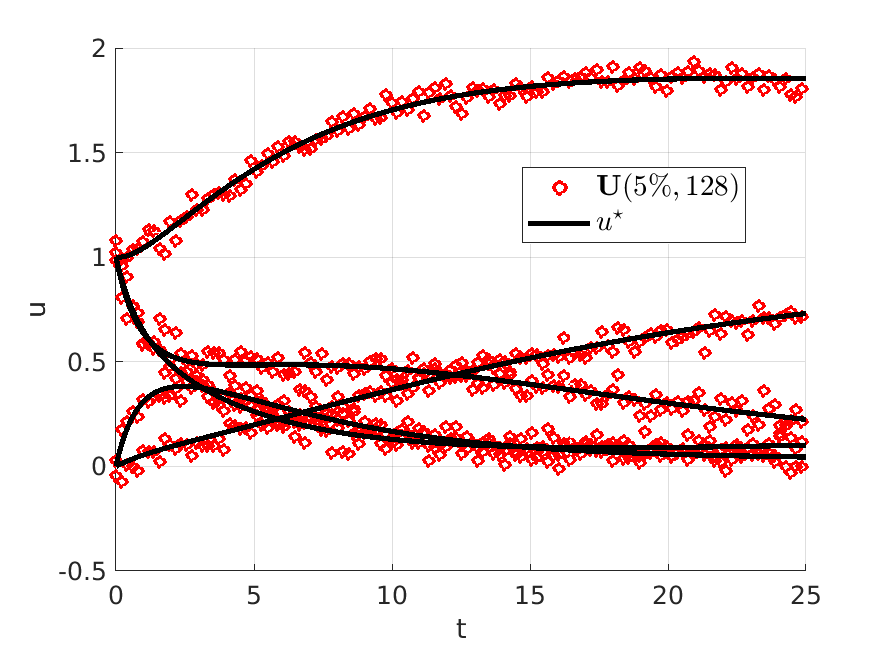} &
    \includegraphics[trim={-20 0 30 20},clip,width=0.48\textwidth]{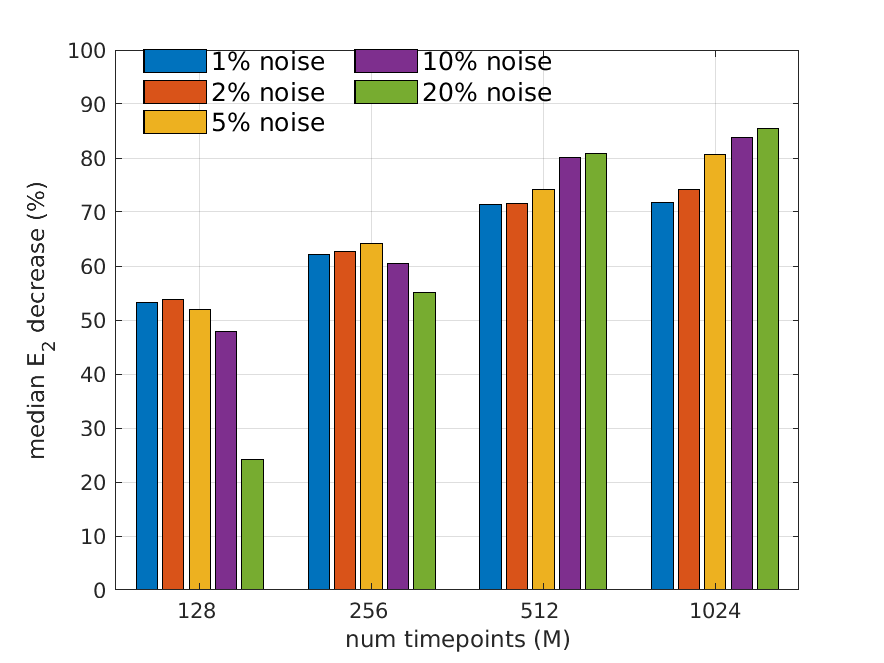} \\
    \includegraphics[trim={0 0 30 20},clip,width=0.48\textwidth]{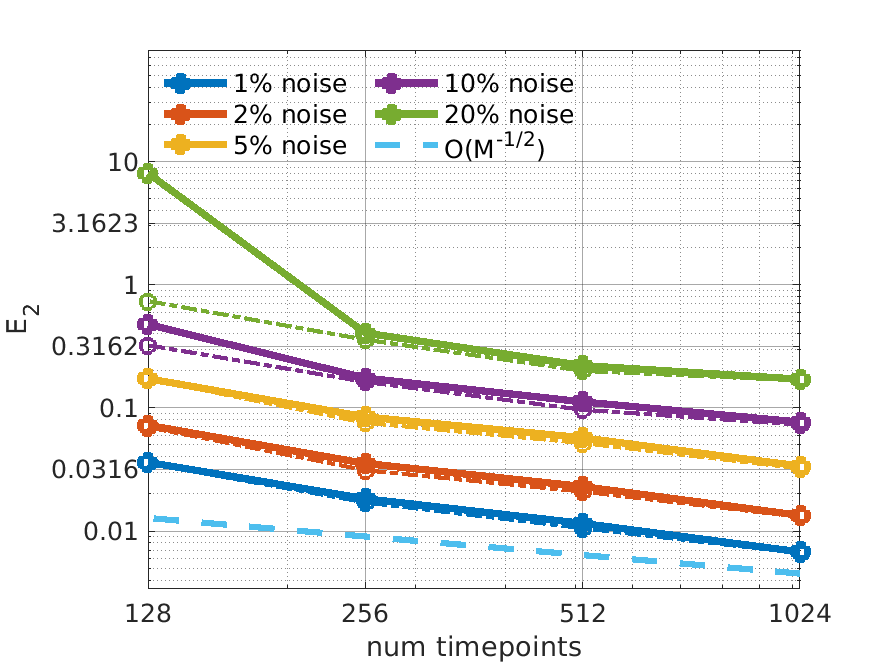}&
    \includegraphics[trim={0 0 30 20},clip,width=0.48\textwidth]{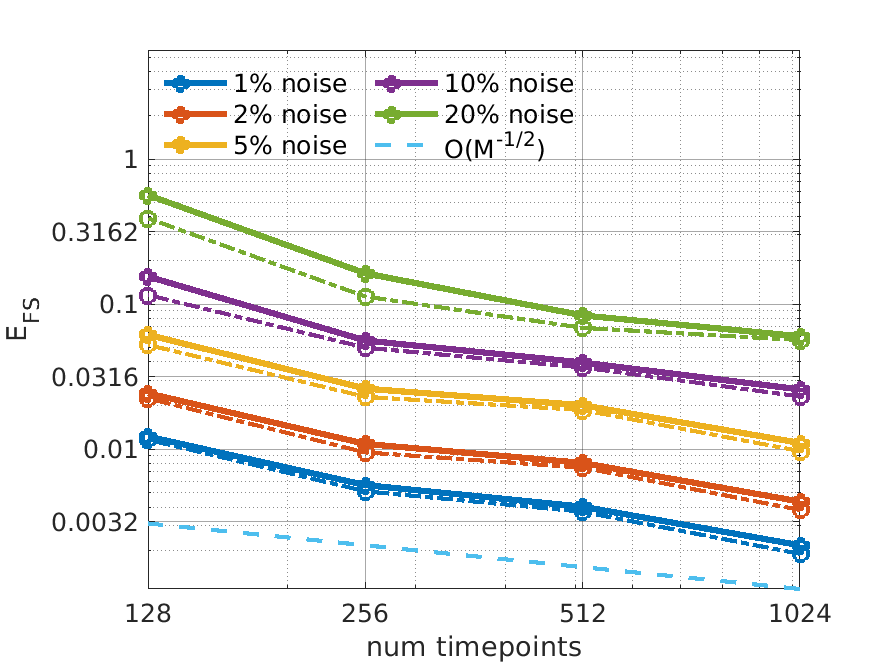}
\end{tabular}
\caption{\textbf{Protein Transduction Benchmark (PTB)}: Estimation of parameters in the PTB model (for plot details see Figure \ref{Logistic_Growth_fig} caption).}
\label{biochemM1_fig}
\end{figure}

The PTB model is a five-compartment protein transduction model identified in \cite{SchoeberlEichler-JonssonGillesEtAl2002NatBiotechnol} as a mechanism in the signaling cascade of epidermal growth factor (EGF). It was used in \cite{VyshemirskyGirolami2008Bioinformatics} to compare between four other models, and has since served as a benchmark for parameter estimation studies in biochemistry \cite{MacdonaldHusmeier2015BioinformaticsandBiomedicalEngineering,NiuRogersFilipponeEtAl2016Proc33rdIntConfMachLearn,KirkThorneStumpf2013CurrOpinBiotechnol}. The nonlinearites are quadratic and sigmoidal, the latter category producing nontrivial transformations of the additive noise. WENDy estimates the 11 parameters with reasonable accuracy when 256 or more timepoints are available (Figure \ref{biochemM1_fig}), which is sufficient to result in forward simulation errors often much less than $10\%$. The benefit of using WENDy over the OLS solution is most apparent for $M\geq 512$, where the coefficient errors are reduced by at least $70\%$, leading to forward simulation errors less than $10\%$, even at $20\%$ noise.

\subsection{Parameter uncertainties using learned covariance}
\label{sec:uncertainty}

\begin{figure}
\centering
\begin{tabular}{@{}c@{}c@{}}
    \includegraphics[trim={30 0 45 5},clip,width=0.5\textwidth]{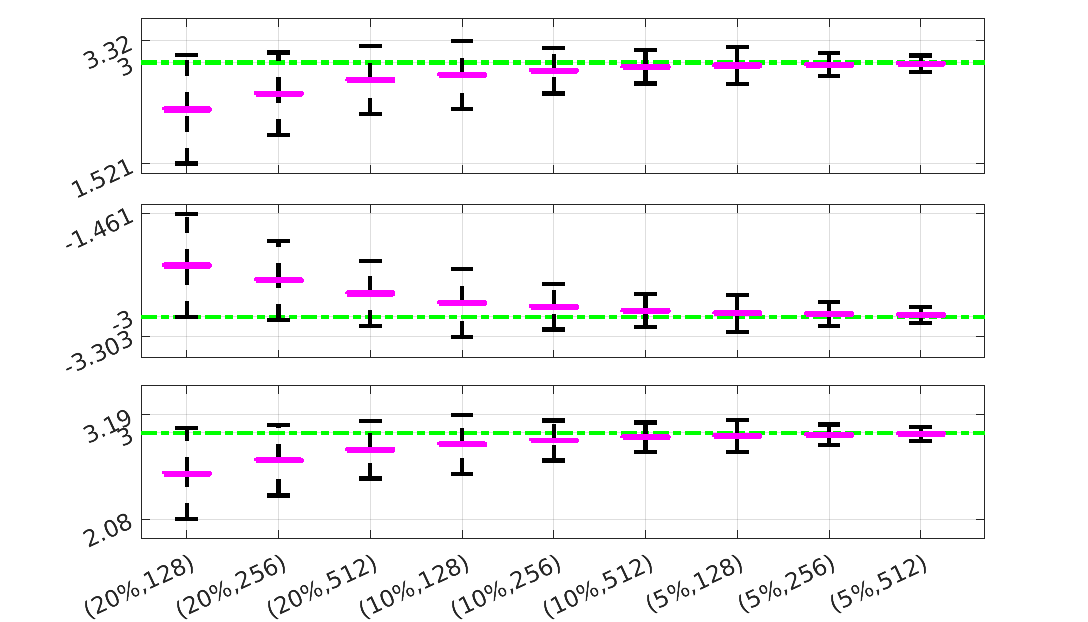}&
    \includegraphics[trim={30 0 45 5},clip,width=0.5\textwidth]{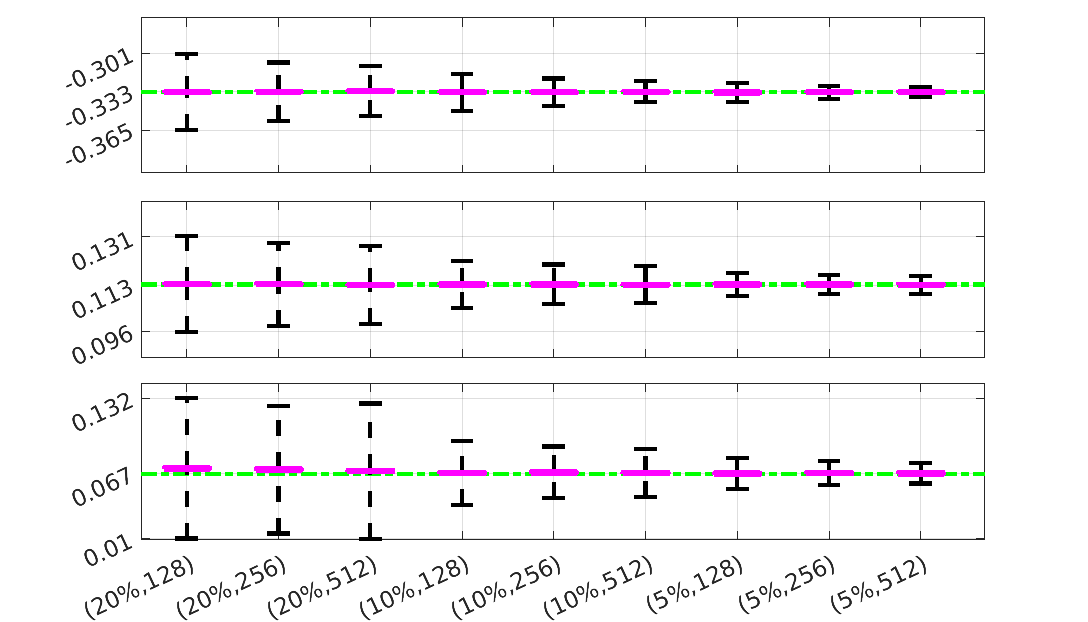}
\end{tabular}
\caption{\textbf{FitzHugh-Nagumo:} Performance of WENDy for all estimated parameters.  The true parameters are plotted in green, the purple lines indicate the average learned parameters over all experiments and the black lines represent the 95\% confidence intervals obtained from averaging the learned parameter covariance matrices $\Sbf$. The $x$-axis indicates noise level and number of timepoints for each interval.}
\label{FHN_CI}
\end{figure}
\begin{figure}
\centering
\begin{tabular}{@{}c@{}c@{}}
    \includegraphics[trim={35 25 55 10},clip,width=0.5\textwidth]{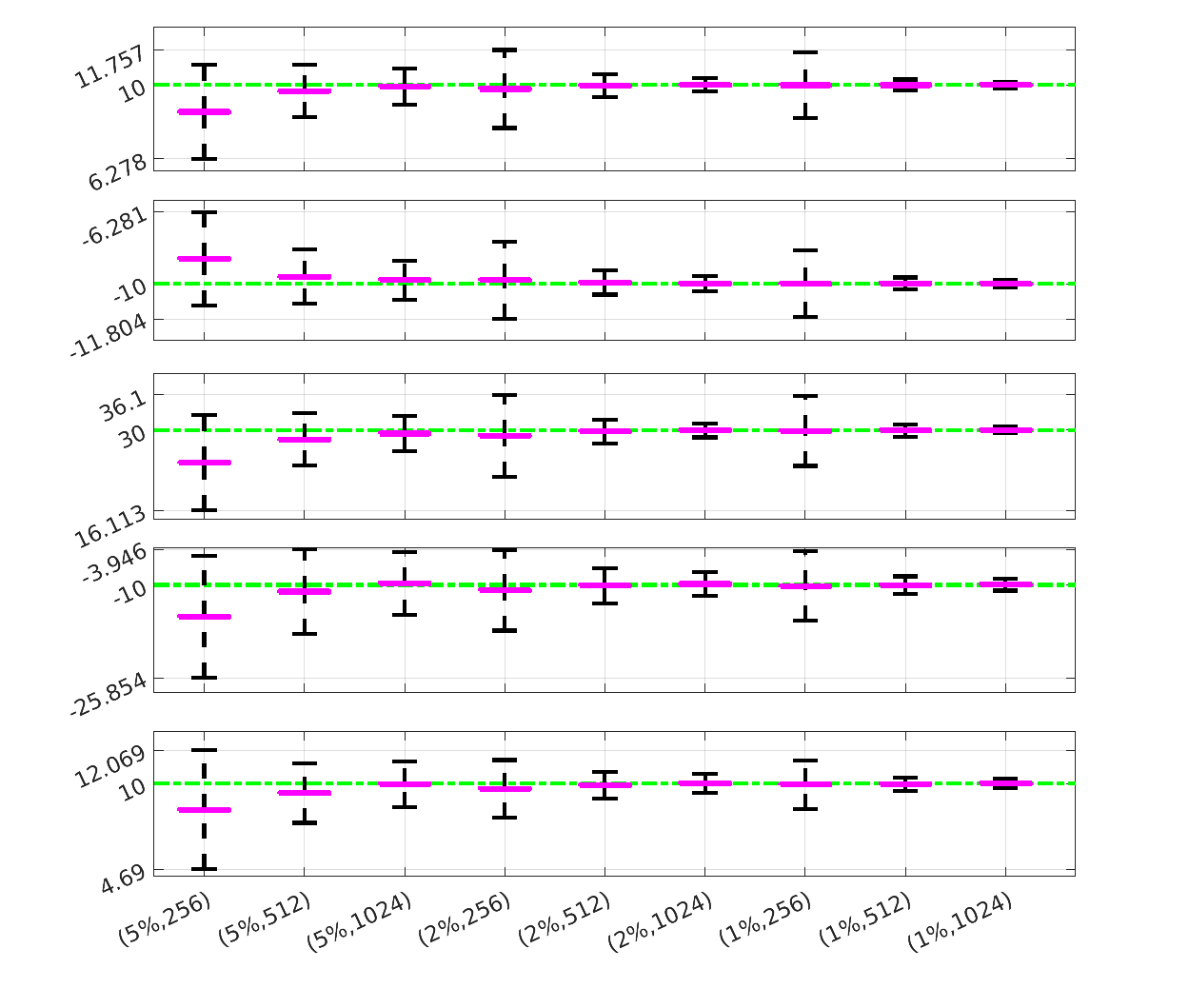}&
    \includegraphics[trim={35 25 55 10},clip,width=0.5\textwidth]{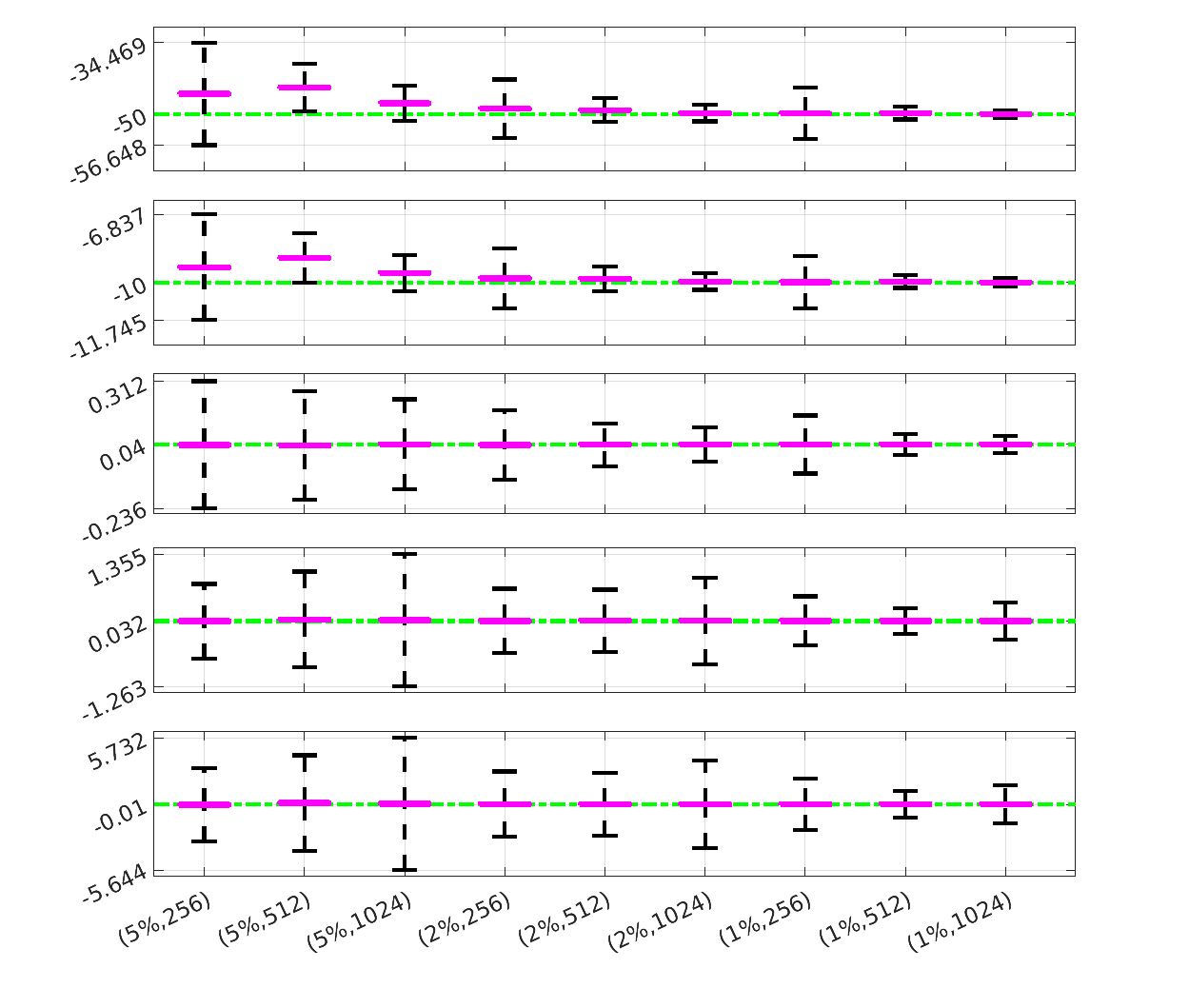}
    \end{tabular}
\caption{\textbf{Hindmarsh-Rose:} Performance of WENDy for all estimated parameters. See Figure \ref{FHN_CI} for a description.}
\label{HR_CI}
\end{figure}

We now demonstrate how the WENDy methodology may be used to inform the user about uncertainties in the parameter estimates. Figures \ref{FHN_CI} and \ref{HR_CI} contain visualizations of confidence intervals around each parameter in the FitzHugh-Nagumo and Hindmarsh-Rose models computed from the diagonal elements of the learned parameter covariance matrix $\Sbf$. Each combination of noise level and number of timepoints yields a 95\% confidence interval around the learned parameter\footnote{Scripts are available at  \url{https://github.com/MathBioCU/WENDy} to generate similar plots for the other examples.}. As expected, increasing the number of timepoints and decreasing the noise level leads to more certainty in the learned parameters, while lower quality data leads to higher uncertainty. Uncertainty levels can be used to inform experimental protocols and even be propagated into predictions made from learned models. One could also examine the off-diagonal correlations in $\Sbf$, which indicate how information flows between parameters. We aim to explore these directions in a future work.

\subsection{Comparison to nonlinear least squares}

We now briefly compare WENDy and forward solver-based nonlinear least squares (FSNLS) using walltime and relative coefficient error $E_2$ as criteria. For nonlinear least-squares one must specify the initial conditions for the ODE solve (IC), a simulation method (SM), and an initial guess for the parameters ($\wbf^{(0)}$). Additionally, stopping tolerances for the optimization method must be specified (Levenberg-Marquardt is used throughout). Optimal choices for each of these hyperparameters is an ongoing area of research. We have optimized FSNLS in ways that are unrealistic in practice in order to demonstrate the advantages of WENDy even when FSNLS  is performing somewhat optimally in both walltime and accuracy. Our hyperparameter selections are collected in Table \ref{NLShp} and discussed below.

To remove some sources of error from FSNLS, we use the true initial conditions $u(0)$ throughout, noting that these would not be available in practice. For the simulation method, we use state-of-the-art ODE solvers for each problem, namely for the stiff differential equations Fitzhugh-Nagumo and Hindmarsh-Rose we use MATLAB's \text{ode15s}, while for Lotka-Volterra and PTB we use \texttt{ode45}. In this way FSNLS is optimized for speed in each problem. We fix the relative and absolute tolerances of the solvers at $10^{-6}$ in order to prevent numerical errors from affecting results without asking for excessive computations. In practice, the ODE tolerance, as well as the solver, must be optimized to depend on the noise in the data, and the relation between simulation errors and parameters errors in FSNLS is an on-going area of research \cite{NardiniBortz2019InverseProbl}.

\begin{figure}
\centering
\begin{tabular}{@{}c@{}c@{}c@{}}
    \includegraphics[trim={0 0 25 0},clip,width=0.33\textwidth]{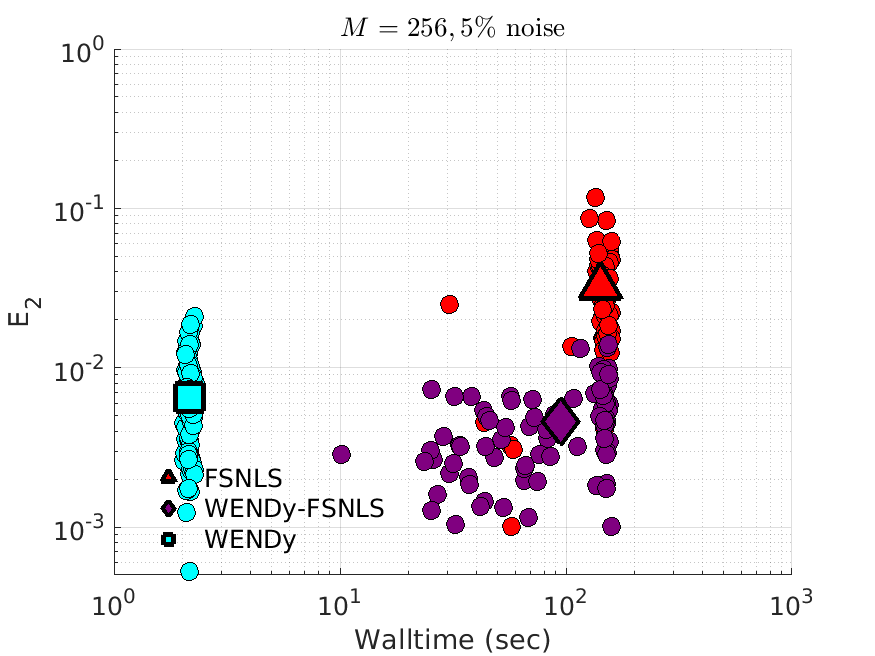}&
    \includegraphics[trim={0 0 25 0},clip,width=0.33\textwidth]{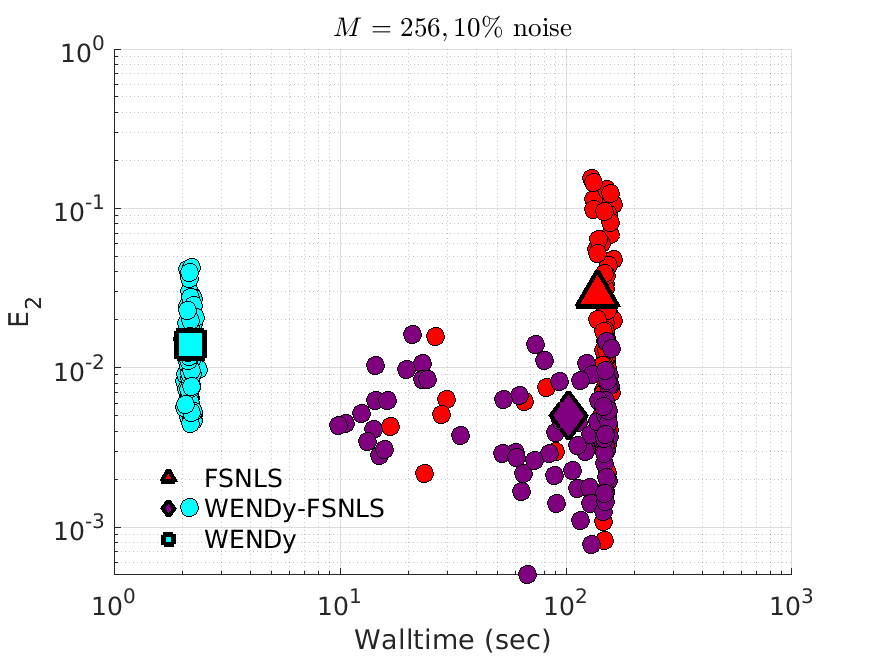}&
    \includegraphics[trim={0 0 25 0},clip,width=0.33\textwidth]{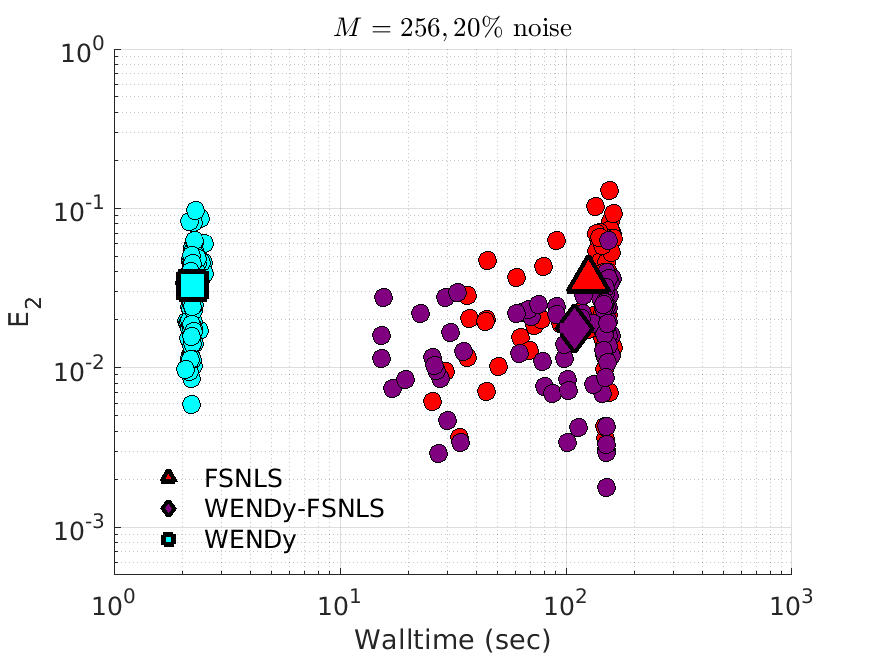}\\
    \includegraphics[trim={0 0 25 0},clip,width=0.33\textwidth]{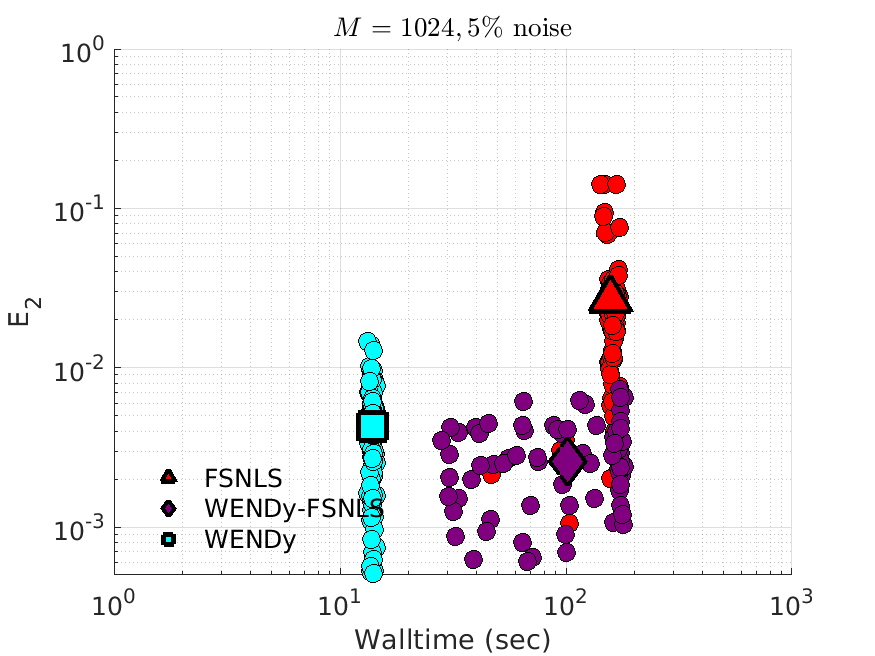}&
    \includegraphics[trim={0 0 25 0},clip,width=0.33\textwidth]{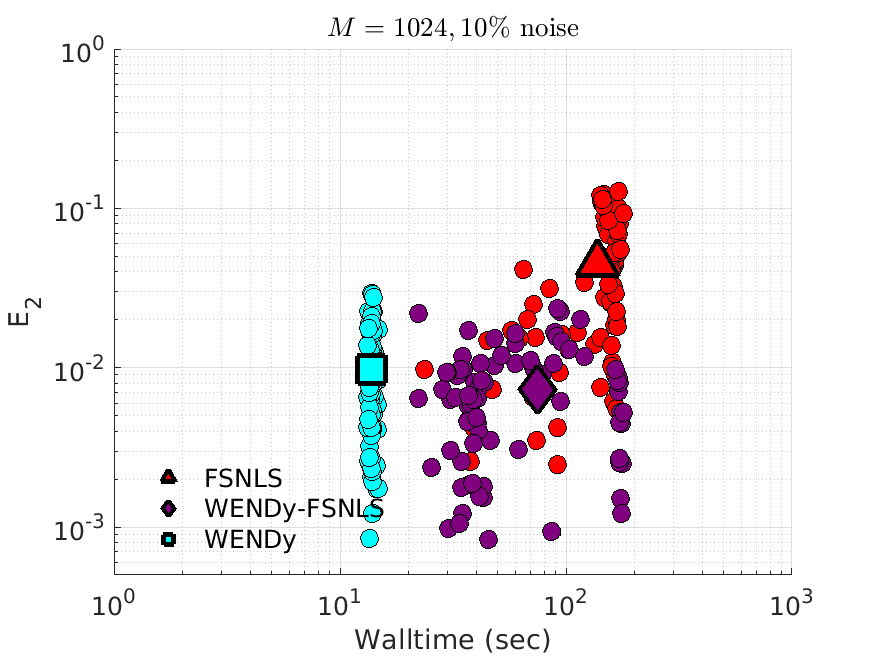}&
    \includegraphics[trim={0 0 25 0},clip,width=0.33\textwidth]{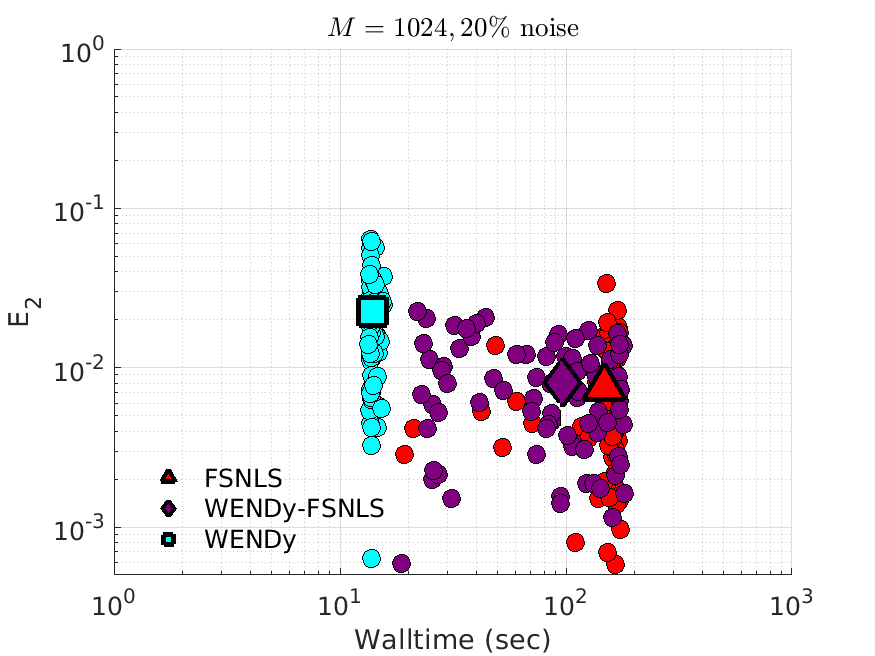}
\end{tabular}
\caption{Comparison between FSNLS, WENDy-FSNLS, and WENDy for the Lotka-Volterra model. Left to right: noise levels $\{5\%,10\%,20\%\}$. Top: 256 timepoints, bottom: 1024 timepoints. We note that the $M=1024$ with $20\%$ noise figure on the lower right suggests that WENDy results in slightly higher errors than the FSNLS.  This is inconsistent with all other results in this work and appears to be an outlier.  Understanding the source of this discrepancy is a topic or future work.}
\label{LV_NLS}
\end{figure}

\begin{figure}
\centering
\begin{tabular}{@{}c@{}c@{}c@{}}
    \includegraphics[trim={0 0 25 0},clip,width=0.33\textwidth]{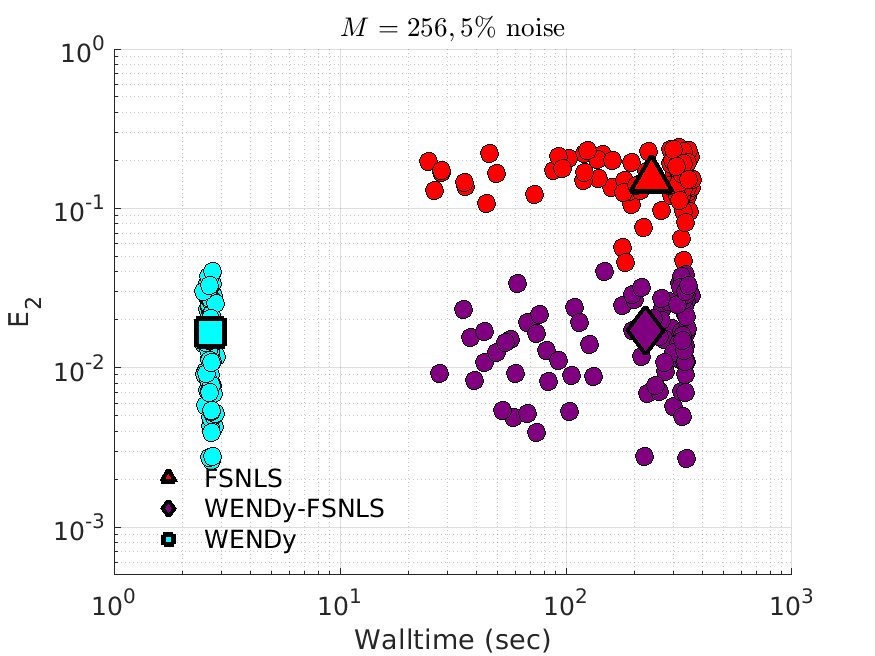}&
    \includegraphics[trim={0 0 25 0},clip,width=0.33\textwidth]{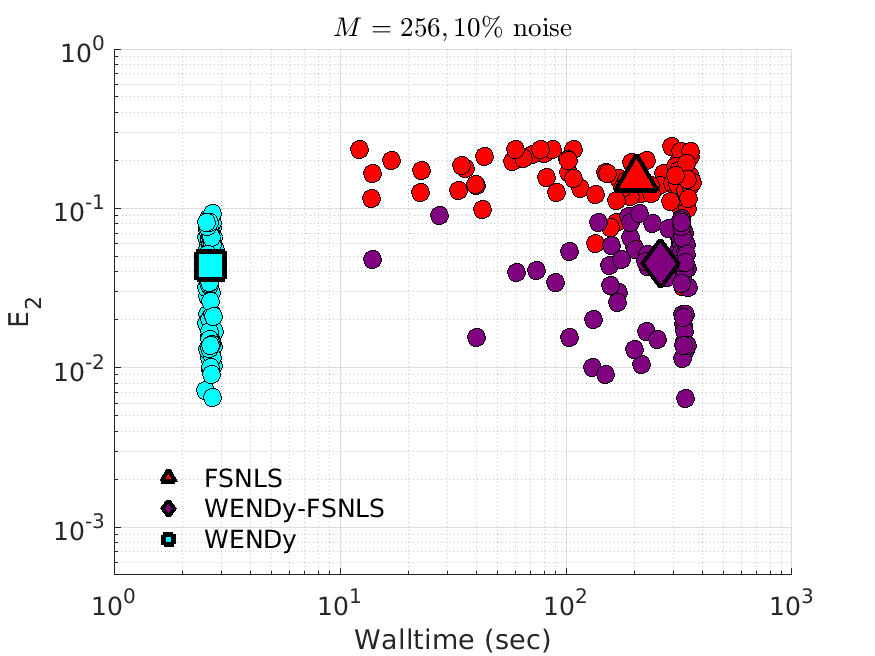}&
    \includegraphics[trim={0 0 25 0},clip,width=0.33\textwidth]{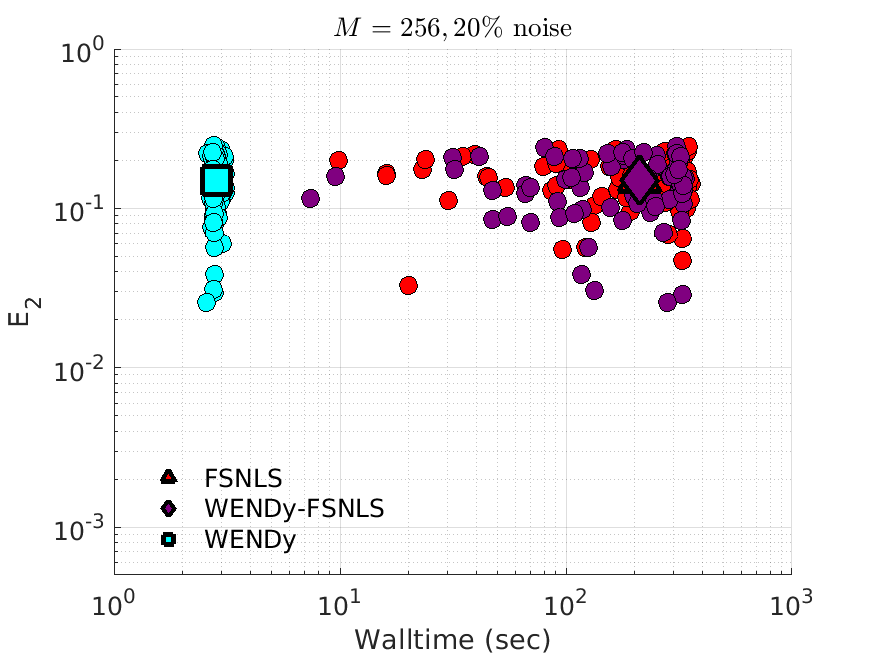}\\
    \includegraphics[trim={0 0 25 0},clip,width=0.33\textwidth]{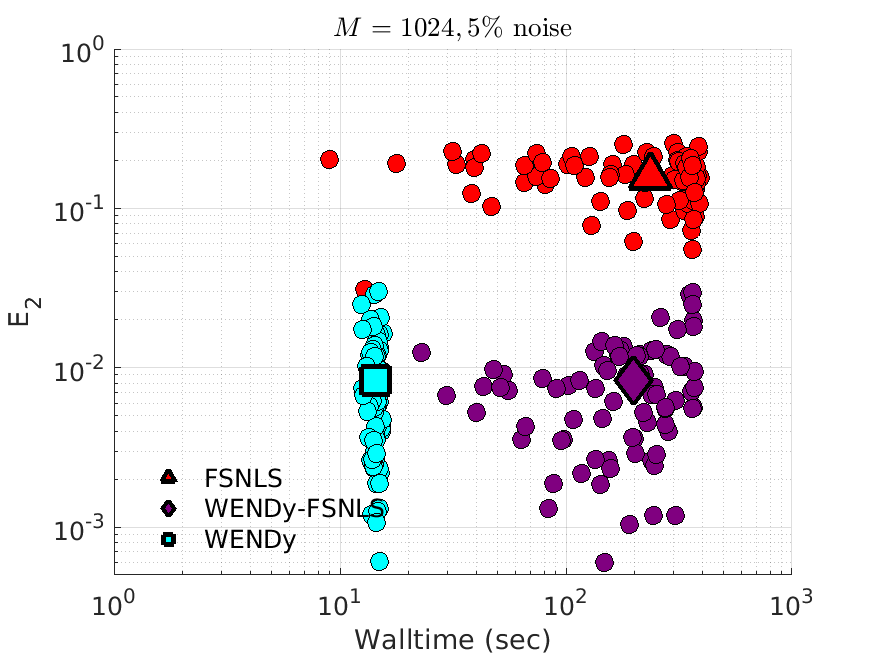}&
    \includegraphics[trim={0 0 25 0},clip,width=0.33\textwidth]{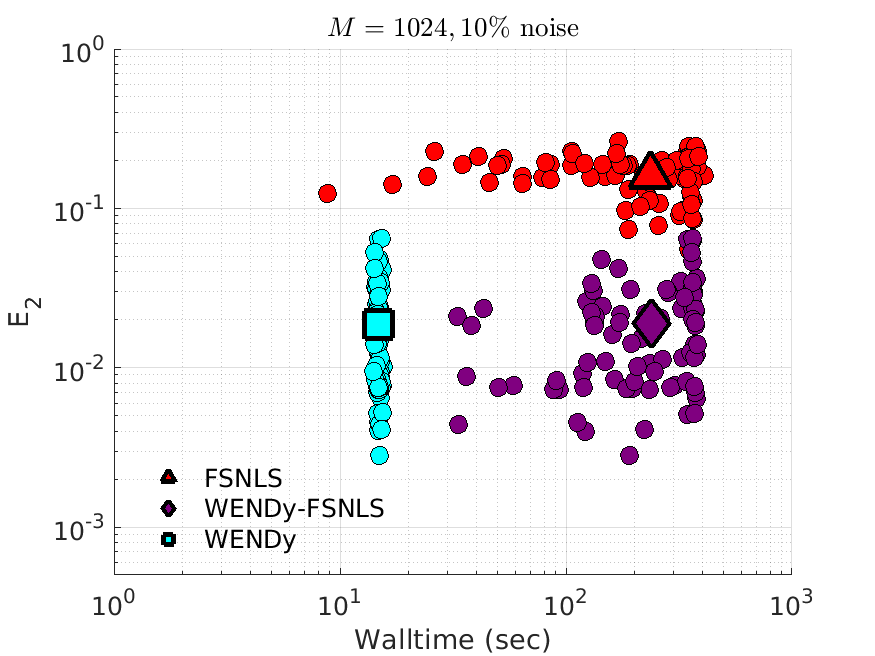}&
    \includegraphics[trim={0 0 25 0},clip,width=0.33\textwidth]{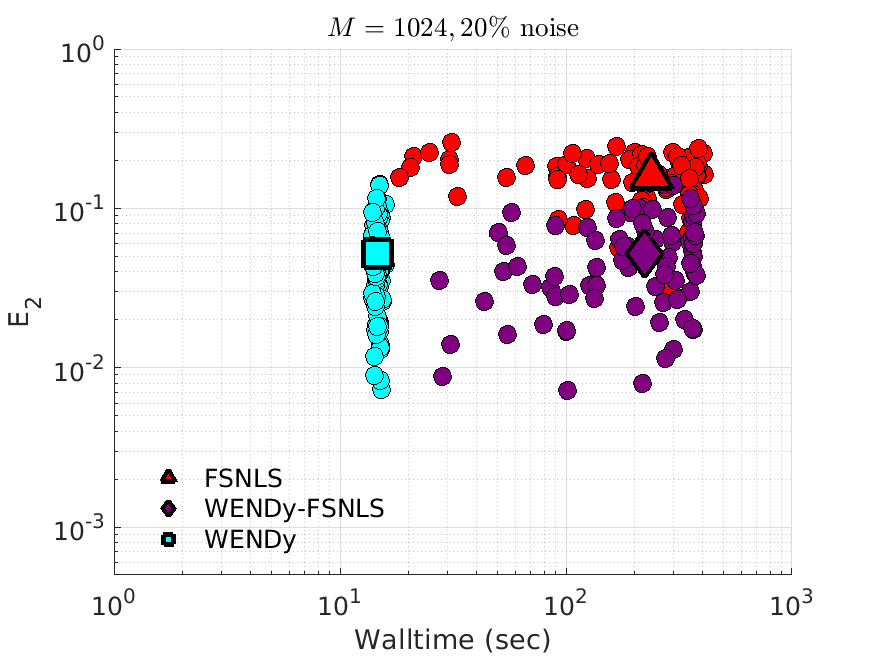}
\end{tabular}
\caption{Comparison between FSNLS, WENDy-FSNLS, and WENDy for the FitzHugh-Nagumo model. Left to right: noise levels $\{5\%,10\%,20\%\}$. Top: 256 timepoints, bottom: 1024 timepoints.}
\label{FHN_NLS}
\end{figure}

\begin{figure}
\centering
\begin{tabular}{@{}c@{}c@{}c@{}}
    \includegraphics[trim={0 0 25 0},clip,width=0.33\textwidth]{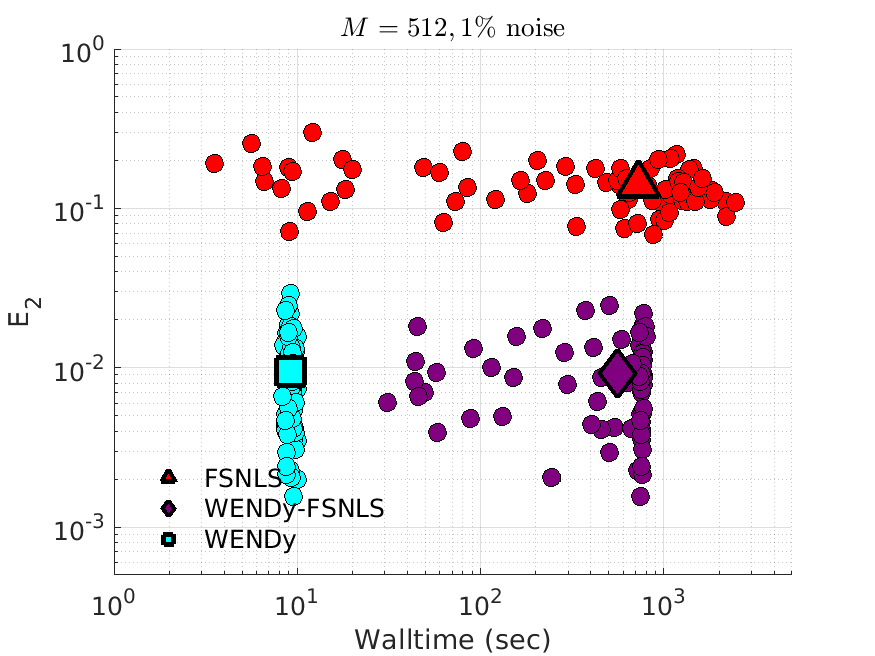}&
    \includegraphics[trim={0 0 25 0},clip,width=0.33\textwidth]{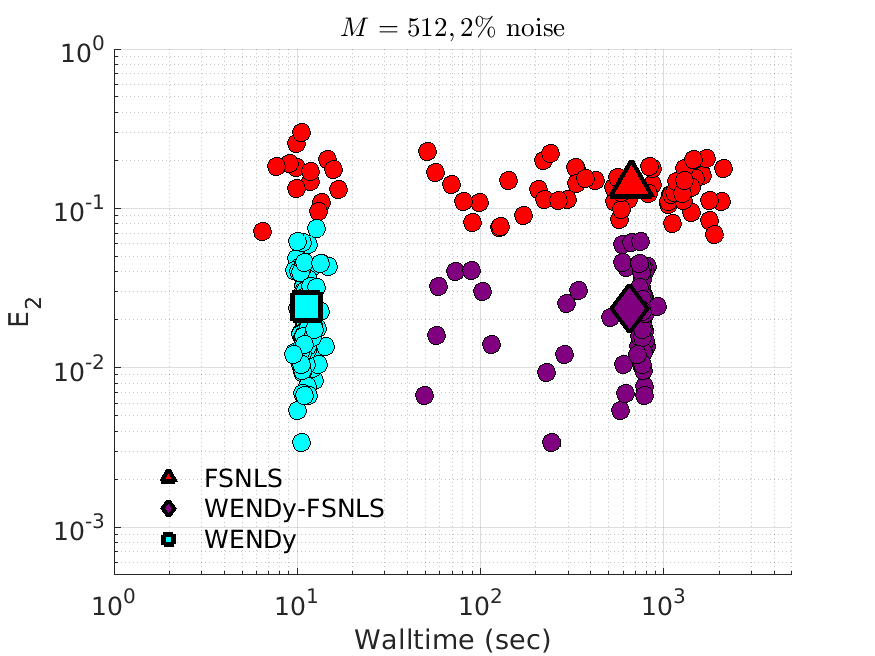}&
    \includegraphics[trim={0 0 25 0},clip,width=0.33\textwidth]{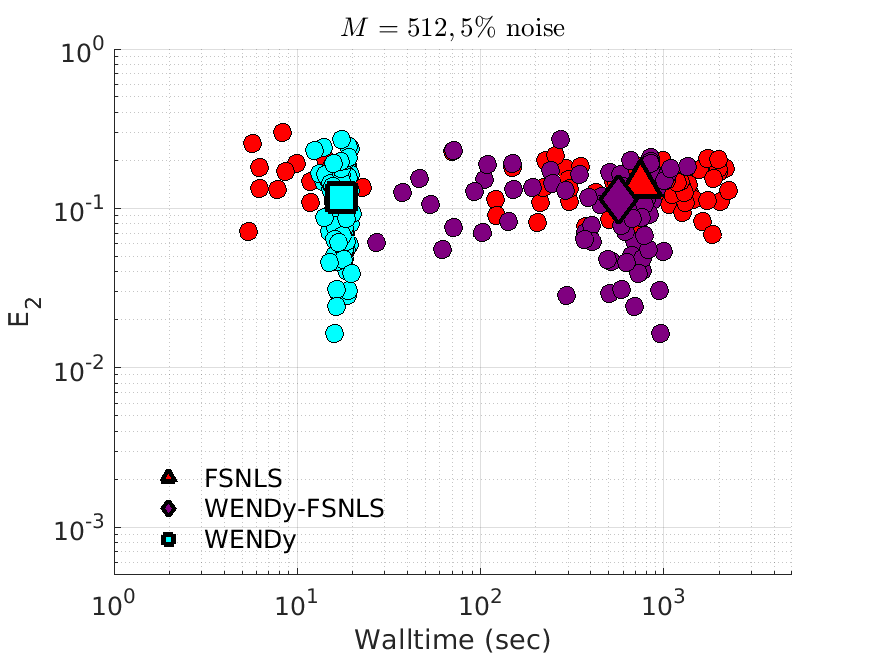}\\
    \includegraphics[trim={0 0 25 0},clip,width=0.33\textwidth]{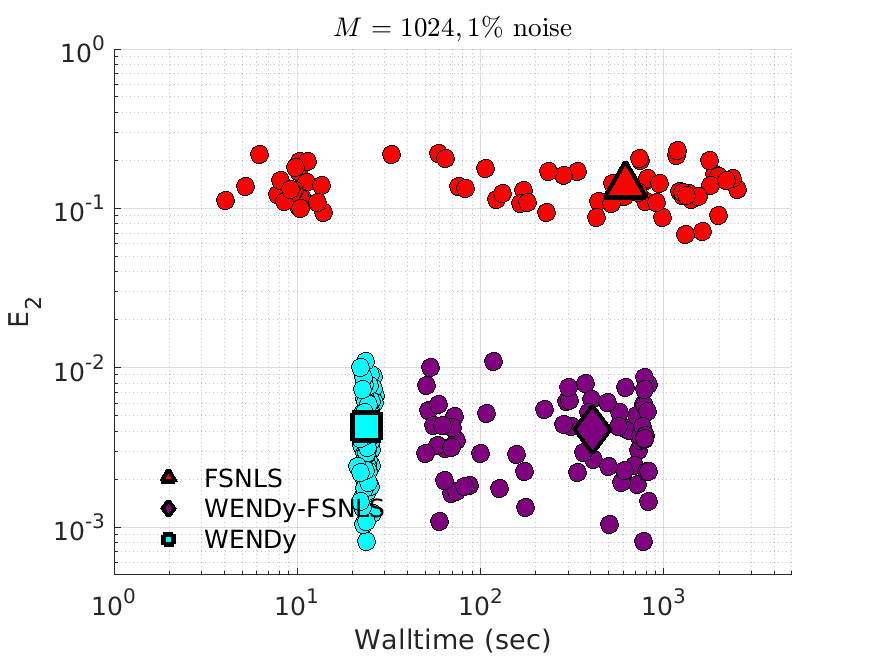}&
    \includegraphics[trim={0 0 25 0},clip,width=0.33\textwidth]{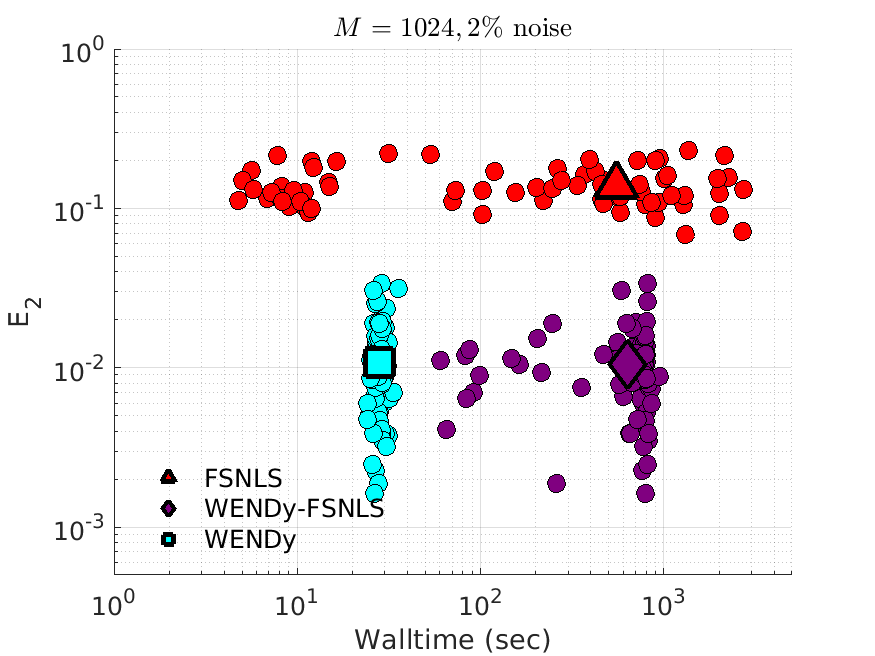}&
    \includegraphics[trim={0 0 25 0},clip,width=0.33\textwidth]{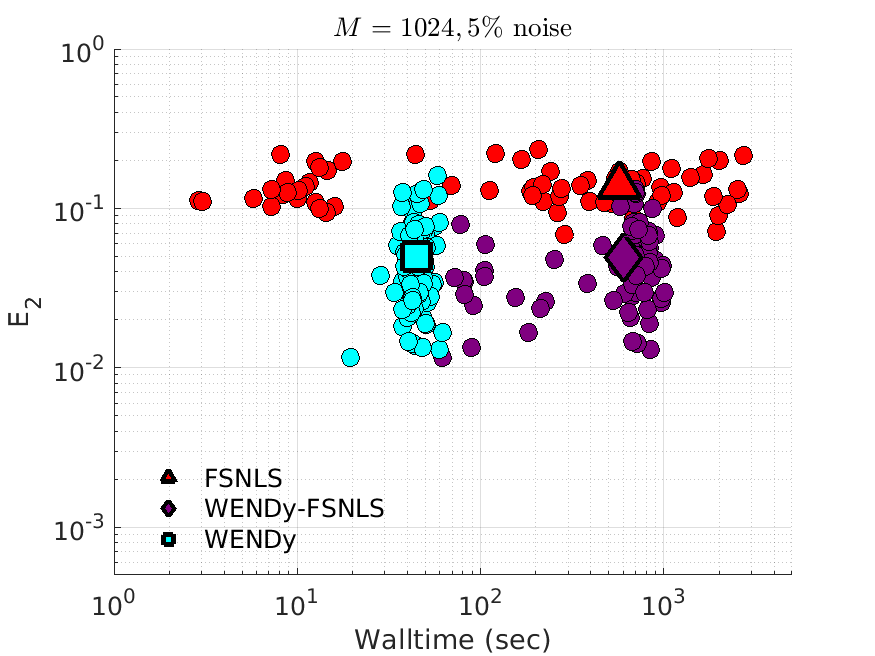}
\end{tabular}
\caption{Comparison between FSNLS, WENDy-FSNLS, and WENDy for the Hindmarsh-Rose model. Left to right: noise levels $\{1\%,2\%,5\%\}$. Top: 512 timepoints, bottom: 1024 timepoints.}
\label{HR_NLS}
\end{figure}

\begin{figure}
\centering
\begin{tabular}{@{}c@{}c@{}c@{}}
    \includegraphics[trim={0 0 25 0},clip,width=0.33\textwidth]{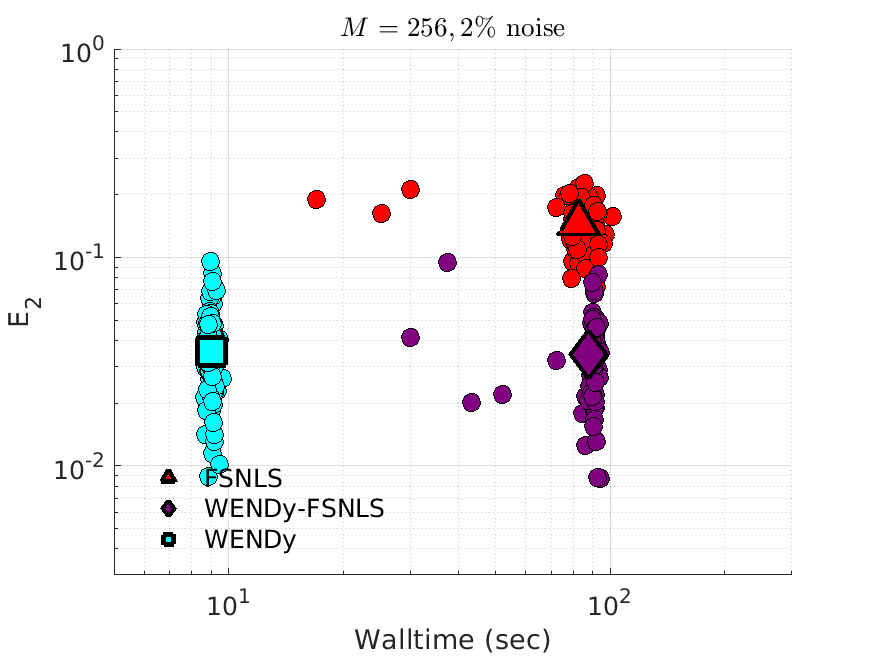}&
    \includegraphics[trim={0 0 25 0},clip,width=0.33\textwidth]{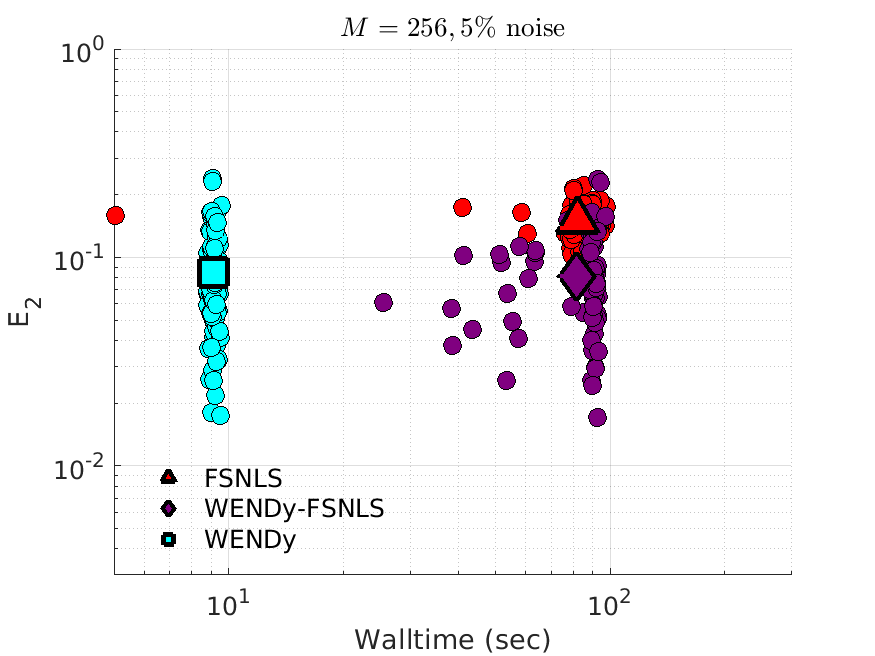}&
    \includegraphics[trim={0 0 25 0},clip,width=0.33\textwidth]{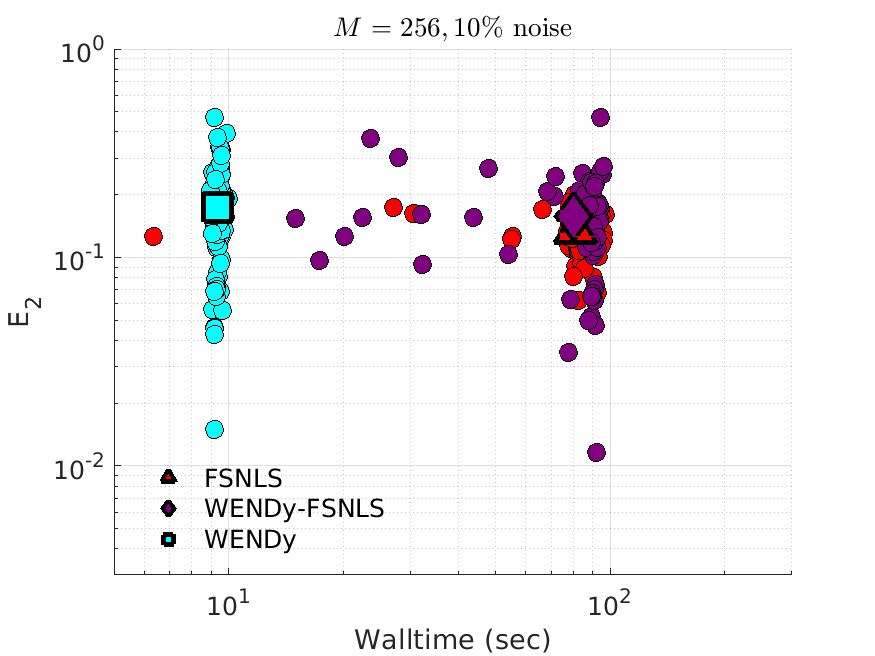}\\
    \includegraphics[trim={0 0 25 0},clip,width=0.33\textwidth]{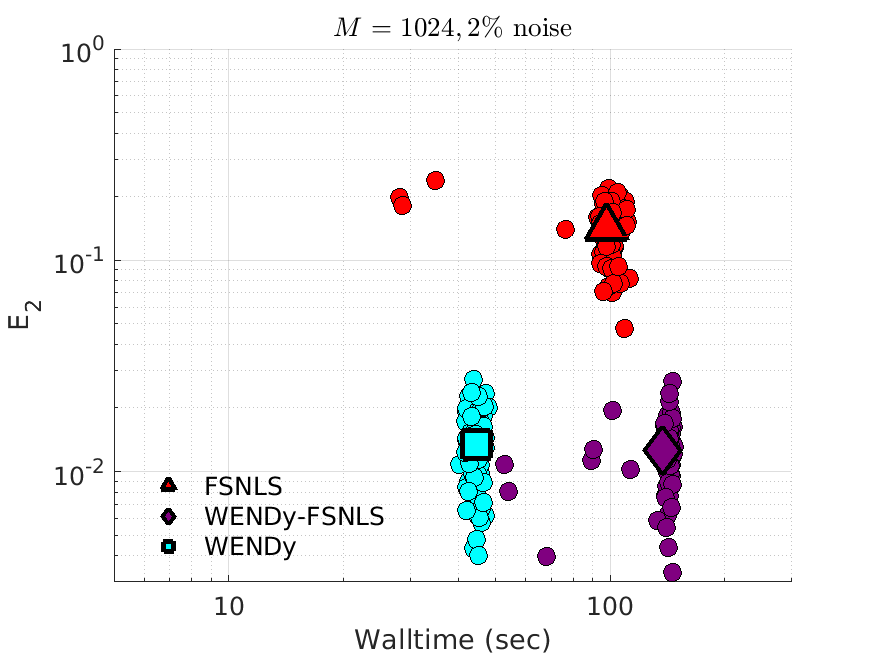}&
    \includegraphics[trim={0 0 25 0},clip,width=0.33\textwidth]{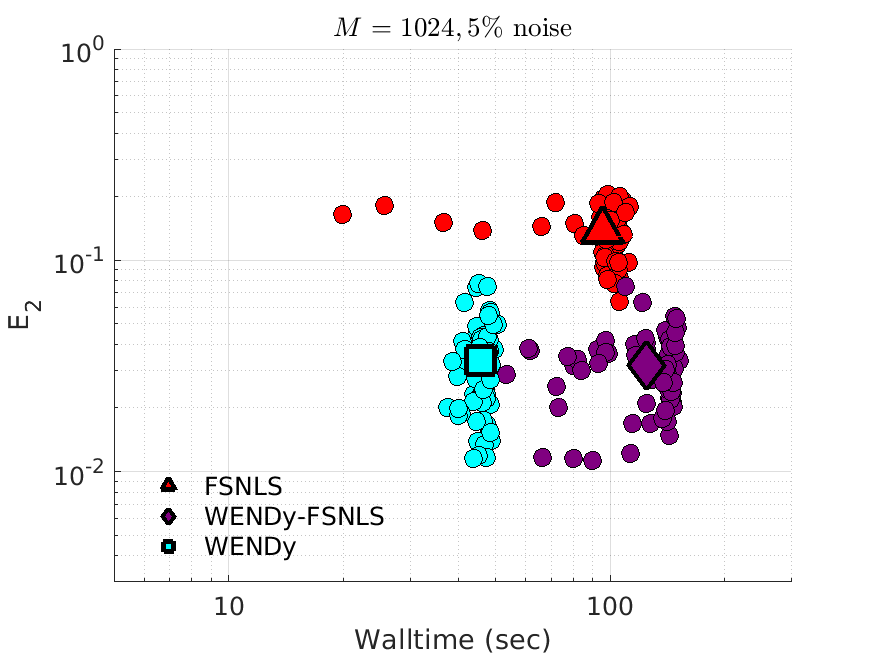}&
    \includegraphics[trim={0 0 25 0},clip,width=0.33\textwidth]{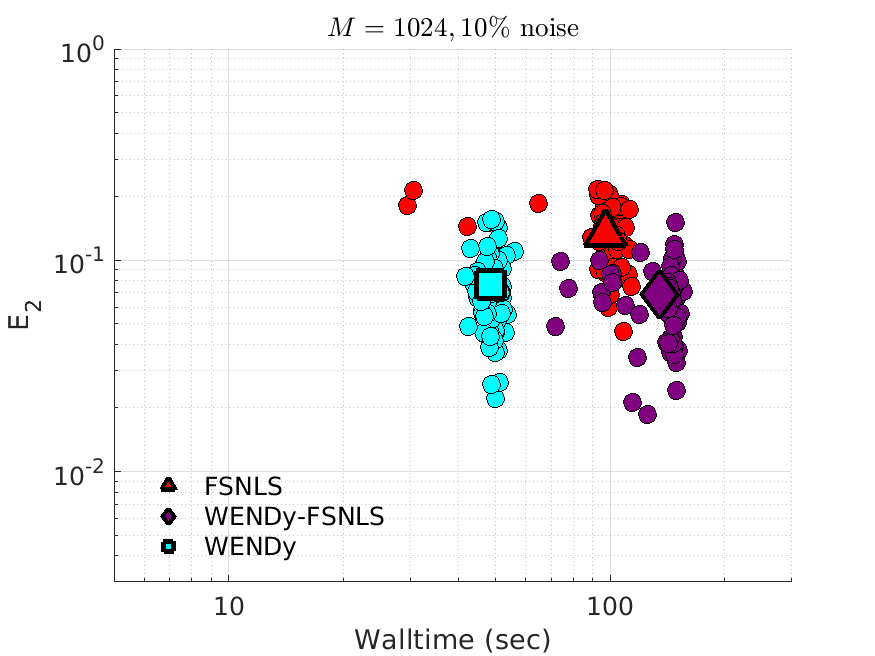}
\end{tabular}
\caption{Comparison between FSNLS, WENDy-FSNLS, and WENDy for the PTB model. Left to right: noise levels $\{2\%,5\%,10\%\}$. Top: 256 timepoints, bottom: 1024 timepoints.}
\label{PTB_NLS}
\end{figure}

Due to the non-convexity of the loss function in FSNLS, choosing a good initial guess $\wbf^{(0)}$ for the parameters $\wstar$ is crucial. For comparison, we use two strategies. The first strategy (simply labeled FSNLS in Figures \ref{LV_NLS}-\ref{PTB_NLS}), consists of running FSNLS on five initial guesses, where each parameter is sampled i.i.d\ from a uniform distribution, i.e., for the $i$th parameter, 
\[\wbf^{(0)}_i\sim \wstar_i+U([-\sigma/2,\sigma/2])\]
and keeping only the best-performing result. Since the sign of coefficients greatly impacts the stability of the ODE, we take the standard deviations to be
\begin{equation}\label{ICstrat1}
\sigma_j = 0.25|\wstar_j|
\end{equation}
so that initial guesses always have the correct sign but with approximately $25\%$ error from the true coefficients. (For cases like Hindmarsh-Rose, this implies that the small coefficients in $\wstar$ are measured to high accuracy relative to the large coefficients.) In practice, one would not have the luxury of selecting the lowest-error result of five independent trials of FSNLS, however it may be possible to combine several results to boost performance.

For the second initial guess strategy we set $\wbf^{(0)} =\what$, the output from WENDy (labeled WENDy-FSNLS in Figures \ref{LV_NLS}-\ref{PTB_NLS}). In almost all cases, this results in an increase in accuracy, and in many cases, also a decrease in walltime. 

\begin{table}
\centering
\begin{tabular}{|c|p{3cm}|p{2.5cm}|c|c|c|c|}
\hline
IC & Simulation method & $\wbf^{(0),\text{batch}}$ & $\wbf^{(0),\text{WENDy}}$ & \textsf{max.\ evals} & \textsf{max.\ iter} & \textsf{min.\ step}\\
\hline $u^\star(0)$ &  L-V, PTB: \texttt{ode45} \newline FH-N, H-R: \texttt{ode15s}\newline (abs/rel tol=$10^{-6}$) & $\wbf^{(0)}\sim ~U(\wstar,\pmb{\sigma})$,\newline best out of 5 & $\wbf^{(0)} = \what$ & 2000 & 500 & $10^{-8}$ \\
\hline
\end{tabular}
\caption{Hyperparameters for the FSNLS algorithm.}
\label{NLShp}
\end{table}

Figures \ref{LV_NLS}-\ref{PTB_NLS} display comparisons between FSNLS, WENDy-FSNLS, and WENDy for Lotka-Volterra, FitzHugh-Nagumo, Hindmarsh-Rose, and PTB models. In general, we observe that WENDy provides significant decreases in walltime and modest to considerable increases in accuracy compared to the FSNLS solution. Due to the additive noise structure of the data, this is surprising because FSNLS corresponds to (for normally distributed measurement errors) a maximum likelihood estimation, while WENDy only provides a first order approximation to the statistical model.  At lower resolution and higher noise (top right plot in Figures \ref{LV_NLS}-\ref{PTB_NLS}), all three methods are comparable in accuracy, and WENDy decreases the walltime by two orders of magnitude. In several cases, such as Lotka-Volterra Figure \ref{LV_NLS}, the WENDy-FSNLS solution achieves a lower error than both WENDy and FSNLS, and improves on the speed of FSNLS. For Hindmarsh-Rose, even with high-resolution data and low noise (bottom left plot of Figure \ref{HR_NLS}), FSNLS is unable to provide an accurate solution ($E_2\approx 0.2$), while WENDy and WENDy-FSNLS result in $E_2\approx 0.005$. The clusters of FSNLS runs in Figure \ref{HR_NLS} with walltimes $\approx 10$ seconds correspond to local minima, a particular weakness of FSNLS, while the remaining runs have walltimes on the order of 20 minutes, compared to 10-30 seconds WENDy. We see a similar trend in $E_2$ for the PTB model (Figure \ref{PTB_NLS}), with $E_2$ rarely dropping below $10\%$, however in this case FSNLS runs in a more reasonable amount of time, taking only $\approx 100$ seconds. The WENDy solution offers speed and error reductions.  For high-resolution data ($M=1024$), WENDy runs in 40-50 seconds on PTB data due to the impact of $M$ and $d$, the number of ODE compartments (here $d=5$), on the computational complexity. It is possible to reduce this using more a sophisticated implementation (in particular, symbolic computations are used to take gradients of generic functions, which could be precomputed). 

\begin{figure}
\centering
\begin{tabular}{cc}
    \includegraphics[trim={0 0 25 0},clip,width=0.48\textwidth]{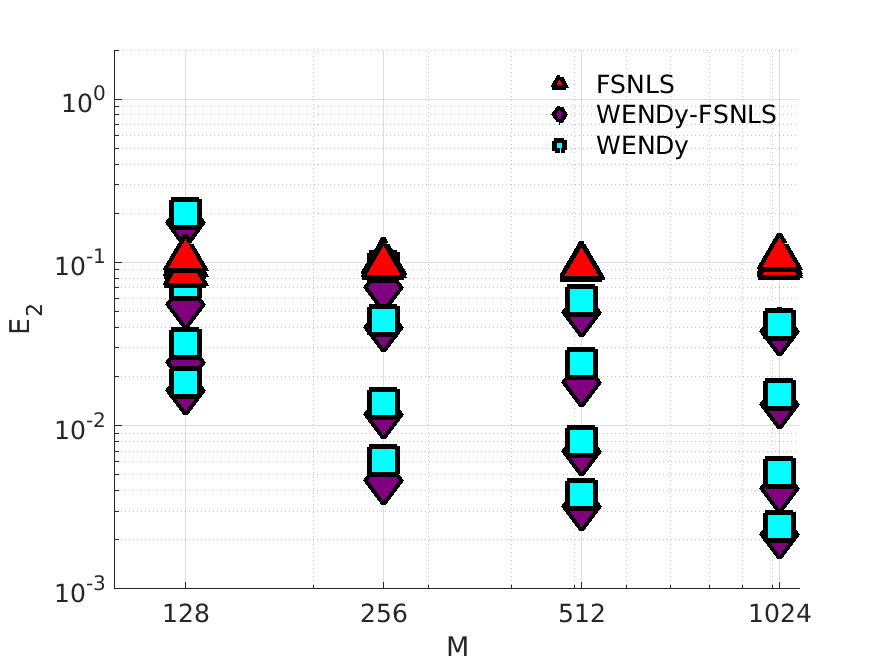}&
    \includegraphics[trim={0 0 25 0},clip,width=0.48\textwidth]{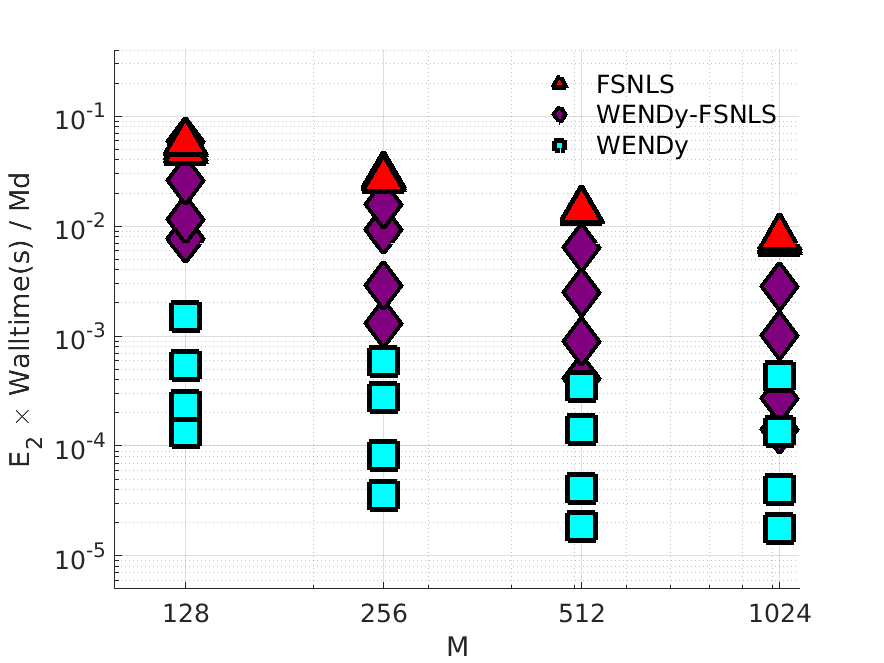}
\end{tabular}
\caption{Average performance of FSNLS, WENDy-FSNLS, and WENDy over Lotka-Volterra, FitzHugh-Nagumo, Hindmarsh-Rose and PTB for noise ratios $\sigma_{NR}\in \{0.01,0.02,0.05,0.1\}$. To account for scaling between examples, the geometric mean across the four examples is reported in each plot. Left: average relative coefficient error $E_2$ vs. number of timepoints $M$; right: relative coefficient error $E_2$ multiplied by walltime per datapoint vs. $M$. In each case, increasing noise levels $\sigma_{NR}$ correspond to increasing values along the $y$-axis. Both plots suggest that WENDy and WENDy-FSNLS each provide accuracy and walltime improvements over FSNLS with best-of-five random initial parameter guesses.}
\label{all_NLS}
\end{figure}

Finally, the aggregate performance of WENDy, WENDy-FSNLS, and FSNLS is reported in Figure \ref{all_NLS}, which reiterates the trends identified in the previous Figures. Firstly, WENDy provides significant accuracy and walltime improvements over FSNLS. It is possible that FSNLS results in lower error for very small sample sizes (see $M=128$ results in the left plot), although this comes at a much higher computational cost. Secondly, WENDy-FSNLS provides similar accuracy improvements over FSNLS and improves the walltime per datapoint score, suggesting that using WENDy as an initial guess may alleviate the computational burden in cases where FSNLS is competitive.

\section{Concluding Remarks\label{sec:ConcDisc}}

In this work, we have proposed the Weak-form Estimation of Nonlinear Dynamics (WENDy) method for directly
estimating model parameters, without relying on forward solvers. The essential feature of the method involves converting the strong form representation of a
model to its weak form and then substituting in the data and solving a regression problem for the parameters.  The method is robust to substantial amounts of noise, and in particular to levels frequently seen in biological experiments.

As mentioned above, the idea of substituting data into the weak form of an equation followed by a least squares solve for the parameters has existed since at least the mid 1950's \cite{Shinbrot1954NACATN3288}.  However, FSNLS-based methods have proven highly successful and are ubiquitous in the parameter estimation literature and software. The disadvantage of FSNLS is that fitting using repeated forward solves comes at a substantial computational cost and with unclear dependence on the initial guess and hyperparameters (in both the solver and the optimizer).  Several researchers over the years have created direct parameter estimation methods (that do not rely on forward solves), but they have historically included some sort of data smoothing step.  The primary issue with this is that projecting the data onto a spline basis (for example) represents the data using a basis which does not solve the original equation\footnote{This is a problem WENDy does not suffer from as there is no pre-smoothing of the data.}. Importantly, that error propagates to the error in the parameter estimates. However, we note that the WENDy framework introduced here is able to encapsulate previous works that incorporate smoothing, namely by including the smoothing operator in the covariance matrix $\widehat{\Cbf}$.

The conversion to the weak form is essentially a weighted integral transform of the equation. As there is no projection onto a non-solution based function basis, the weak-form approach bypasses the need to estimate the true solution to directly estimate the parameters. 

The main message of this work is that weak-form-based direct parameter estimation offers intriguing advantages over FSNLS-based methods. In almost all the examples shown in this work and in particular for larger dimensional systems with high noise, the WENDy method is faster and more accurate by orders of magnitude. In rare cases where an FSNLS-based approach yields higher accuracy, WENDy can be used as an efficient method to identify a good initial guess for parameters.

\begin{acknowledgements}
The authors would like to thank Dr.~Michael Zager (Pfizer) and Dr.~Clay Thompson
(SAS) for offering insight into the state of the art parameter estimation
methods used in industry.
\end{acknowledgements}

\section*{\textemdash \textemdash \textemdash \textemdash \textemdash \textemdash \textemdash{}}

\bibliographystyle{spmpsci}
\addcontentsline{toc}{section}{\refname}\bibliography{mathbioCU}

\begin{thebibliography}{10}
\providecommand{\url}[1]{{#1}}
\providecommand{\urlprefix}{URL }
\expandafter\ifx\csname urlstyle\endcsname\relax
  \providecommand{\doi}[1]{DOI~\discretionary{}{}{}#1}\else
  \providecommand{\doi}{DOI~\discretionary{}{}{}\begingroup
  \urlstyle{rm}\Url}\fi

\bibitem{BanksKunisch1989}
Banks, H.T., Kunisch, K.: Estimation {{Techniques}} for {{Distributed Parameter
  Systems}}, \emph{Systems and {{Control}}: {{Foundations}} and
  {{Applications}}}, vol.~1.
\newblock {Birkh\"auser Boston}, {Boston, MA} (1989)

\bibitem{Bellman1969MathematicalBiosciences}
Bellman, R.: A new method for the identification of systems.
\newblock Mathematical Biosciences \textbf{5}(1-2), 201--204 (1969).
\newblock \doi{10.1016/0025-5564(69)90042-X}

\bibitem{BertsimasGurnee2023NonlinearDyn}
Bertsimas, D., Gurnee, W.: Learning sparse nonlinear dynamics via mixed-integer
  optimization.
\newblock Nonlinear Dyn.  (2023).
\newblock \doi{10.1007/s11071-022-08178-9}

\bibitem{BollerslevWooldridge1992EconomRev}
Bollerslev, T., Wooldridge, J.M.: Quasi-maximum likelihood estimation and
  inference in dynamic models with time-varying covariances.
\newblock Econom. Rev. \textbf{11}(2), 143--172 (1992).
\newblock \doi{10.1080/07474939208800229}

\bibitem{BonyadiMichalewicz2017EvolComput}
Bonyadi, M.R., Michalewicz, Z.: Particle {{Swarm Optimization}} for {{Single
  Objective Continuous Space Problems}}: {{A Review}}.
\newblock Evol. Comput. \textbf{25}(1), 1--54 (2017).
\newblock \doi{10.1162/EVCO_r_00180}

\bibitem{Bortz2006JCritCare}
Bortz, D.M.: Accurate {{Model Selection Computations}}.
\newblock J. Crit. Care \textbf{21}(4), 359 (2006)

\bibitem{Brunel2008ElectronJStat}
Brunel, N.J.B.: Parameter estimation of {{ODE}}'s via nonparametric estimators.
\newblock Electron. J. Stat. \textbf{2}(0), 1242--1267 (2008).
\newblock \doi{10.1214/07-EJS132}

\bibitem{BrunelClairondAlche-Buc2014JAmStatAssoc}
Brunel, N.J.B., Clairon, Q., {d'Alch{\'e}-Buc}, F.: Parametric {{Estimation}}
  of {{Ordinary Differential Equations With Orthogonality Conditions}}.
\newblock J. Am. Stat. Assoc. \textbf{109}(505), 173--185 (2014).
\newblock \doi{10.1080/01621459.2013.841583}

\bibitem{BruntonProctorKutz2016ProcNatlAcadSci}
Brunton, S.L., Proctor, J.L., Kutz, J.N.: Discovering governing equations from
  data by sparse identification of nonlinear dynamical systems.
\newblock Proc. Natl. Acad. Sci. \textbf{113}(15), 3932--3937 (2016).
\newblock \doi{10.1073/pnas.1517384113}

\bibitem{CalderheadGirolamiLawrence2008AdvNeuralInfProcessSyst}
Calderhead, B., Girolami, M., Lawrence, N.D.: Accelerating {{Bayesian
  Inference}} over {{Nonlinear Differential Equations}} with {{Gaussian
  Processes}}.
\newblock In: D.~Koller, D.~Schuurmans, Y.~Bengio, L.~Bottou (eds.) Adv.
  {{Neural Inf}}. {{Process}}. {{Syst}}., vol.~21. {Curran Associates, Inc.}
  (2008)

\bibitem{Dattner2021WIREsCompStat}
Dattner, I.: Differential equations in data analysis.
\newblock WIREs Comp Stat \textbf{13}(6) (2021).
\newblock \doi{10.1002/wics.1534}

\bibitem{DattnerMillerPetrenkoEtAl2017JRSocInterface}
Dattner, I., Miller, E., Petrenko, M., Kadouri, D.E., Jurkevitch, E., Huppert,
  A.: Modelling and parameter inference of predator\textendash prey dynamics in
  heterogeneous environments using the direct integral approach.
\newblock J. R. Soc. Interface. \textbf{14}(126), 20160,525 (2017).
\newblock \doi{10.1098/rsif.2016.0525}

\bibitem{DingWu2014StatSin}
Ding, A.A., Wu, H.: Estimation of ordinary differential equation parameters
  using constrained local polynomial regression.
\newblock Stat. Sin. \textbf{24}(4), 1613--1631 (2014).
\newblock \doi{10.5705/ss.2012.304}

\bibitem{DuistermaatKolk2010}
Duistermaat, J., Kolk, J.: Distributions.
\newblock {Birkh\"auser Boston}, {Boston} (2010).
\newblock \doi{10.1007/978-0-8176-4675-2}

\bibitem{FaselKutzBruntonEtAl2021ArXiv211110992CsMath}
Fasel, U., Kutz, J.N., Brunton, B.W., Brunton, S.L.: Ensemble-{{SINDy}}:
  {{Robust}} sparse model discovery in the low-data, high-noise limit, with
  active learning and control.
\newblock ArXiv211110992 Cs Math  (2021)

\bibitem{FitzHugh1961BiophysJ}
FitzHugh, R.: Impulses and {{Physiological States}} in {{Theoretical Models}}
  of {{Nerve Membrane}}.
\newblock Biophys. J. \textbf{1}(6), 445--466 (1961).
\newblock \doi{10.1016/S0006-3495(61)86902-6}

\bibitem{Fornberg1988MathComput}
Fornberg, B.: Generation of finite difference formulas on arbitrarily spaced
  grids.
\newblock Math. Comput. \textbf{51}(184), 699--699 (1988).
\newblock \doi{10.1090/S0025-5718-1988-0935077-0}

\bibitem{Greenberg1951NACATN2340}
Greenberg, H.: A survey of methods for determining stability parameters of an
  airplance from dyanmics flight measurements.
\newblock Tech. Rep. NACA TN 2340, {Ames Aeronautical Laboratory}, {Moffett
  Field, CA} (1951)

\bibitem{GurevichReinboldGrigoriev2019Chaos}
Gurevich, D.R., Reinbold, P.A.K., Grigoriev, R.O.: Robust and optimal sparse
  regression for nonlinear {{PDE}} models.
\newblock Chaos \textbf{29}(10), 103,113 (2019).
\newblock \doi{10.1063/1.5120861}

\bibitem{HindmarshRose1984ProcRSocLondB}
Hindmarsh, J.L., Rose, R.M.: A model of neuronal bursting using three coupled
  first order differential equations.
\newblock Proc. R. Soc. Lond. B. \textbf{221}(1222), 87--102 (1984).
\newblock \doi{10.1098/rspb.1984.0024}

\bibitem{Jorgensen2012EncyclopediaofEnvironmetrics}
Jorgensen, M.: Iteratively {{Reweighted Least Squares}}.
\newblock In: A.H. El-Shaarawi, W.W. Piegorsch (eds.) Encyclopedia of
  {{Environmetrics}}, first edn. {Wiley} (2012).
\newblock \doi{10.1002/9780470057339.vai022}

\bibitem{KaptanogludeSilvaFaselEtAl2022JOSS}
Kaptanoglu, A., {de Silva}, B., Fasel, U., Kaheman, K., Goldschmidt, A.,
  Callaham, J., Delahunt, C., Nicolaou, Z., Champion, K., Loiseau, J.C., Kutz,
  J., Brunton, S.: {{PySINDy}}: {{A}} comprehensive {{Python}} package for
  robust sparse system identification.
\newblock JOSS \textbf{7}(69), 3994 (2022).
\newblock \doi{10.21105/joss.03994}

\bibitem{KhanmohamadiXu2009Chaos}
Khanmohamadi, O., Xu, D.: Spatiotemporal system identification on nonperiodic
  domains using {{Chebyshev}} spectral operators and system reduction
  algorithms.
\newblock Chaos \textbf{19}(3), 033,117 (2009).
\newblock \doi{10.1063/1.3180843}

\bibitem{KirkThorneStumpf2013CurrOpinBiotechnol}
Kirk, P., Thorne, T., Stumpf, M.P.: Model selection in systems and synthetic
  biology.
\newblock Curr. Opin. Biotechnol. \textbf{24}(4), 767--774 (2013).
\newblock \doi{10.1016/j.copbio.2013.03.012}

\bibitem{LaxMilgram1955ContributionstotheTheoryofPartialDifferentialEquations}
Lax, P.D., Milgram, A.N.: {{IX}}. {{Parabolic Equations}}, \emph{Annals of
  {{Mathematical Studies}}}, vol.~33, pp. 167--190.
\newblock {Princeton University Press} (1955).
\newblock \doi{10.1515/9781400882182-010}

\bibitem{LiangWu2008JournaloftheAmericanStatisticalAssociation}
Liang, H., Wu, H.: Parameter {{Estimation}} for {{Differential Equation Models
  Using}} a {{Framework}} of {{Measurement Error}} in {{Regression Models}}.
\newblock Journal of the American Statistical Association \textbf{103}(484),
  1570--1583 (2008).
\newblock \doi{10.1198/016214508000000797}

\bibitem{Ljung1999}
Ljung, L.: System Identification: Theory for the User, second edn.
\newblock Prentice {{Hall}} Information and System Sciences Series. {Prentice
  Hall PTR}, {Upper Saddle River, NJ} (1999)

\bibitem{Ljung2017WileyEncyclopediaofElectricalandElectronicsEngineering}
Ljung, L.: System {{Identification}}, pp. 1--19.
\newblock {John Wiley \& Sons, Inc.}, {Hoboken, NJ, USA} (2017).
\newblock \doi{10.1002/047134608X.W1046.pub2}

\bibitem{LoebCahen1965Automatisme}
Loeb, J., Cahen, G.M.: Extraction a partir des enregistrements de mesures, des
  parametres dynamiques d'un systeme.
\newblock Automatisme \textbf{8}, 479--486 (1965)

\bibitem{LoebCahen1965IEEETransAutomControl}
Loeb, J., Cahen, G.M.: More about process identification.
\newblock IEEE Trans. Autom. Control \textbf{10}(3), 359--361 (1965).
\newblock \doi{10.1109/TAC.1965.1098172}

\bibitem{Lotka1978TheGoldenAgeofTheoreticalEcology1923-1940}
Lotka, A.J.: The Growth of Mixed Populations: {{Two}} Species Competing for a
  Common Food Supply, vol.~22, pp. 274--286.
\newblock {Springer Berlin Heidelberg}, {Berlin, Heidelberg} (1978).
\newblock \doi{10.1007/978-3-642-50151-7_12}

\bibitem{MacdonaldHusmeier2015BioinformaticsandBiomedicalEngineering}
Macdonald, B., Husmeier, D.: Computational {{Inference}} in {{Systems
  Biology}}.
\newblock In: F.~Ortu{\~n}o, I.~Rojas (eds.) Bioinformatics and {{Biomedical
  Engineering}}, vol. 9044, pp. 276--288. {Springer International Publishing},
  {Cham} (2015).
\newblock \doi{10.1007/978-3-319-16480-9_28}

\bibitem{Martina-PerezSimpsonBaker2021ProcRSocA}
{Martina-Perez}, S., Simpson, M.J., Baker, R.E.: Bayesian uncertainty
  quantification for data-driven equation learning.
\newblock Proc. R. Soc. A. \textbf{477}(2254), 20210,426 (2021).
\newblock \doi{10.1098/rspa.2021.0426}

\bibitem{MessengerBortz2021JComputPhys}
Messenger, D.A., Bortz, D.M.: Weak {{SINDy For Partial Differential
  Equations}}.
\newblock J. Comput. Phys. \textbf{443}, 110,525 (2021).
\newblock \doi{10.1016/j.jcp.2021.110525}

\bibitem{MessengerBortz2021MultiscaleModelSimul}
Messenger, D.A., Bortz, D.M.: Weak {{SINDy}}: {{Galerkin-Based Data-Driven
  Model Selection}}.
\newblock Multiscale Model. Simul. \textbf{19}(3), 1474--1497 (2021).
\newblock \doi{10.1137/20M1343166}

\bibitem{MessengerBortz2022arXiv221116000}
Messenger, D.A., Bortz, D.M.: Asymptotic consistency of the {{WSINDy}}
  algorithm in the limit of continuum data.
\newblock arXiv:2211.16000 (submitted) (2022)

\bibitem{MessengerBortz2022PhysicaD}
Messenger, D.A., Bortz, D.M.: Learning mean-field equations from particle data
  using {{WSINDy}}.
\newblock Physica D \textbf{439}, 133,406 (2022).
\newblock \doi{10.1016/j.physd.2022.133406}

\bibitem{MessengerDallAneseBortz2022ProcThirdMathSciMachLearnConf}
Messenger, D.A., Dall'Anese, E., Bortz, D.M.: Online {{Weak-form Sparse
  Identification}} of {{Partial Differential Equations}}.
\newblock In: Proc. {{Third Math}}. {{Sci}}. {{Mach}}. {{Learn}}. {{Conf}}.,
  \emph{Proceedings of {{Machine Learning Research}}}, vol. 190, pp. 241--256.
  {PMLR} (2022)

\bibitem{MessengerWheelerLiuEtAl2022JRSocInterface}
Messenger, D.A., Wheeler, G.E., Liu, X., Bortz, D.M.: Learning {{Anisotropic
  Interaction Rules}} from {{Individual Trajectories}} in a {{Heterogeneous
  Cellular Population}}.
\newblock J. R. Soc. Interface \textbf{19}(195) (2022).
\newblock \doi{10.1098/rsif.2022.0412}

\bibitem{NardiniBortz2019InverseProbl}
Nardini, J.T., Bortz, D.M.: The influence of numerical error on parameter
  estimation and uncertainty quantification for advective {{PDE}} models.
\newblock Inverse Probl. \textbf{35}(6), 065,003 (2019).
\newblock \doi{10.1088/1361-6420/ab10bb}

\bibitem{NicolaouHuoChenEtAl2023arXiv230102673}
Nicolaou, Z.G., Huo, G., Chen, Y., Brunton, S.L., Kutz, J.N.: Data-driven
  discovery and extrapolation of parameterized pattern-forming dynamics (2023)

\bibitem{NiuRogersFilipponeEtAl2016Proc33rdIntConfMachLearn}
Niu, M., Rogers, S., Filippone, M., Husmeier, D.: Fast {{Inference}} in
  {{Nonlinear Dynamical Systems}} using {{Gradient Matching}}.
\newblock In: Proc. 33rd {{Int}}. {{Conf}}. {{Mach}}. {{Learn}}., vol.~48, pp.
  1699--1707. {PMLR} (2016)

\bibitem{PantazisTsamardinos2019Bioinformatics}
Pantazis, Y., Tsamardinos, I.: A unified approach for sparse dynamical system
  inference from temporal measurements.
\newblock Bioinformatics \textbf{35}(18), 3387--3396 (2019).
\newblock \doi{10.1093/bioinformatics/btz065}

\bibitem{PerdreauvilleGoodson1966JBasicEng}
Perdreauville, F.J., Goodson, R.E.: Identification of {{Systems Described}} by
  {{Partial Differential Equations}}.
\newblock J. Basic Eng. \textbf{88}(2), 463--468 (1966).
\newblock \doi{10.1115/1.3645880}

\bibitem{PoytonVarziriMcAuleyEtAl2006ComputersChemicalEngineering}
Poyton, A., Varziri, M., McAuley, K., McLellan, P., Ramsay, J.: Parameter
  estimation in continuous-time dynamic models using principal differential
  analysis.
\newblock Computers \& Chemical Engineering \textbf{30}(4), 698--708 (2006).
\newblock \doi{10.1016/j.compchemeng.2005.11.008}

\bibitem{PreisigRippin1993ComputChemEng}
Preisig, H., Rippin, D.: Theory and application of the modulating function
  method\textemdash{{I}}. {{Review}} and theory of the method and theory of the
  spline-type modulating functions.
\newblock Comput. Chem. Eng. \textbf{17}(1), 1--16 (1993).
\newblock \doi{10.1016/0098-1354(93)80001-4}

\bibitem{RamsayHookerCampbellEtAl2007JRStatSocSerBStatMethodol}
Ramsay, J.O., Hooker, G., Campbell, D., Cao, J.: Parameter estimation for
  differential equations: A generalized smoothing approach.
\newblock J. R. Stat. Soc. Ser. B Stat. Methodol. \textbf{69}(5), 741--796
  (2007).
\newblock \doi{10.1111/j.1467-9868.2007.00610.x}

\bibitem{ReinboldGurevichGrigoriev2020PhysRevE}
Reinbold, P.A.K., Gurevich, D.R., Grigoriev, R.O.: Using noisy or incomplete
  data to discover models of spatiotemporal dynamics.
\newblock Phys. Rev. E \textbf{101}(1), 010,203 (2020).
\newblock \doi{10.1103/PhysRevE.101.010203}

\bibitem{RudyBruntonProctorEtAl2017SciAdv}
Rudy, S.H., Brunton, S.L., Proctor, J.L., Kutz, J.N.: Data-driven discovery of
  partial differential equations.
\newblock Sci. Adv. \textbf{3}(4), e1602,614 (2017).
\newblock \doi{10.1126/sciadv.1602614}

\bibitem{Sangalli2021InternationalStatisticalReview}
Sangalli, L.M.: Spatial {{Regression With Partial Differential Equation
  Regularisation}}.
\newblock International Statistical Review \textbf{89}(3), 505--531 (2021).
\newblock \doi{10.1111/insr.12444}

\bibitem{SchaefferMcCalla2017PhysRevE}
Schaeffer, H., McCalla, S.G.: Sparse model selection via integral terms.
\newblock Phys. Rev. E \textbf{96}(2) (2017).
\newblock \doi{10.1103/PhysRevE.96.023302}

\bibitem{SchoeberlEichler-JonssonGillesEtAl2002NatBiotechnol}
Schoeberl, B., {Eichler-Jonsson}, C., Gilles, E.D., M{\"u}ller, G.:
  Computational modeling of the dynamics of the {{MAP}} kinase cascade
  activated by surface and internalized {{EGF}} receptors.
\newblock Nat Biotechnol \textbf{20}(4), 370--375 (2002).
\newblock \doi{10.1038/nbt0402-370}

\bibitem{Schwartz1950}
Schwartz, L.: Th\'eorie Des Distributions, vol.~I.
\newblock {Hermann et Cie}, {Paris, France} (1950)

\bibitem{ShapiroWilk1965Biometrika}
Shapiro, S.S., Wilk, M.B.: An analysis of variance test for normality (complete
  samples).
\newblock Biometrika \textbf{52}(3-4), 591--611 (1965).
\newblock \doi{10.1093/biomet/52.3-4.591}

\bibitem{Shinbrot1954NACATN3288}
Shinbrot, M.: On the analysis of linear and nonlinear dynamical systems for
  transient-response data.
\newblock Tech. Rep. NACA TN 3288, {Ames Aeronautical Laboratory}, {Moffett
  Field, CA} (1954)

\bibitem{VanHuffelLemmerling2002}
Van~Huffel, S., Lemmerling, P. (eds.): Total {{Least Squares}} and
  {{Errors-in-Variables Modeling}}: {{Analysis}}, {{Algorithms}} and
  {{Applications}}.
\newblock {Springer Netherlands}, {Dordrecht} (2002).
\newblock \doi{10.1007/978-94-017-3552-0}

\bibitem{vanLaarhovenAarts1987}
{van Laarhoven}, P.J.M., Aarts, E.H.L.: Simulated {{Annealing}}: {{Theory}} and
  {{Applications}}.
\newblock {Springer Netherlands}, {Dordrecht} (1987).
\newblock \doi{10.1007/978-94-015-7744-1}

\bibitem{Varah1982SIAMJSciandStatComput}
Varah, J.M.: A {{Spline Least Squares Method}} for {{Numerical Parameter
  Estimation}} in {{Differential Equations}}.
\newblock SIAM J. Sci. and Stat. Comput. \textbf{3}(1), 28--46 (1982).
\newblock \doi{10.1137/0903003}

\bibitem{VyshemirskyGirolami2008Bioinformatics}
Vyshemirsky, V., Girolami, M.A.: Bayesian ranking of biochemical system models.
\newblock Bioinformatics \textbf{24}(6), 833--839 (2008).
\newblock \doi{10.1093/bioinformatics/btm607}

\bibitem{WangZhou2021IntJUncertaintyQuantification}
Wang, H., Zhou, X.: Explicit estimation of derivatives from data and
  differential equations by {{Gaussian}} process regression.
\newblock Int. J. UncertaintyQuantification \textbf{11}(4), 41--57 (2021).
\newblock \doi{10.1615/Int.J.UncertaintyQuantification.2021034382}

\bibitem{WangHuanGarikipati2019ComputMethodsApplMechEng}
Wang, Z., Huan, X., Garikipati, K.: Variational system identification of the
  partial differential equations governing the physics of pattern-formation:
  {{Inference}} under varying fidelity and noise.
\newblock Comput. Methods Appl. Mech. Eng. \textbf{356}, 44--74 (2019).
\newblock \doi{10.1016/j.cma.2019.07.007}

\bibitem{WenkAbbatiOsborneEtAl2020AAAI}
Wenk, P., Abbati, G., Osborne, M.A., Sch{\"o}lkopf, B., Krause, A., Bauer, S.:
  {{ODIN}}: {{ODE-Informed Regression}} for {{Parameter}} and {{State
  Inference}} in {{Time-Continuous Dynamical Systems}}.
\newblock AAAI \textbf{34}(04), 6364--6371 (2020).
\newblock \doi{10.1609/aaai.v34i04.6106}

\bibitem{XuKhanmohamadi2008Chaos}
Xu, D., Khanmohamadi, O.: Spatiotemporal system reconstruction using
  {{Fourier}} spectral operators and structure selection techniques.
\newblock Chaos \textbf{18}(4), 043,122 (2008).
\newblock \doi{10.1063/1.3030611}

\bibitem{YangWongKou2021ProcNatlAcadSciUSA}
Yang, S., Wong, S.W.K., Kou, S.C.: Inference of dynamic systems from noisy and
  sparse data via manifold-constrained {{Gaussian}} processes.
\newblock Proc Natl Acad Sci USA \textbf{118}(15), e2020397,118 (2021).
\newblock \doi{10.1073/pnas.2020397118}

\bibitem{ZhangNanshanCao2022StatComput}
Zhang, N., Nanshan, M., Cao, J.: A {{Joint}} estimation approach to sparse
  additive ordinary differential equations.
\newblock Stat Comput \textbf{32}(5), 69 (2022).
\newblock \doi{10.1007/s11222-022-10117-y}

\end{thebibliography}

\end{document}